\documentclass{article}

\usepackage{PRIMEarxiv}
\usepackage{multirow}
\usepackage[utf8]{inputenc} 
\usepackage[T1]{fontenc}    
\usepackage{hyperref}       
\usepackage{url}            
\usepackage{booktabs}       
\usepackage{amsfonts}       
\usepackage{nicefrac}       
\usepackage{microtype}      
\usepackage{lipsum}
\usepackage{fancyhdr}       
\usepackage{graphicx}       
\graphicspath{{media/}}     
\usepackage{amsmath}
\usepackage{algorithm}
\usepackage{algorithmic}
\title{Uncertainty quantification with approximate variational learning for wearable photoplethysmography prediction tasks
}

\author{
  Ciaran Bench \\
  Department of Data Science and AI \\
  National Physical Laboratory \\
  Teddington, UK\\
  \texttt{ciaran.bench@npl.co.uk} \\
  \And
  Vivek Desai \\
  Department of Data Science and AI \\
  National Physical Laboratory \\
  Teddington, UK\\
  \texttt{vivek.desai@npl.co.uk} \\
  \And
  Mohammad Moulaeifard \\
  Carl von Ossietzky Universität Oldenburg \\ Oldenburg, Germany \\
  \texttt{mohammad.moulaeifard@uni-oldenburg.de} \\
  \And
  Nils Strodthoff \\
  Carl von Ossietzky Universität Oldenburg \\ Oldenburg, Germany \\
  \texttt{nils.strodthoff@uni-oldenburg.de} \\
  \And
  Philip Aston \\
  Department of Data Science and AI \\
  National Physical Laboratory \\
  Teddington, UK\\
  \texttt{philip.aston@npl.co.uk} \\
  \And
  Andrew Thompson \\
  Department of Data Science and AI \\
  National Physical Laboratory \\
  Teddington, UK\\
  \texttt{andrew.thompson@npl.co.uk} \\
}

\begin{document}
\maketitle

\begin{abstract}
Photoplethysmography (PPG) signals encode information about relative changes in blood volume that can be used to assess various aspects of cardiac health non-invasively, e.g.\ to detect atrial fibrillation (AF) or predict blood pressure (BP). Deep networks are well-equipped to handle the large quantities of data acquired from wearable measurement devices. However, they lack interpretability and are prone to overfitting, leaving considerable risk for poor performance on unseen data and misdiagnosis. Here, we describe the use of two scalable uncertainty quantification techniques: Monte Carlo Dropout and the recently proposed Improved Variational Online Newton. These techniques are used to assess the trustworthiness of models trained to perform AF classification and BP regression from raw PPG time series. We find that the choice of hyperparameters has a considerable effect on the predictive performance of the models and on the quality and composition of predicted uncertainties. E.g. the stochasticity of the model parameter sampling determines the proportion of the total uncertainty that is aleatoric, and has varying effects on predictive performance and calibration quality dependent on the chosen uncertainty quantification technique and the chosen expression of uncertainty. We find significant discrepancy in the quality of uncertainties over the predicted classes, emphasising the need for a thorough evaluation protocol that assesses local and adaptive calibration. This work suggests that the choice of hyperparameters must be carefully tuned to balance predictive performance and calibration quality, and that the optimal parameterisation may vary depending on the chosen expression of uncertainty.
\end{abstract}


\section{Introduction}

Several indicators of patient health can be monitored from measured changes in blood flow and composition, such as arterial blood oxygen saturation, cardiac rhythm, heart rate,  blood pressure, and atrial fibrillation \cite{park2022photoplethysmogram}. This information can be detected from photoplethysmography (PPG) signals which encode changes in the optical properties of the skin's microvascular bed driven by variations in blood volume. 

With PPG, the skin bed is illuminated with near infrared light at various wavelengths, where the tissue's optical properties change over the course of measurement due to a variety of physiological factors (including changes in arterial and venous blood volume/flow). This results in a variation in the amount of light reflected/backscattered into a sensor positioned around the same location as the illumination source (e.g. as is the case with smartwatch-based measurement devices); this can be collected to form a time-series signal \cite{charlton2022wearable}. Sensors may also be placed opposite the illumination source to collect transmitted light (e.g. for pulse oximeters that can be placed on the finger).

More recent advancements in measurement technology have given rise to wearable PPG sensing devices that enable patient monitoring outside of typical clinical settings. This can drastically improve the frequency, accuracy, and ease with which one can detect risk factors related to cardiovascular disease \cite{charlton2022wearable, Torres-Soto_Ashley_2020,shcherbina2017accuracy,wasserlauf2019smartwatch}. However, classical methods of analysis are not suited to efficiently processing the large quantities of time-series data collected by these devices. Automated techniques for analysing PPG data are needed to realise the potential of these new sensing capabilities.

Two tasks of particular clinical relevance are blood pressure estimation, and the detection of atrial fibrillation. It is known that blood pressure measured in the clinic can differ from levels experienced at home or in the office. Therefore, home monitoring of blood pressure is often recommended for more effective management of hypertension \cite{krakoff2016blood,o2008ambulatory,piper2015diagnostic,parati2008european}. However, this usually involves cumbersome cuff-based devices that cannot be used for continuous monitoring, spurring a desire for more convenient technologies. Episodes of atrial fibrillation involve periods of rapid, uncoordinated heartbeat and correlate with an increased risk of stroke and other ailments. Anticoagulation treatments are effective in reducing this risk, but episodes are underdiagnosed using current clinical procedures \cite{zungsontiporn2018newer,ding2020emerging}. For both prediction tasks, the use of wearables data could enhance screening capabilities by improving the ease of measurement, providing a means of continuous monitoring and increasing the availability of measurement devices. Therefore, their use holds considerable potential to improve patient outcomes.

The prediction of various physiological parameters from PPG signals is often highly non-trivial, owing to the nonlinearity of the mappings, and due to the presence of confounding experimentals factors such as motion artefacts, noise, and poor calibration. Analytical approaches have been proposed to perform a wide range of prediction tasks using PPG signals \cite{almarshad2022diagnostic}, such as the use of a single generalised transfer function for predicting arterial stiffness from a a single digital volume pulse waveform \cite{brillante2008arterial}, the combination of ensemble empirical mode decomposition with independent component analysis and non-negative matrix factorization in the prediction of respiratory rate \cite{lei2020estimation}, and the use of physical light propagation models in the estimation of arterial blood oxygen saturation from measured R values \cite{nitzan2014calibration}. However, data-driven approaches based on deep neural networks stand apart as particularly effective and versatile given (i) their capacity to approximate highly complex nonlinear mappings without the need to define some convenient/domain specific analytical form of the mapping function, (ii) their tolerance to experimental artefacts/noise, and (iii) their ability to learn feature extraction automatically, allowing them to be used effectively in cases where ideal feature representations are not straightforward to manually extract or are not known \textit{a priori}.

Here we focus on the AF classification and blood pressure prediction tasks. Indeed, deep neural networks have been trained to estimate blood pressure from single PPG time-series measurements \cite{gonzalez2023benchmark,el2020review,maqsood2022survey,paviglianiti2022comparison}, providing a novel means to avoid challenges related to using multiple sensor measurements such as mis-matched sampling rates, synchronisation, and the unappealing nature of having to perform per-patient calibration \cite{el2020review,gonzalez2023benchmark}. Similar work has been conducted for the classification task of detecting atrial fibrillation \cite{shashikumar2017deep,tison2018passive,shen2019ambulatory,gotlibovych2018end}. Their capabilities for automatic feature extraction have been found to be useful for both tasks given (i) the challenges with formulating hand-crafted features that are consistent across PPG environments \cite{Torres-Soto_Ashley_2020}, and (ii) because they do not rely on the unreliable distance-based metrics often involved with these hand-crafted feature-based approaches. The use of manually extracted features as inputs to networks are reported in \cite{tison2018passive,shashikumar2017deep} for AF classification and in \cite{kurylyak2013neural} for BP estimation. However, a fully end-to-end approach using raw time-series as inputs remains appealing as this removes the potential risk of degrading model performance with misguided assumptions about the most informative features to use for training.

Here, we train models for the AF and BP predictions tasks, and assess the trustworthiness of their outputs by applying uncertainty quantification techniques to predict the uncertainty in each prediction. While it is known that the parameterisation of the model/optimisation can have an effect on the quality of uncertainty estimates, this is often underexplored in practical implementations. We analyse the effect using different parameterisations can have on both calibration quality (i.e. how well the predicted  uncertainties reflect the underlying doubt in a given prediction), as well as predictive performance. Furthermore, we show how these effects may change depending on the chosen UQ technique, and expression of uncertainty. The optimal balance between calibration quality and predictive performance may be assessed with a rigorous evaluation protocol. We use several metrics to assess predictive performance and calibration quality. We also disentangle uncertainties to see how the choice of hyperparameters may affect the composition of uncertainties.

\subsection{Uncertainty quantification for deep neural networks}

Deep neural networks are often used as deterministic models, outputting just a single estimate for a given input. However, in a typical measurement setting, the uncertainty of a model's prediction is inferred from a distribution of possible outputs, where the variance of this distribution encodes the degree of doubt about the prediction \cite{witkovsky2017brief}. The training/evaluation of deep networks can be reformulated to provide distributions of predictions that (to varying degrees) accurately represent uncertainty \cite{gal2016dropout}. While classification models output a distribution over the classes, this does not necessarily capture all sources of uncertainty relevant to a prediction task. In any given measurement scenario, there are several possible sources of uncertainty \cite{hullermeier2021aleatoric}. Two types of uncertainty in neural networks are often distinguished: \emph{aleatoric} and \emph{epistemic} uncertainty. Aleatoric uncertainty is the inherent uncertainty in the learning task, including that which is caused by uncertainty in the data used for training/evaluation, such as noise. Epistemic uncertainty is the uncertainty concerning the model. This type of uncertainty can be reduced with more data or a more efficient modelling framework (e.g.\ formulated using additional knowledge about the problem/task), whereas aleatoric uncertainty is irreducible. Insight into whether uncertainty is inherent to the task or a property of the modelling approach is valuable. It is therefore advantageous to adopt an approach in which aleatoric and epistemic contributions to prediction uncertainty can be disentangled (though, most commonly implemented approaches are not particularly effective \cite{gruber2023sources,mucsanyi2024benchmarking}). In particular, isolating the effect of epistemic uncertainty can be used to detect out-of-distribution data to which the model is not robust~\cite{kendall2017uncertainties}. It is also important to capture \emph{both} aleatoric and epistemic uncertainty. A frequently observed practice is to use a technique which captures only one of the two, usually epistemic uncertainty. Such an approach, however, cannot be considered to give a comprehensive quantification of uncertainty.

In this paper, we investigate the use of methods for capturing both aleatoric and epistemic uncertainty for AF detection and BP estimation deep learning models. The choice of model parameterisation can have a sigificant effect on the quality of uncertainties, so here, we investigate how various choices may affect calibration quality.

\subsection{Methods for capturing both aleatoric and epistemic uncertainty}

Aleatoric uncertainty for neural networks is typically estimated by assuming some statistical distribution for the model output and selecting a loss function which maximises the corresponding likelihood. For classification problems (such as AF detection), aleatoric uncertainty can be captured by the output class probabilities; the popular categorical cross-entropy loss function corresponds to a categorical distribution on the output classification (or a Bernoulli distribution for the special case of binary classification). Extensions to this approach which also take into account uncertainties on the class probabilities themselves have also been proposed, for example in~\cite{kendall2017uncertainties}. For regression problems (such as BP prediction), it is common to model aleatoric uncertainty as additive (heteroscedastic) noise and to learn the parameters of the predictive distribution, for example the mean and variance of a Gaussian distribution as in~\cite{kendall2017uncertainties,lakshminarayanan2017simple}.

For epistemic uncertainty, Bayesian inference is a commonly used framework which can, in principle, be applied to deep neural networks (Bayesian neural networks) \cite{jospin2022hands,mackay1995bayesian}. In one common implementation, a prior is placed over the model parameters, and Bayes' theorem is used to derive the posterior distribution over the parameters. However, in most real world applications, the number of model parameters is high, which makes this approach intractable. This remains the case even with the use of well-known approximation schemes such as variational inference \cite{gal2016dropout}. Several approaches have been implemented to overcome these challenges. 

One class of approaches makes use of assumptions about the form of the posterior that give rise to more convenient optimisation schemes. Laplace approximation, for example, involves the Gaussian approximation of the posterior over its mode (acquired with maximum a posteriori estimation) \cite{mackay1992practical,duffield2024scalable}. Here, a model is trained with generic supervised learning, and then subsequently trained for an additional epoch to acquire an estimate of the covariance matrix. However, storing/evaluating this covariance matrix is intractable for larger models, making it unappealing in practical applications.

Natural gradient methods are another approach, where the stochastic gradient of an expected, regularised loss is used to update parameter values under a Gaussian prior. This expectation is computed over several evaluations of a given input, each corresponding to a model whose parameters are sampled from the variational distribution \cite{osawa2019practical}. When the variational distribution also has a Gaussian form, the expressions for updating parameters become similar to those associated with common optimisers, such as stochastic gradient descent and Adam \cite{kingma2014adam}. Natural gradient methods can be thought of as adapted versions of common optimisation schemes that perform variational learning. The Improved Variational Online Newton (IVON) is one form that employs Adam-like updates with a Newton-like estimate for the Hessian, which improves efficiency over other natural gradient methods (e.g. the Variational Online Newton, which employs a Gauss-Newton estimate), and has been shown to be effective on larger scale models \cite{shen2024variational}. However, this recently-proposed technique has not been applied to a wide range of tasks, and it remains to be seen whether its proposed benefits are easily realised in practical training scenarios. Indeed, it could impose limitations that hinder its use on more challenging tasks; for example, the optimiser does not work well with models featuring batch normalisation \cite{shen2024variational}.

Monte Carlo Dropout (MCD) is another technique that employs an approximation of variational learning. Here, a single model is trained with dropout regularisation, usually applied liberally throughout the architecture \cite{gal2016dropout,kendall2015bayesian}. At evaluation, dropout regularisation remains active, and each example is evaluated several times. The resulting distribution of outputs encodes the predictive variance, from which one can compute an uncertainty. The method is reasoned from Bayesian principles, where training/evaluating with dropout has been shown to approximate variational inference \cite{gal2016dropout}. Although reports have shown that the technique may produce poorly calibrated uncertainties compared to other approaches, it remains an appealing option for uncertainty quantification as it only requires a single training run \cite{fort2019deep}. This makes MCD compatible with larger models and datasets, making it appealing for models trained to predict AF and BP from large datasets of raw time-series. That said, the magnitude of the estimated uncertainties are known to depend on the values of certain hyperparameters such as the dropout rate and weight regularisation, where optimal values may be derived with considerable effort using some kind of grid search \cite{gal2015bayesian} or optimised as part of the training process e.g.\ Concrete Dropout \cite{gal2017concrete}.

In contrast to the Bayesian neural network framework in which a prior is placed over the model weights, evidential deep learning is a framework in which training is treated as an evidence gathering process where a prior is placed directly on the likelihood function \cite{amini2020deep}. With this framework, the model predicts the parameters to a distribution that, when sampled, will produce lower-order likelihood functions from which the data was drawn. Recent works suggest that generic methods estimate epistemic uncertainty \cite{gao2024comprehensive}.  Only a single forward pass through the trained model is needed to predict uncertainties. However, the training objective is more complicated than other uncertainty quantification schemes, resulting in less interpretable outputs and a potentially more challenging optimisation. 

Deep Ensembles is another approach for quantifying epistemic uncertainty where the predictive distribution is derived by independently training an ensemble of deep network models, each with a distinct random initialisation of their weights \cite{lakshminarayanan2017simple}. Once trained, each model is fed the same input example, and the resulting distribution of outputs are aggregated to derive an uncertainty. However, the technique can be computationally prohibitive to implement when the model architectures and/or datasets are quite large. While Deep Ensembles may have a Bayesian interpretation \cite{wilson2020bayesian,fort2019deep} (i.e.\ each model is a maximum \textit{a-posteriori} estimate for a given mode of the posterior), the optimisation scheme is not inherently Bayesian. 

By using an appropriate likelihood function, the above techniques for estimating epistemic uncertainty can all in principle be combined with the maximum likelihood approach for estimating aleatoric uncertainty. Approaches of this kind for both regression and classification problems were proposed for MCD in~\cite{kendall2017uncertainties}, and an approach for regression problems was proposed for Deep Ensembles in~\cite{lakshminarayanan2017simple}. Methods for summarising uncertainty quantifications and disentangling aleatoric and epistemic contributions to them were proposed in~\cite{valdenegro2022deeper} for ensembling methods such as MCD and Deep Ensembles.

In this paper, we focus on techniques for capturing epistemic uncertainty derived from Bayesian inference/variational learning, and their combination with the maximum likelihood approach for aleatoric uncertainty. We aim to compare the results of using two scalable approximations of variational learning: IVON and MCD. We consider both the ease of implementation in addition to predictive performance and the quality of uncertainty estimates.

\section{Related work}
\subsection{Monte Carlo Dropout applied to relevant health models}
MCD has been applied to models trained to perform various PPG analysis tasks (in times series form or otherwise) \cite{song2023uncertainty,harper2019end,trudaquantifying,liu2022videocad}. For example, Song et al. \cite{song2023uncertainty}, applied MCD to heart rate estimation models processing remote PPG signals. They disentangled the aleatoric `data' uncertainty from the epistemic `model' uncertainty, and calibrated their uncertainty estimates using a post-hoc technique. They compared their approach to Deep Ensembles, and assessed the quality of estimates using reliability diagrams. They found that the total uncertainty was dominated by the aleatoric uncertainty, and that while both techniques produced accurate uncertainties, Deep Ensembles appeared to outperform MCD. However, they did not motivate their choice of training hyperparamaters for their implementation, and consequently, it is unclear whether their uncalibrated uncertainty estimates could have been further improved with a different choice for the dropout rate. This is of interest as, in some cases, there may not be sufficient data to set aside for the calibration task, and it would be useful to know how this parameter affects the quality of `raw' uncertainty estimates.
\subsubsection{Blood pressure}
In regards to BP prediction, Asgharnezhad et al. \cite{asgharnezhad2023improving} provided a notable example where an enhanced form of MCD was applied to networks trained to estimate normal, prehypertensive, or hypertensive levels of blood pressure (classification task) from raw PPG time-series. Here, a penalty term was added to the loss to reduce the average entropy of output predictions. Additionally, the dropout rate was optimised using a form of Bayesian optimisation. They validated their estimates using a custom metric describing the proportion of examples where incorrect estimates corresponded to high predicted uncertainties and \textit{vice versa}. They reported well calibrated uncertainties for this classification task. 
 
Han et al. \cite{han2023non} used features manually extracted from remote imaging PPG signals as inputs to models trained to predict blood pressure. The estimated uncertainty was shown to increase with model errors, broadly indicating some degree of calibration. 

However, both studies mentioned here were small scale, and a larger test set is needed to provide more conclusive evidence of the quality of the estimated uncertainties. Furthermore, neither implements a rigorous evaluation procedure for assessing calibration quality for the BP regression problem.

\subsubsection{Atrial fibrillation}
MCD was also applied to models that were (in part) trained to predict AF from electrocardiogram (ECG) time-series in \cite{vranken2021uncertainty}. The models were trained to assign one of eight diagnoses: normal, atrial fibrillation, left bundle branch block, right bundle branch block, premature atrial contraction, premature ventricular contraction, ST-segment depression, and ST-segment elevation (Where S and T refer to the S-wave and T-wave of the ECG time series). These results were compared to those acquired using several other uncertainty quantification approaches including variational inference and various forms of deep ensembles. Although the authors did not directly validate their predicted uncertainties, they found that removing samples with high uncertainty from the pool of test examples generally improved accuracy metrics, indicating that the uncertainties were calibrated to some degree. However, a more precise evaluation is needed to assess whether the predicted uncertainty is accurate enough for clinical application. They also showed that the implementation of uncertainty quantification had the effect of improving the performance of their models, where techniques that considered both aleatoric and epistemic uncertainty outperformed those that only considered epistemic uncertainty. They justified this observation with the reasoning that with large amounts of data (as is the case in their study), the model uncertainty will likely be low, and that modelling the dominant aleatoric uncertainty as part of the learning objective should improve performance. They reported that one of their MCD based implementations produced overconfident predictions on out of domain test data, while one of their ensemble-based techniques was the most effective. 

\cite{chen2022quantifying} also report the use of MCD to estimate epistemic uncertainty from models trained on ECG time-series (in this case, acquired from patients that were checked into the intensive care unit). A surrogate classifier was used to generate labels for the data, where the subsequent use of these weak labels in the training of a separate model via confident learning, and the incorporation of both epistemic and aleatoric uncertainty quantification (the latter estimated via test-time augmentation) was shown to provide the most optimal balance between prediction performance and uncertainty calibration.

An example involving PPG time-series was reported in \cite{das2020bayesbeat}, where a Bayesian neural network (optimised with Bayes by backpropagation) was trained to predict uncertainties on predictions of AF. Ultimately the technique is less scalable than other uncertainty quantification techniques, and some demonstration of a more efficient technique applied to larger models that have the capacity to learn from large datasets is highly desirable.

\subsection{Applications of the Improved Variational Online Newton}
IVON was applied to several models in \cite{shen2024variational}, e.g.\ for a range of image classification benchmarks, and LLM pretraining/finetuning tasks. It achieved comparable or better predictive performance/calibration quality for several tasks when compared to other uncertainty quantification techniques, and performed well on out-of-distribution detection tasks. This provides a strong indication that the technique can scale well, and provide meaningful uncertainty estimates. However, in contrast to MCD, it has yet to be demonstrated on a large set of problems featuring more realistic datasets. Here we aim to provide an example using real measurement data from wearable devices. The performance of the optimiser is reported to depend on the values of various hyperparameters. We assess the influence these values have on the quality of uncertainty estimates.

\subsection{Our contribution}

MCD and IVON are appealing techniques for quantifying the uncertainties of models trained to predict physiological parameters given their scalable nature and ease of implementation. However, there remains a need to robustly evaluate the quality of uncertainties that can be estimated with these techniques before they can be considered for use in clinical settings. There are several factors to consider when developing an evaluation framework.

For example, epistemic and aleatoric uncertainties are often not completely disentangled from predictions, limiting the insights that can be drawn about model behaviour. This, for example, may complicate efforts to determine how much of the total uncertainty may be epistemic, and thus, whether a different architecture or more data could be used to improve performance. 

Calibration is often assessed globally (i.e.\ over the whole test set), limiting the precision with which one can assess calibration quality over subsets of the data, e.g.\ per class (which may inform procedures for using model-outputs to inform diagnosis), or by signal quality, skin tone, etc.

For MCD, related works mentioned in the previous sections fail to address the effect that the value of the dropout rate has on the accuracy of the predicted uncertainties (and therefore the practical need to optimise this parameter). Although, Concrete Dropout has been proposed as a means to optimise this parameter, it is not used as frequently. This motivates the need for some assessment of how calibration quality varies with the parameter. This can help derive a more optimal configuration for learning calibrated uncertainties. Similarly, the performance of IVON depends on the choice of hyperparameter values (e.g.\ the initialisation of the Hessian) and their effects should ideally be assessed as part of an evaluation of calibration quality. Given the novelty of IVON, it has not yet been implemented in a wide range of domains; this work provides an application to a medical domain involving 1D time series data.

Here, we apply MCD to classification models trained on raw PPG time-series to predict AF, and regression models that simultaneously predict systolic (SBP) and diastolic blood pressure (DBP). We showcase the effect different dropout rates have on the predicted uncertainties. We perform a similar analysis for the initialisation of the Hessian for IVON which we apply to the classification models (which similarly plays a role in the degree of stochasticity in the sampling). We report the novel incorporation of aleatoric uncertainty modelling into a model trained with IVON. We also provide a general and novel method (i.e. distinct from that proposed in \cite{valdenegro2022deeper}) for disentangling the aleatoric uncertainty from the total uncertainty of the classification model outputs to observe how changes in these parameters affects how each source of uncertainty is modelled. Our evaluation is conducted over large-scale datasets utilising a range of metrics to assess overall predictive performance, as well as the sensitivity/adaptivity of the uncertainty estimates for each class \cite{pernot2023validation}; this provides a thorough assessment of model performance. We also perform a similar analysis for uncertainties estimated with MCD applied to models that predict BP. Here, we additionally assess the effectiveness of the uncertainty disentanglement, raising important questions about interpreting estimates as representing distinct sources of uncertainties.

Ultimately, this work highlights various factors that should be considered when applying MCD and IVON to models intended for clinical application and proposes an evaluation framework that can be used to more effectively prototype models with these factors in mind. 

\section{Uncertainty quantification} 

Models trained on raw time-series data acquired with wearable devices have clear potential to automate diagnostic protocols. However, existing studies often do not estimate uncertainties for their predictions. Deep neural networks are prone to overfitting, and lack interpretability. Therefore, there remains considerable risk that poor generalisation to previously unseen test data (i.e.\ poor performance) is left undetected, rendering them unsuitable for routine use in clinical and other settings given the possibility of misdiagnosis. It follows that an accurate estimate of the uncertainty associated with a given prediction is needed to determine whether it can be used to reliably inform diagnoses. 

Neural networks are typically trained to be deterministic models, where evaluating the same input multiple times will always return the same output. This is unappealing in the context of uncertainty quantification, where the doubt about a particular prediction made by a model is usually assessed by observing the distribution of outputs it provides for a given input. Uncertainty quantification methods for deep networks recast training and evaluation so that models will produce a distribution of outputs for a given input, from which one can derive uncertainty estimates. While classification models are deterministic,  and output a distribution over the classes, this does not encode all sources of uncertainty (in particular epistemic uncertainty). Several approaches for modelling this epistemic uncertainty are inspired by Bayesian optimisation, which is a framework for acquiring uncertainties from predictive models in the non-neural network case. 

\subsection{Bayesian optimisation}
\label{sec:bayes_opt}
Given a non-neural network model $f_\theta(x)$, that performs a mapping from some input $x_i \in D_x$ to predict some corresponding ground truth quantity $y_i \in D_y$, with parameters $\theta$ (where the inputs and ground truths are collated into a dataset $D$), one can place a prior distribution over the model's parameters and use Bayes' theorem to derive a posterior distribution (here, forcing independence between the inputs and the parameters) \cite{jospin2022hands},
\begin{equation}
    p(\theta|D) = \frac{p(D_y|D_x,\theta)p(\theta)}{\int_\theta p(D_y|D_x,\theta')p(\theta')d\theta'}.
    \label{eq:posterior}
\end{equation}

This posterior distribution can then be used to evaluate the predictive `marginal' distribution,
\begin{equation}
    p(y|x,D) = \int_{\theta}p(y|x,\theta')p(\theta'|D)d\theta',
\end{equation}
where the variance of this predictive distribution encodes information about the range of plausible predictions for the given input. This can be used to derive an estimate of the uncertainty. A `wider' (higher entropy) distribution implies that the set of possible predictions spans a large range, and that there is significant uncertainty in the prediction for the given input example.

Given that a neural network can ultimately be treated as some analytical function composed of a large number of parameters, this approach, in principle, can be used to optimise its parameters and derive predictive distributions on test data. Models optimised in this way (and with other similar schemes reasoned from Bayesian principles) are referred to as Bayesian Neural Networks \cite{jospin2022hands}. However, the evaluation of the denominator of (\ref{eq:posterior}) (the evidence) is intractable given the large number of parameters. Instead, this term can be approximated by optimising over the evidence lower bound, in a procedure known as variational inference \cite{jospin2022hands}. Yet, even this approximation cannot be implemented efficiently on models composed of a large number of parameters such as deep networks \cite{gal2016dropout}. Thus, there remains a desire for more efficient ways to derive uncertainties from deep networks.

\subsection{ Approximate variational inference with Monte Carlo Dropout}
MCD \cite{gal2016dropout} is one approach that efficiently approximates variational inference. Dropout is applied liberally throughout the architecture during training, and left active at the evaluation stage where each input example is evaluated several times. In essence, each `dropout model' can be interpreted as a sample from the model's posterior distribution, where the subsequent aggregation of their predictions into distributions allows one to derive uncertainty estimates. 

The implementation of MCD varies depending on the type of uncertainties that are considered/estimated. Here, we consider epistemic and aleatoric uncertainty as they both are present in virtually all scenarios in which inferences are made from measurements. Furthermore, we treat the aleatoric uncertainty as heteroscedastic (parameterising the models to predict variance terms). This enables use to model a non-constant aleatoric uncertainty for each example, which is appealing given the effects of confounding experimental factors such as noise are not constant for all measurements. 

\subsubsection{Regression}
For regression, we implement the procedure outlined by Kendall and Gal in \cite{kendall2017uncertainties}. When considering both epistemic and aleatoric uncertainty, a model is not configured to output the desired value directly as is the case in a generic training framework. Instead, the model outputs parameters used to define a Gaussian distribution (e.g.\ a mean $\mu$ and variance $\sigma^2$). The model is typically constructed to have two outputs; one that predicts the mean, and another that predicts the variance. This Gaussian distribution can be sampled to acquire a point prediction. A likelihood-based loss function (i.e.\ a Gaussian log-likelihood that takes the predicted mean and variance as inputs alone) is used to learn the parameters of the model that will ideally output the parameters used to define Gaussian distributions that `describe the data well'. For example, if a particular input example has large amounts of noise, then it could have a large range of plausible predictions associated with it. Within this framework, the model should ideally learn to detect the features associated with high uncertainty/non-uniqueness, and consequently predict a distribution with a large variance if these features are present. A likelihood-based optimisation, therefore, provides a means to model uncertainty as part of the training process. 

For evaluation (pseudocode given in Algorithm 1), dropout is left active, and each example is evaluated $K$ times, where each evaluation is indexed by $k$. Here, the law of total variance can be applied to disentangle the uncertainties, where the epistemic uncertainty can be evaluated by computing the variance of the predicted means $\mu_k$,
 \begin{equation}
      \sigma_{\text{epi}}^{2} = \frac{1}{K}\sum^{K}_{k=1}(\mu_k - \frac{1}{K}\sum_{k=1}^{K}\mu_k)^2 ,
 \end{equation}

and the aleatoric uncertainty can be evaluated by taking the mean of the predicted variances $\sigma_k^2$, 
 \begin{equation}
      \sigma_{\text{ale}}^{2} = \frac{1}{K}\sum_{k=1}^{K}\sigma^{2}_{k} .
 \end{equation}
$K$ should be chosen such that the number of samples adequately approximates the model's posterior distribution. This depends strongly on the task/model/data. Here, we choose the maximum number of samples that still allowed for reasonable execution times. 
 \begin{algorithm}
 \label{alg:eval_regression}
\caption{Monte Carlo Dropout Regression Evaluation Loop}
\begin{algorithmic}
\FOR{each input}
\FOR{$k=1$ to $K$}
\STATE Compute predicted mean $\mu_{k}$ and variance $\sigma^2_{k}$
\ENDFOR
\STATE $\sigma_{\text{epi}}^2= \frac{1}{K}\sum^{K}_{k=1}(\mu_k - \frac{1}{K}\sum_{k=1}^{K}\mu_k)^2$
\STATE $\sigma_{\text{ale}}^2= \frac{1}{K}\sum_{k=1}^{K}\sigma_k^2$
\ENDFOR
\end{algorithmic}
\end{algorithm}

The magnitude of the uncertainties estimated with MCD depend strongly on the chosen dropout rate and weight regularisation, where a grid search over some validation metric is recommended to choose sensible values \cite{gal2015bayesian}. In an effort to derive well-calibrated uncertainties (without using complex schemes for optimising the hyperparameter, like Concrete dropout \cite{gal2017concrete}), we advise the use of a grid search over the dropout rate where, for each model, one computes a calibration quality score for the predicted uncertainties. The optimal dropout rate is that which produces the best calibrated values on a validation set.

\

\subsubsection{Classification}
\textbf{Training:}

Here, we follow the procedure outlined by Kendall and Gal in \cite{kendall2017uncertainties}. When considering both epistemic and aleatoric uncertainty, the model is configured to predict a mean and standard deviation unique to each logit, which are used to parameterise multiple Gaussian distributions (one for each class). Therefore the model is constructed to have four outputs. The value of a given logit can be acquired by sampling from its respective predicted Gaussian distribution. A vector of probabilities may be acquired by applying a Softmax function to the vector of sampled logits. The aim is to optimise the model over the expected negative log-likelihood (NLL) of these probabilities.

Unlike the regression case, a generic Gaussian log-likelihood loss function is not applicable as we are predicting categorical rather than continuous variables. Another more suitable analytical loss is not straightforward to implement in this case, and instead, the training objective is evaluated using Monte Carlo sampling (separate from dropout sampling, which is used during training and evaluation), where for each input, we acquire $T$ sampled logits $x_{i,t}$ (where $i$ indexes each logit, and $t$ is the index of each sampled value), from Normal distributions parameterised by the predicted mean $f_i$ and variance $\sigma^2_i$ of each logit. In practice, this is executed by sampling $T$ instances of Gaussian noise for each logit $\epsilon_{i,t} \sim  \mathcal{N}(0,\sigma_i^2)$, and computing $x_{i,t} = f_i + \epsilon_{i,t}$. Each vector of logits $\textbf{x}_t$ is then passed through a Softmax, producing a vector of probabilities $\textbf{p}_t = \text{Softmax}(\textbf{x}_t)$, which are then averaged over $t$, $\bar{\textbf{p}} = \frac{1}{T}\sum_{t=1}^{T}\textbf{p}_t$. The loss is evaluated as the negative log-likelihood of $\bar{\textbf{p}}$ assuming a categorical distribution. $T$ must have a high enough value to adequately perform the Monte Carlo sampling. Here, we limit the value to 100 to reduce memory consumption, as also selected in a similar study \cite{vranken2021uncertainty}. We propose a novel method for expressing predicted uncertainties, given in Algorithm 3.

\begin{algorithm}

\caption{Monte Carlo Dropout Classification Training Loop}
\begin{algorithmic}
\label{alg:train_class}
\FOR{each batch}
\STATE Compute predicted mean $f_{l,i}$ and variance $\sigma^2_{l,i}$ for each logit $i$ for each example $l$ in the batch
\FOR{$t = 1$ to $T$}
\FOR{each logit $i$}
\STATE Sample $\epsilon_{l,i,t} \sim \mathcal{N}(0,1)$
\STATE Compute sampled logit $x_{l,i,t} = f_{l,i} + \sigma_{l,i}^2\epsilon_{l,i,t}$
\ENDFOR
\STATE Compute $\textbf{p}_{l,t} = \text{Softmax}(\textbf{x}_{l,t})$
\ENDFOR
\STATE Compute average $\bar{\textbf{p}_l} = \frac{1}{T}\sum_{t=1}^{T}\textbf{p}_{l,t}$
\STATE Evaluate NLL on $P$ where $P$ is the batch of $\bar{\textbf{p}}_l$ values.

\ENDFOR
\end{algorithmic}
\end{algorithm}
\textbf{Evaluation:}

Similar to the regression case, for evaluation we keep dropout regularisation active. Each input is passed through the network $K$ times, where for each pass we acquire a predicted mean and variance. These are used to apply the same noise sampling procedure implemented during training to produce $\bar{\textbf{p}}_k$.

\begin{algorithm}
\label{alg:eval_class}
\caption{Monte Carlo Dropout Classification Evaluation Loop}
\begin{algorithmic}
\FOR{each input}
\FOR{$k=1$ to $K$}
\STATE Acquire the predicted mean $f_{i,k}$ and variance $\sigma^2_{i,k}$ for each logit $i$
\FOR{$t = 1$ to $T$}
\FOR{each logit $i$}
\STATE Sample $\epsilon_{i,k,t} \sim \mathcal{N}(0,1)$
\STATE Compute sampled logit $x_{i,k,t} = f_{i,k} + \sigma_{i,k}^2\epsilon_{i,k,t}$
\ENDFOR
\STATE Compute $\textbf{p}_{k,t} = \text{Softmax}(\textbf{x}_{k,t})$
\ENDFOR
\STATE Compute average $\bar{\textbf{p}}_{k} = \frac{1}{T}\sum_{t=1}^{T}\textbf{p}_{k,t}$
\ENDFOR
\STATE $H_{\text{ale}} = \frac{1}{K}\sum^K_{k=1}H(\bar{\textbf{p}}_{k})$
\STATE $H_{\text{total}} = H(\frac{1}{K}\sum^K_{k=1}\bar{\textbf{p}}_{k})$
\ENDFOR
\end{algorithmic}
\end{algorithm}

Uncertainty should be expressed in a way that captures the variability in the predictions produced by our model; this is achieved here using careful averaging of the distribution of outputs/entropies. The aleatoric uncertainty for each prediction can be expressed in terms of the entropy of $\bar{\textbf{p}}_k$ (i.e.\ the class probabilities).   To acquire an aggregate estimate of the aleatoric uncertainty alone, we can compute the average entropy over the predicted distributions:

\begin{equation}
    H_{ale} = \frac{1}{K}\sum^K_{k=1}H(\bar{\textbf{p}}_k),
    \label{eq:H_2}
\end{equation}

where the operator $H$ is the entropy $H(\textbf{p}) = \sum_{i=1}^{n}p_i\text{log}_{2}(p_i)$ where $n$ is the number of classes. The variability in the predictions due to model uncertainty is described by the variability of $\bar{\textbf{p}}_k$ over $K$ passes. To capture the variability introduced by the epistemic uncertainty, we can average over $K$, where a large variance in the predictions will result in a broadened distribution with a high entropy. Therefore, we compute: 
\begin{equation}
    H_{total} = H\left(\frac{1}{K}\sum^K_{k=1}\bar{\textbf{p}}_k\right).
    \label{eq:H_1}
\end{equation}

We note that $H_{ale}\le H_{total}$ due to the concavity of the entropy function.

\subsection{Improved Variational Online Newton}
IVON \cite{shen2024variational} is an optimiser that can be used to approximate variational learning (mentioned in Sec. \ref{sec:bayes_opt}) on larger neural networks, avoiding issues with memory consumption associated with the classical implementation. Its formulation is as follows. If both the variational distribution and prior placed over model parameters has a Gaussian form, then the updates for model parameters resemble the form of generic optimisers (e.g.\ stochastic gradient descent or Adam). The Variational Online Gauss-Newton is one form, where the Hessian is approximated with a Gauss-Newton estimate to avoid negative variances in a given update \cite{khan2018fast}. However, the Gauss-Newton estimate is expensive to implement, as per-sample squaring is not a standard operation in deep learning. With IVON, the Hessian estimator is instead performed with a Newton-like update which is more efficient and remains valid for losses that are not twice differentiable.

The training and evaluation loops for classification and regression are similar to those used for MCD, with the exceptions that: i) during training, each batch is evaluated several times, each time using a randomly sampled model parameterisation (indexed by $j$ in Algorithm 4) where the gradients from each evaluation are aggregated and then used to update the model, ii) for testing, dropout layers are not active, and iii) for testing, iterative dropout sampling is replaced with the model sampling procedure described in i). The training loop for classification and is shown in Algorithm 4. The evaluation loops for IVON are the same as that used for MCD (Algorithm 3), where $K$ dropout samples are replaced with $J$ model evaluations. For regression tasks, aleatoric uncertainty can be estimated using a likelihood-based loss function (e.g.\ Gaussian negative log-likelihood). Consequently, uncertainty disentanglement can be performed following the same procedure as that used for MCD. 

The initialisation of the Hessian ($h_0$) controls the noise at initialisation, where larger values correspond to a more concentrated/deterministic initial posterior. Higher values may stabilise training at the cost of performance. The number of model samples used during training also significantly affects performance, where larger numbers of samples typically confer performance benefits that saturate. Using just a single sampled model per batch, IVON has comparable efficiency to other optimisers. However, using larger numbers of sampled models can significantly slow training.

\begin{algorithm}
\caption{IVON Classification Training Loop}
\begin{algorithmic}
\label{alg:train_class_ivon}
\FOR{each batch}
\FOR{$j=1$ to $J$ model evaluations}
\STATE Compute the predicted means $f_{l,i}$ and variance $\sigma^2_{l,i}$ for each logit $i$ for each example $l$ in the batch using sampled parameters $\theta$
\FOR{$t = 1$ to $T$}
\FOR{each logit $i$}
\STATE Sample $\epsilon_{l,i,t} \sim \mathcal{N}(0,1)$
\STATE Compute sampled logit $x_{l,i,t} = f_{l,i} + \sigma_{l,i}^2\epsilon_{l,i,t}$
\ENDFOR
\STATE Compute $\textbf{p}_{l,t} = \text{Softmax}(\textbf{x}_{l,t})$
\ENDFOR
\STATE Compute average $\bar{\textbf{p}_l} = \frac{1}{T}\sum_{t=1}^{T}\textbf{p}_{l,t}$
\STATE Evaluate NLL loss using all elements of $\bar{\textbf{p}_l}$ in a batch. 
\ENDFOR
\STATE Aggregate gradients over $J$ evaluations, and then update the model
\ENDFOR
\end{algorithmic}
\end{algorithm}

\subsection{Validating predicted uncertainties}
Here, we describe the various metrics we used to evaluate the calibration quality of predicted uncertainties for all tasks, as well as uncertainty disentanglement for the BP regression task.

\subsubsection{Regression}
Several metrics have been proposed to validate the uncertainties predicted by a neural network. For example, \cite{kuleshov2018accurate} proposed a metric based on the use of confidence intervals. A model will predict a Gaussian distribution for each input; if the uncertainty (variance of this distribution) is well-calibrated, then the ground truth for a corresponding prediction should fall within the predicted distribution's X \% confidence interval X \% of the time, i.e.\ $\frac{1}{N}\sum_{n=1}^{N} \mathbb{I}\{y_n \leq F_{n}^{-1}(p)\} = p \text{ for all } p \in [0,1] $, where $F^{-1}$ is the quantile function of the predicted distribution evaluated at the confidence interval $p$, and $y_n$ is the corresponding ground truth. A calibration curve can be plotted showcasing the observed frequency at which the ground truths lies within each chosen confidence interval. The sum of the squared offsets for each point of this plot is from a line with a slope of one can be used as a scalar metric, which we call the coverage calibration error (CCE).

It has been reported that if the prediction error distribution is roughly Gaussian, then a perfect calibration curve may be produced by setting all of the uncertainties to be equal (i.e.\ the errors and uncertainty may have no correlation, but a perfect calibration may be reported) \cite{levi2022evaluating}. The questionable effectiveness of this metric has motivated the development of another metric reported in \cite{levi2022evaluating}, that assesses the difference between the predicted uncertainty and the actual offset between the prediction and ground truth; well-calibrated uncertainties should approximate the observed prediction error. They propose the use of the expected normalised calibration error (ENCE) as a summary metric, which aggregates the normalised absolute difference between the the mean of two quantities computed for $N$ uncertainty bins of equal population (each indexed with $j$, where each example in a bin is indexed with $t$), 
\begin{equation}
    \text{ENCE} = \frac{1}{N} \sum_{j=1}^{N} \frac{|\text{mVAR}(j) - \text{RMSE}(j)|}{\text{mVAR}(j)}, 
\end{equation}
where,
\begin{equation}
    \text{mVAR}(j) = \sqrt{\frac{1}{B_j}\sum_{t \in B_j}\sigma_{t}^2}, 
\end{equation}
and, 
\begin{equation}
    \text{RMSE}(j) = \sqrt{\frac{1}{B_j}\sum_{t \in B_j}(y_t - \hat{y}_t)^2},  
\end{equation}

where $y_t$ is ground truth, and $\hat{y}_t$ is the predicted mean, and $\sigma_t^2$ is the predicted variance. The value of $N$ is chosen based on the desired precision with which one would like to assess the calibration, while also ensuring each bin still has sufficient examples so that the most extreme bins may not be so sensitive to outliers. Unlike the count-based calibration curved proposed in \cite{kuleshov2018accurate}, this metric scales with the magnitude of the offset between the prediction errors and the predicted uncertainties, making it sensitive to outliers in the predicted uncertainties. In contrast to the more global coverage metrics which are computed over this whole test set, the ENCE enables an assessment of how the calibration of the predicted uncertainties varies over subsets of the predictions/data. This is known as conditional calibration, and provides a more nuanced picture of how calibration varies across distinct populations of samples \cite{pernot2023validation}.

In contrast to these global and conditional summary calibration metrics, it may be of interest to understand how the accuracy of the predicted uncertainties vary over subsets of the predictions/data or at the level of individual predictions. This is known as individual calibration \cite{pernot2023validation}. Individual calibration is of practical importance, given diagnostic decisions may be informed from single estimates as opposed to distributions of estimates. Bivariate histograms of prediction error vs.\ predicted uncertainty can be used to increase the locality with which calibration may be assessed.

\subsubsection{Classification}
For classification, we used the expected calibration error (ECE) \cite{guo2017calibration}, which is the weighted average of the absolute difference between the model's accuracy and confidence. This is achieved by assessing the calibration of the model predictions by binning the predicted confidences (the maximum class probability) in equal-width bins, and determining the difference between the fraction of correct predictions in the bin (accuracy) and the average of the confidences in the bin. The ECE calculates a weighted average of these local calibration errors across all bins, and is given by

\begin{equation}
    \text{ECE} = \sum_{m=1}^{M} \frac{|B_{m}|}{N} |\text{acc}(B_{m}) - \text{conf}(B_{m})|, 
\end{equation}

where $M$ is the chosen number of bins, $N$ the total number of predictions, $|B_{m}|$ the size of bin $m$, and $\text{acc}(B_{m})$ and $\text{conf}(B_{m})$ are the accuracy and average confidence within bin $m$ respectively. The ECE scalar lies within the interval [0, 1], and allows for comparison of calibration techniques for different models. The mean confidence and mean accuracy can be plotted against one another to construct a calibration curve, which can be useful for visualisation.

Taking inspiration from \cite{guo2017calibration}, an analogous uncertainty calibration curve was proposed in \cite{laves2020calibration} where the estimated uncertainty is binned into equal-width bins, like the ECE, and the average number of incorrect responses in each bin, as well as the average entropy in each bin, is calculated. These quantities are then plotted against one another. A linear uncertainty calibration curve with a slope of 0.5 would provide a highly interpretable metric (i.e.\ entropy would be an interpretable metric if it happens to be linearly related to accuracy), and we treat this as our target for assessing the quality of our calibrations. Similar to for the ENCE, the authors of~\cite{laves2020calibration} propose a summary metric called the uncertainty calibration error (UCE) which, for the binary classification case, may be written as
\begin{equation}
    \text{UCE} = \sum_{m=1}^{M}\frac{|B_m|}{n}\left|\text{err}(B_m) - \frac{\text{uncert}(B_m)}{2}\right| ,
\end{equation}
where $B_m$ are the predictions in a bin (indexed by $m$) parsed from all $n$ test examples, 
\begin{equation}
    \text{err}(B_m) = \frac{1}{B_m} \sum_{i \in B_m} 1(\hat{y}_i \neq y),
\end{equation}
and 
\begin{equation}
    \text{uncert}(B_m) = \frac{1}{B_m} \sum_{i \in B_m} H_{\text{total},i},
\end{equation}
where the metric is bounded within the interval [0,1].
While these calibration curves may be plotted considering the whole population of test examples, it may be of interest to assess calibration quality for subsets of the data, e.g.\ for each class. This notion is referred to as adaptivity. Here, we plot calibration curves over the whole population of test data, and for each class. Per-class calibration quality is important to assess in medical contexts given the minority class is often of the most clinical importance, and therefore the reliability of their corresponding uncertainties determines their practical utility in informing diagnosis.

\subsubsection{Validating uncertainty disentanglement}
While we employ generic methods for disentangling the uncertainties for our regression models, it has been reported that uncertainties estimated with MCD (or other methods) and subsequently disentangled with this approach do not represent distinct sources of uncertainty \cite{mucsanyi2024benchmarking}. This is assessed by observing whether estimated aleatoric and epistemic uncertainty are highly correlated. With that said, some degree of correlation is unavoidable given the non-uniform distribution of aleatoric uncertainties in a  training set (i.e.\ there are often lower numbers of high aleatoric inputs, meaning the corresponding epistemic uncertainty is also high).  We assess this here by calculating a Pearson's correlation coefficient between the two, and observing how high the correlation between them may be.

\section{Datasets}
\subsection{Atrial fibrillation}
The DeepBeat dataset is composed of a mixture of raw time-series collected at Stanford University from subjects undergoing elective stress tests/cardioversions and a separate publicly-available dataset parsed from the IEEE Signal Processing Cup 2015 \cite{Torres-Soto_Ashley_2020}. We used a custom split of the dataset \cite{moulaeifard2025machine}, ensuring there is no patient overlap and a balanced label distribution across training, validation, and test sets (see Table \ref{tab:DeepBeat}). Each example is composed of a raw time-series, preprocessed using low-pass, high-pass, and adaptive filters to remove artefacts, noise, and baseline wander. Each example has a duration of 25 seconds, with a sampling rate of 32 Hz. Table \ref{tab:DeepBeat} shows the population of the training, validation, and test splits.

\begin{table}[ht]
    \centering
    \caption{   DeepBeat dataset split}
    \scalebox{0.8}{ %
    \begin{tabular}{|l|c|c|c|c|}
        \hline
        \textbf{Set}       & \textbf{Patients} & \textbf{Total samples} & \textbf{AF samples} & \textbf{Non-AF samples}  \\ \hline
        Training              & 88                   & 106249                 & 40603               & 65646                            \\ \hline
        Validation         & 26                   & 15256                  & 5800                & 9456                               \\ \hline
        Test               & 24                   & 15377                  & 5797                & 9580                                \\ \hline
    \end{tabular}
    }
    \label{tab:DeepBeat}
\end{table}

\subsection{Blood pressure}
\begin{figure}
    \centering
    \includegraphics[width=0.9\columnwidth]{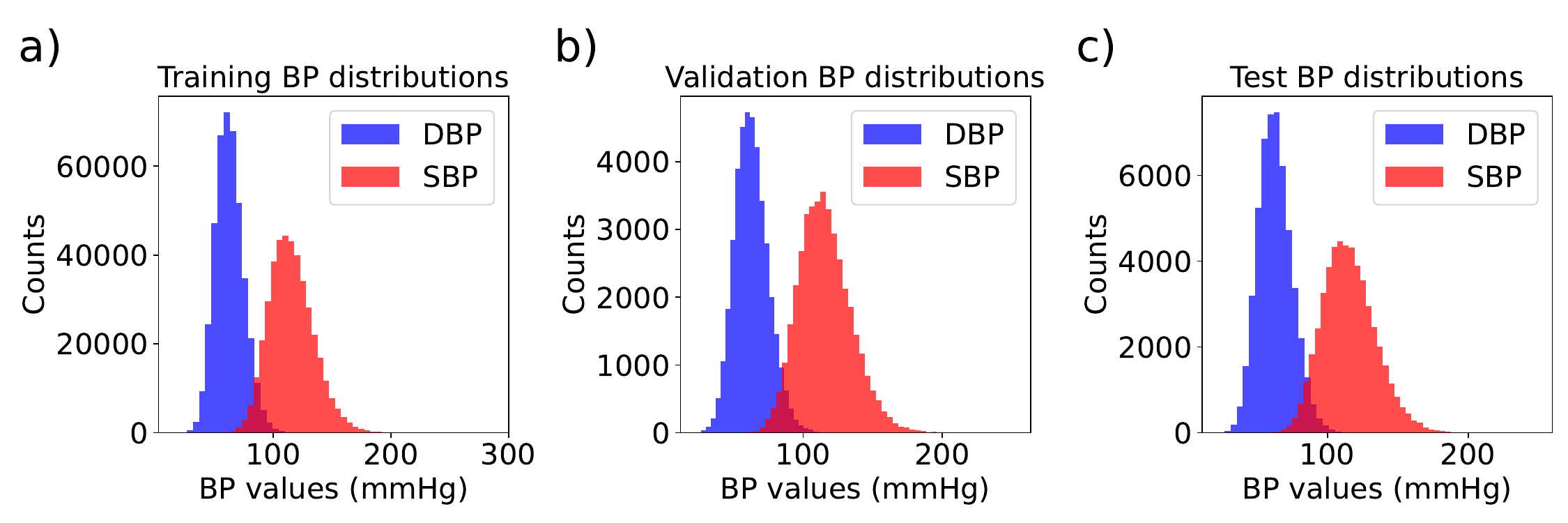}
    \caption{Distributions of ground truth blood pressure (BP) values in the a) training set, b) validation set, and c) test set.}
    \label{fig:bp_gts_hist}
\end{figure}

We used the PPG signals included in the VitalDB dataset, acquired from 6,090 ICU patients who underwent surgical procedures at Seoul National University Hospital for training the blood pressure regression models \cite{wang2023pulsedb}. We have 418,986 examples in the training set, 40,673 examples in the validation, and 51,720 in the test set (Fig. \ref{fig:bp_gts_hist} shows the distributions of ground truths in each set). Each example has a length of 1,250 elements. Data from the same patient is by design included in each split \cite{moulaeifard2025machine}, in order to approximate a scenario in which models are calibrated in a patient-specific way.

\begin{table}[h]
    \centering
    \renewcommand{\arraystretch}{1}
    \caption{Dataset statistics for VitalDB.}
    \begin{tabular}{l c}
        \toprule
        \textbf{Training (samples / subjects)} & 418986 / 1293 \\
        \textbf{Validation (samples / subjects)} & 40673 / 1293 \\
        \textbf{Test (samples / subjects)} & 51720 / 1293 \\
        \textbf{Gender (F \%)} & 42.31 \\
        \textbf{Age (mean ± SD)} & 58.98 ± 15.03 \\
        \textbf{SBP (mmHg, mean ± SD)} & 115.48 ± 18.92 \\
        \textbf{DBP (mmHg, mean ± SD)} & 62.92 ± 12.08 \\
        \bottomrule
    \end{tabular}
    
    \label{tab:vitaldb_calib}
\end{table}

\section{Learning to detect atrial fibrillation from raw PPG time-series}
Blood flow encodes information about the rhythm of the beating heart. PPG signals are sensitive to changes in blood volume, and consequently, can be used to infer pulse rate and rhythm from measurements at the vascular periphery. AF is a condition where, for short periods, the heart beats rapidly in an uncoordinated manner. The resultant effects on blood volume can be detected with PPG signals \cite{pereira2020photoplethysmography}, providing a means to monitor the ocurrence of AF episodes outside of the clinic with the use of wearable sensors. 

More specifically, PPG waveforms are known to have components corresponding to events in the cardiac cycle. For example, the initial positive slope of a PPG pulse emerges due to the contraction of the left ventricle, which will eject blood along the arterial tree and into the vasculature. The maximum of the waveform corresponds to the systolic peak and the dichrotic notch separating the systolic and diastolic phases (corresponding to the closing of the aortic valves) is the minimum following this peak \cite{pereira2020photoplethysmography}. While normal heart rhythm is indicated by little variation in the morphology of pulses, which tend to be spaced out at regular intervals, AF can be recognised by irregular spacing between pulses that exhibit significant variance in their morphology. Motion artefacts may cause similar changes in signal properties and may be easily misinterpreted as being caused by an arrhythmia.

While statistical/analytical prediction models utilise features hand-crafted  from the raw time-series as inputs (e.g.\ successive differences in peak-to-peak interval lengths, Shannon entropy, Poincar\'e plot analysis, and signal slope changes)  \cite{pereira2020photoplethysmography,chong2018motion,corino2017detection,tang2017identification}, deep networks have the ability to  automatically learn the most relevant features from the raw time-series themselves. This may confer significant performance benefits where hand-crafted feature extraction may not be robust to experimental artefacts or noise (deep networks implicitly take this into account), or where there is inadequate prior knowledge of the most relevant features. 

We aim to learn the optimal parameters $\theta$ of a deep neural network model $G_\theta$ that, when given a PPG time-series $x_1, x_2, ... x_i, ... x_N \in X$ as inputs, will output $\mu_{AF,i}$, $\mu_{NoAF,i}$, $\sigma_{AF,i}^2$ and $\sigma_{NoAF,i}^2$ used to parameterise two Gaussian distributions $\mathcal{N}_{AF}(\mu_{AF,i},\sigma_{AF,i}^2)$ and $\mathcal{N}_{NoAF}(\mu_{NoAF,i},\sigma_{NoAF,i}^2)$ that, when independently sampled (using a custom procedure represented here by $\textbf{f}_i \sim \mathcal{N}_{G_\theta}(x_i)$ and described in Algorithms 2) and 3), produce a vector of logits $\textbf{f}_i$ where the subsequent application of a Softmax should produce a vector of probabilities such that \text{argmax}($\mathbb{E}[\text{Softmax($\textbf{f}_i \sim \mathcal{N}_{G_\theta(x_i)}$)}]) = \text{argmax}(y_i)$, where $y_1, y_2, .... y_i, ... y_N  \in Y$ are the corresponding one-hot encoded classification labels. Furthermore, we would like the uncertainies estimated by the model to be well-calibrated according to the chosen evaluation metrics.

\subsection{Model and training}
We note some works implying that the use of batch norm \cite{mukhoti2020batch} and skip connections \cite{kendall2015bayesian} are compatible with MCD, and that there is some degree of flexibility with how one may apply dropout regularisation to an architecture. For example, one can optimise the accuracy of predictions/ease of training by carefully choosing which layers to regularise \cite{gal2015bayesian,kendall2015bayesian}. IVON, however, is known to suffer performance issues with models trained with batch norm \cite{duffield2024scalable}. 

Here, we trained a generic 1D convolutional network with four output nodes in the final layer for both MCD and IVON. We opted for a smaller network, as this reduces the computational expense of training the model with IVON using a large number of model evaluations, making it feasible to prototype parameterisations within a day of training. For MCD, we applied dropout regularisation liberally throughout the architecture, and used stochastic gradient descent for optimisation. Two nodes output the mean of each logit distribution, while the others output the corresponding variances. We used a batch size of 64, with an initial learning rate of 0.001, for MCD following the procedure described in Algorithm \ref{alg:train_class}, where we used $T=100$. For evaluation, we used $K=100$ dropout samples. For IVON, we used a batch size of 128, a learning rate of 0.02, and $J=100$ model evaluations for testing.

The parameters of the model that produced the lowest validation loss achieved over the course of training were chosen for evaluation. The validation loss was computed using the procedure outlined in Algorithm \ref{alg:train_class}. We trained identical models parameterised with a range of dropout rates for MCD. The same architecture was used for both uncertainty quantification techniques. For models optimised with IVON, we used 60 samples per batch for training. Other than this difference in the model sampling, the same evaluation loop used for MCD was also used for IVON. The IVON training loop is given in Algorithm 4. Here, we also trained several identical models, each parameterised by a different initialisation of the Hessian ($h_0$). For both uncertainty quantification techniques, we used weighted random sampling to account for class imbalance.

\section{Learning blood pressure estimation from raw PPG time-series}
Blood travels at higher velocities when the pressure on the walls of the vasculature system is high, resulting in short pulse transit times (the delay between the proximal and distal arterial waveforms) \cite{haddad2021continuous}. This can be used to infer information about blood pressure. Pulse arrival time is another metric that measures the interval between the instance of the heart beat (as detected by ECG, using the time associated with the R peak) and the resulting blood pulse's arrival at the periphery. Several features of the PPG waveform have been shown to correspond with this arrival time, allowing one to derive an idea of the blood pulse velocity and subsequently derive values of blood pressure \cite{sola2009parametric}. However, these approaches require multiple sensors and additional calibration, making them generally unappealing to implement and incompatable with wearable technologies.

Pulse wave analysis is another class of approaches that directly correlate PPG waveform features to blood pressure values. These can be calibration free (do not require a reference BP measurement), and may require only one sensor. For example, one can take the second derivative of a PPG to form an acceleration plethysmogram (APG), that contains several peaks whose amplitudes have been shown to correspond with various stages of the cardiac cycle, and to vary with BP \cite{haddad2021continuous,atomi2017cuffless}. However, the exact physiological mechanism by which blood pressure induces change in these features is not widely discussed \cite{mehta2023can,suzuki2008cuffless}. Arterial blood pressure (ABP) waveforms indicate the cyclic changes in arterial pressure that correspond with the beating heart (i.e.\ increases in pressure during systole when blood is being pushed into the vasculature, and decreases during diastole). Some studies have noted similarities in the morphology of the ABP waveform with the PPG waveform, and have conjectured that PPG signals may contain information related to BP; indeed several groups have found features that may correlate with BP \cite{li2021central, samimi2022cuffless, samimi2023ppg, ding2015continuous, khalid2018blood, gesche2012continuous, mase2011feasibility}. 

Similar to the AF classification task, we used raw PPG signals as an input. We aim to learn the optimal parameters $\theta$ of a deep neural network model $O_\theta$ that, when given a PPG time-series $x_1, x_2, ... x_i, ... x_N \in X$ as an input, will output a mean $\mu_{SBP,i}$, and variance $\sigma_{SBP,i}^2$ used to parameterise a Gaussian distribution $\mathcal{N}_{SBP}(\mu_{SBP,i},\sigma_{SBP,i}^2)$ that when sampled (represented here by $\textbf{f}_i \sim \mathcal{N}_{SBP}(x_i)$), produces an estimate of SBP, where ideally, $\mathbb{E}[\mathcal{N}_{SBP}(x_i)] \approx p_{sbp,i}$, where $p_{sbp,1}, p_{sbp,2}, .... p_{sbp,i}, ... p_{sbp,N} \in P_{sbp}$ are the corresponding ground truths. Simultaneously, we would like it to output the corresponding $\mu_{DBP,i}$, and variance $\sigma_{DBP,i}^2$ used to parameterise a Guassian distribution for DBP $\mathcal{N}_{DBP}(\mu_{DBP,i},\sigma_{DBP,i}^2)$, with the same properties. Furthermore, we would like $\sigma_{DBP,i}$ and $\sigma_{SBP,i}$ to have values that produce accurate estimates of predictive uncertainty when used with the evaluation procedure described in Algorithm 1. We train a single model to predict both SBP and DBP.

\subsection{Model and training}
Here, we learn blood pressure regression using a 1D ResNet-style architecture with 19 layers (8 residual blocks). While an extensive hyperparameter search/optimisation was not conducted, we found our chosen parameterisations were sufficient to produce results comparable to the existing literature \cite{wang2023pulsedb}. Here, two nodes output the predicted mean for each blood pressure parameter, and the other two predicted the corresponding variances. The outputs are used as inputs to two Gaussian negative log-likelihood losses, that are summed together to give the total loss. 

Each model was trained for 100 epochs, where the state used for evaluation was chosen based on that which produced the lowest validation loss. We studied how the chosen dropout rate affects uncertainty calibration by training identical models parameterised with different values. Here, we assessed calibration quality over the test set. We used $K=50$ dropout samples.

We did not train our ResNet-style regression models with IVON given the computational expense associated with using a larger number of model evaluations (which confer performance benefits), and challenges with choosing suitable hyperparameters, both of which complicated model prototyping. While in principle we could instead implement a smaller architecture like that used for the AF classification task to reduce computation time, we found that this architecture was not compatible with the use of MCD given challenges with optimising the model with dropout regularisation. This highlights one disadvantage of using MCD; it may only be used effectively for models that can tolerate dropout regularisation. Given our interest in comparing calibration quality of uncertainties derived using the two techniques for similar model architectures, we opted to not include the results of using a smaller model here.

\section{Results}
\subsection{Regression}

\begin{figure}
    \centering
    \LARGE{\sffamily BP Regression: MCD \\ Uncertainty Calibration}\\
    \includegraphics[width=.32\columnwidth]{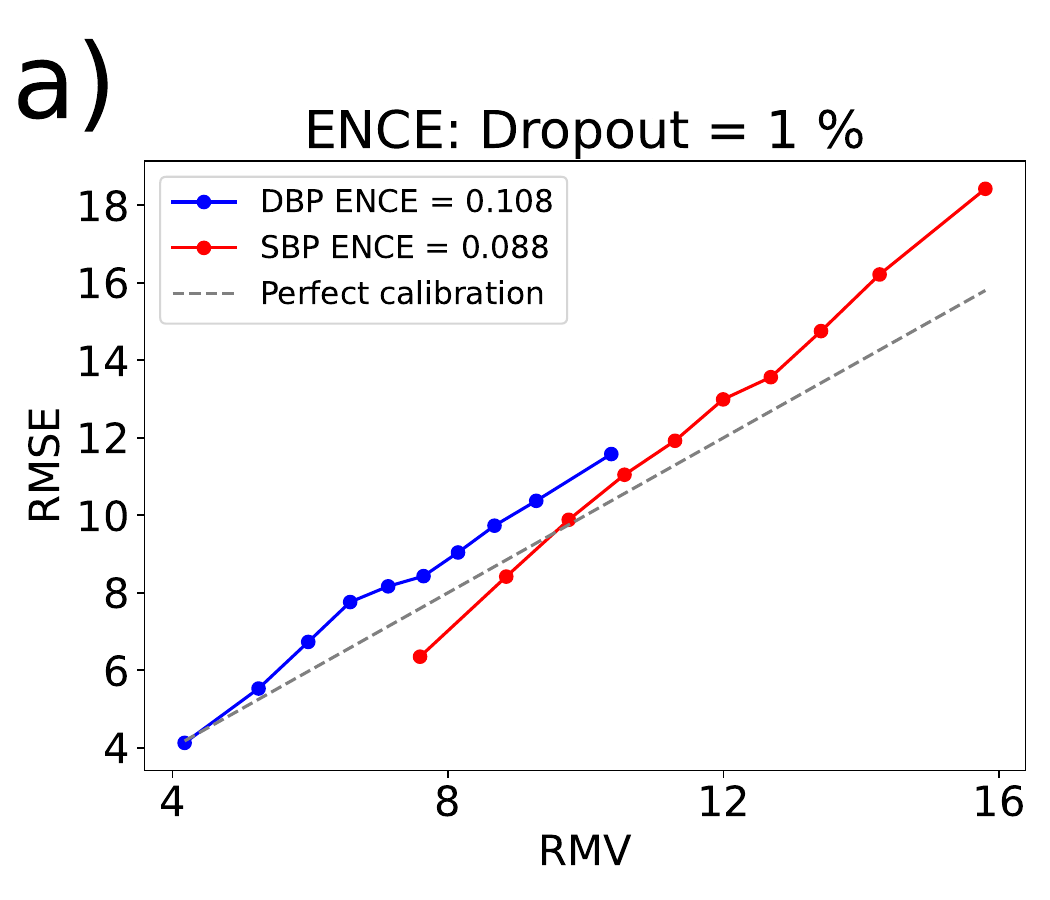}
    \includegraphics[width=.32\columnwidth]{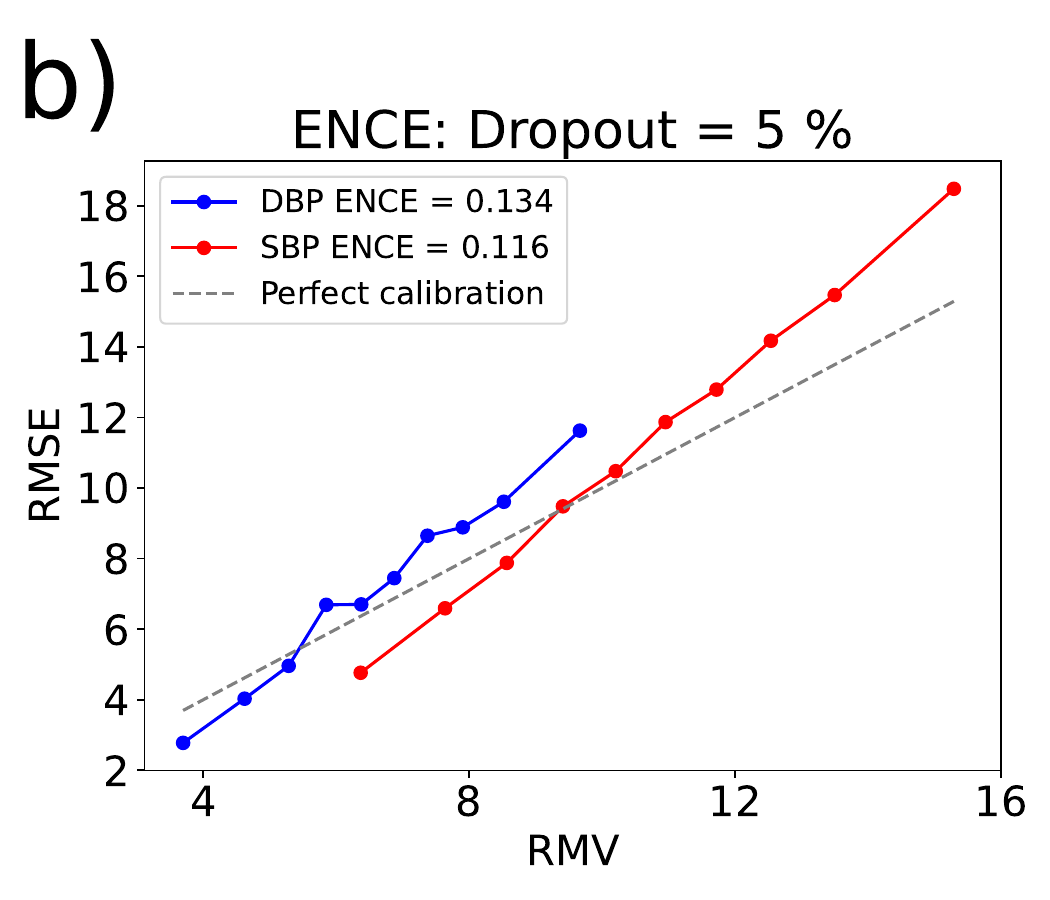}
    \includegraphics[width=.32\columnwidth]{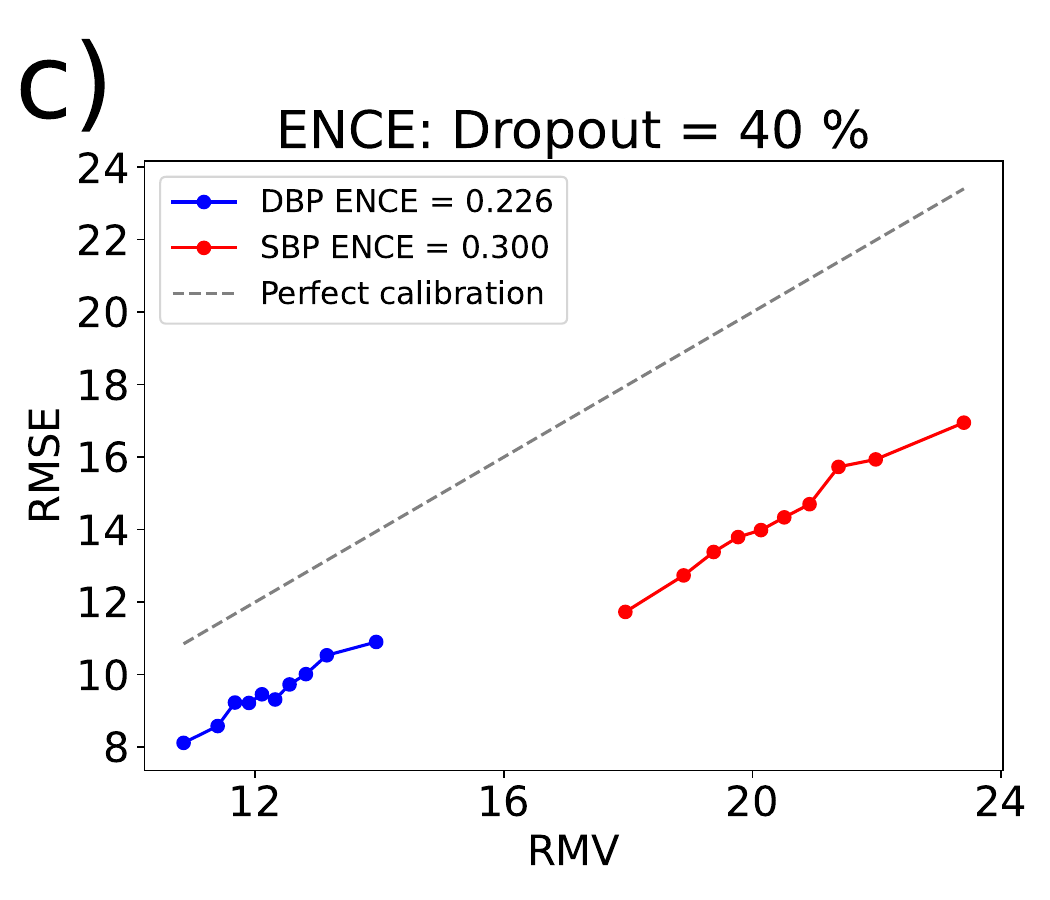}
    \includegraphics[width=.32\columnwidth]{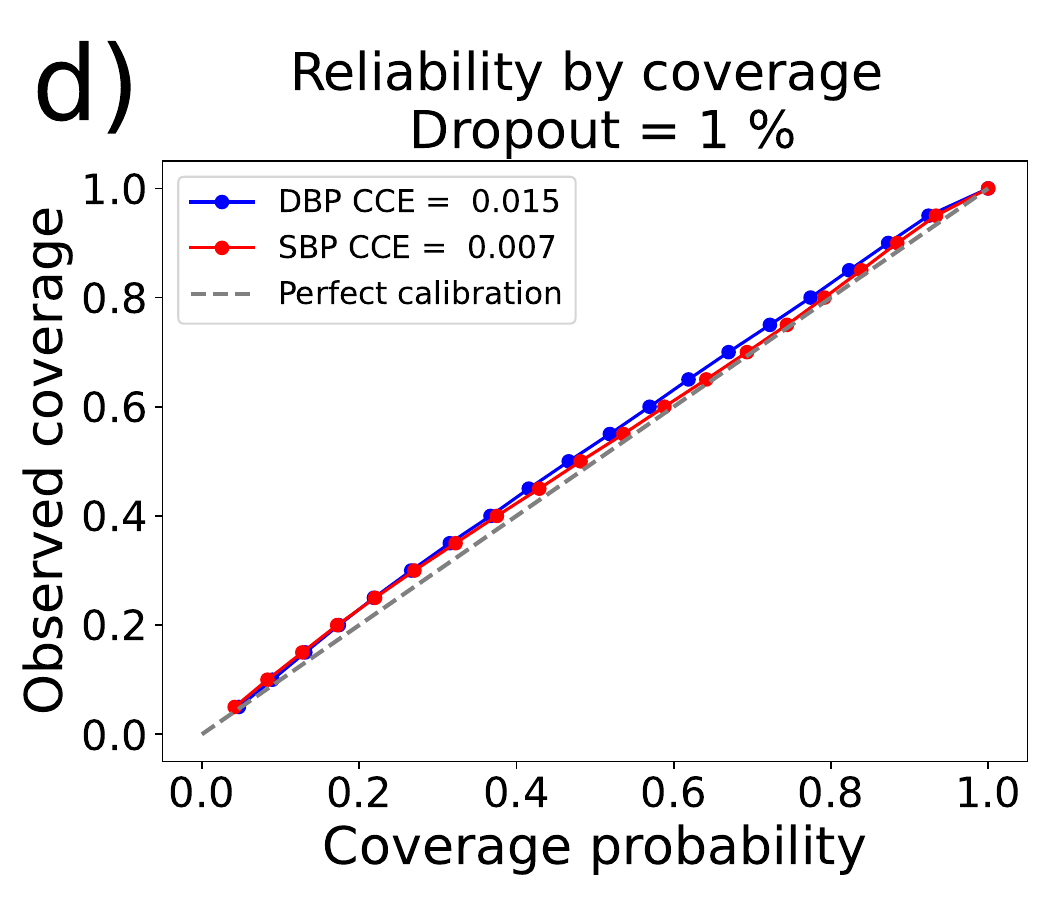}
    \includegraphics[width=.32\columnwidth]{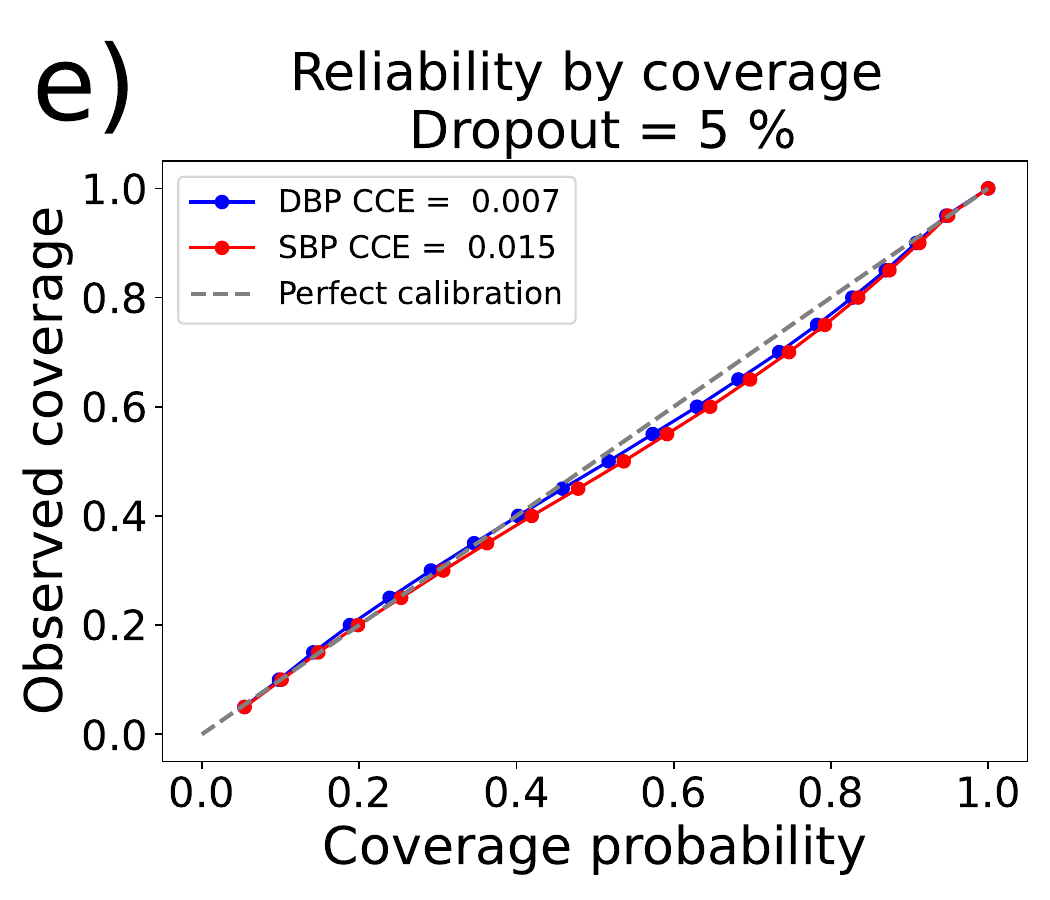}
    \includegraphics[width=.32\columnwidth]{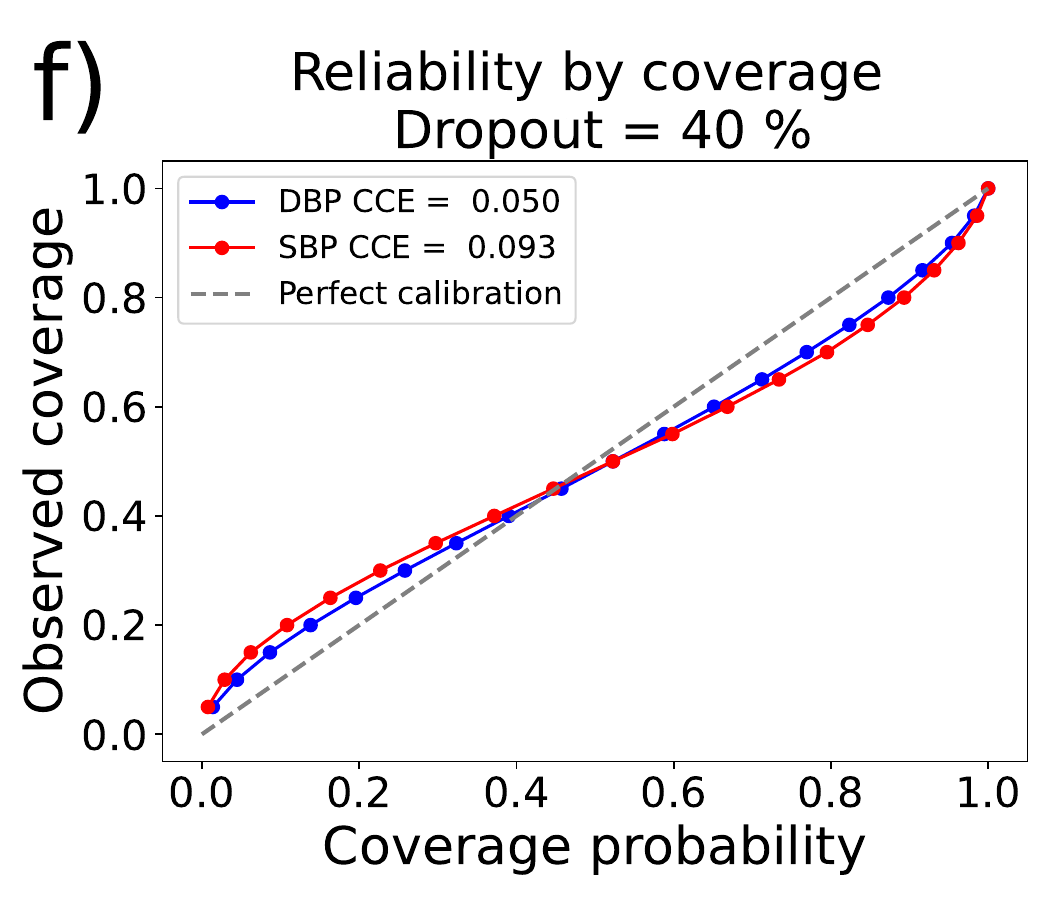}
    \includegraphics[width=.32\columnwidth]{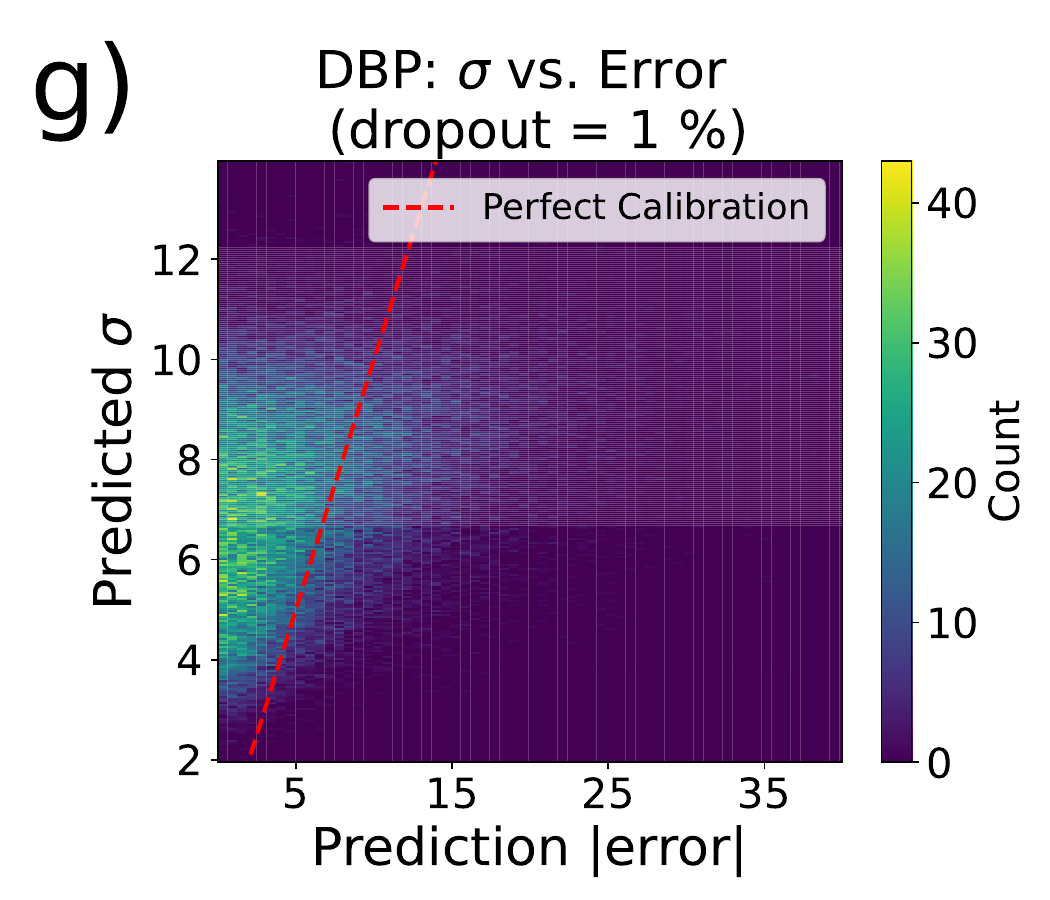}
    \includegraphics[width=.32\columnwidth]{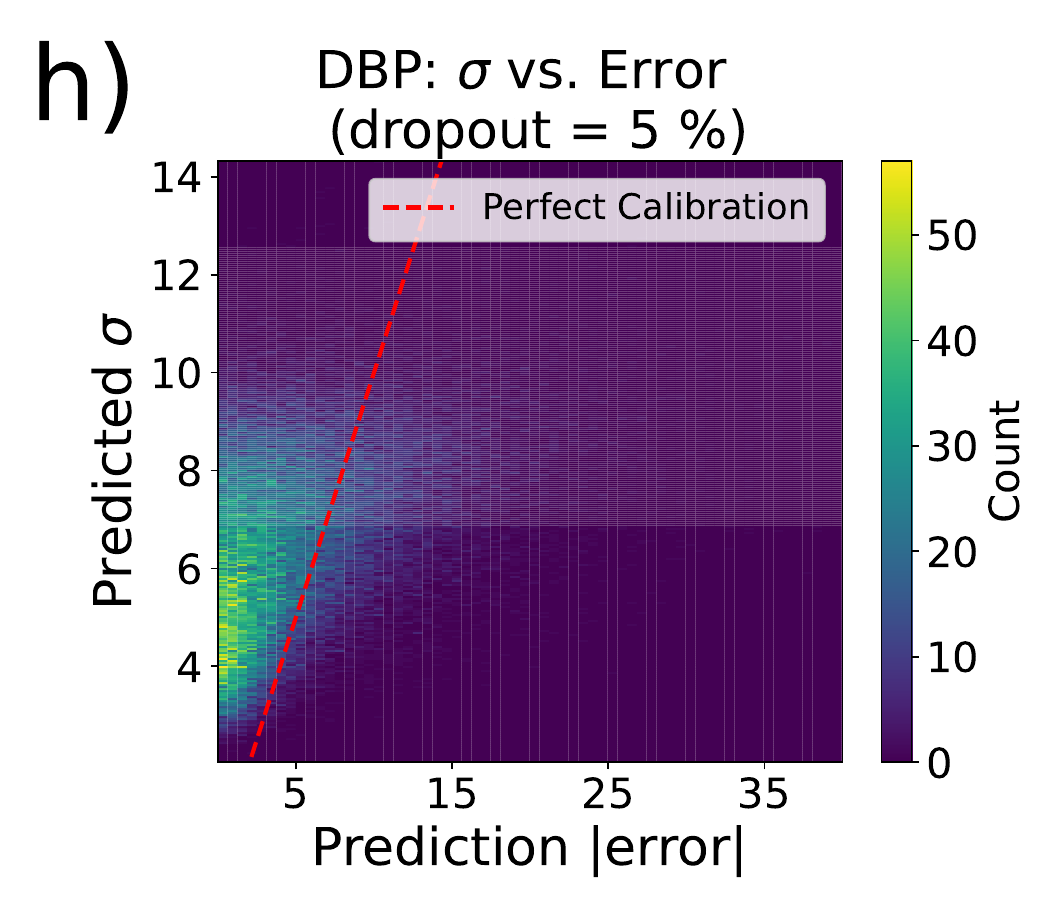}
    \includegraphics[width=.32\columnwidth]{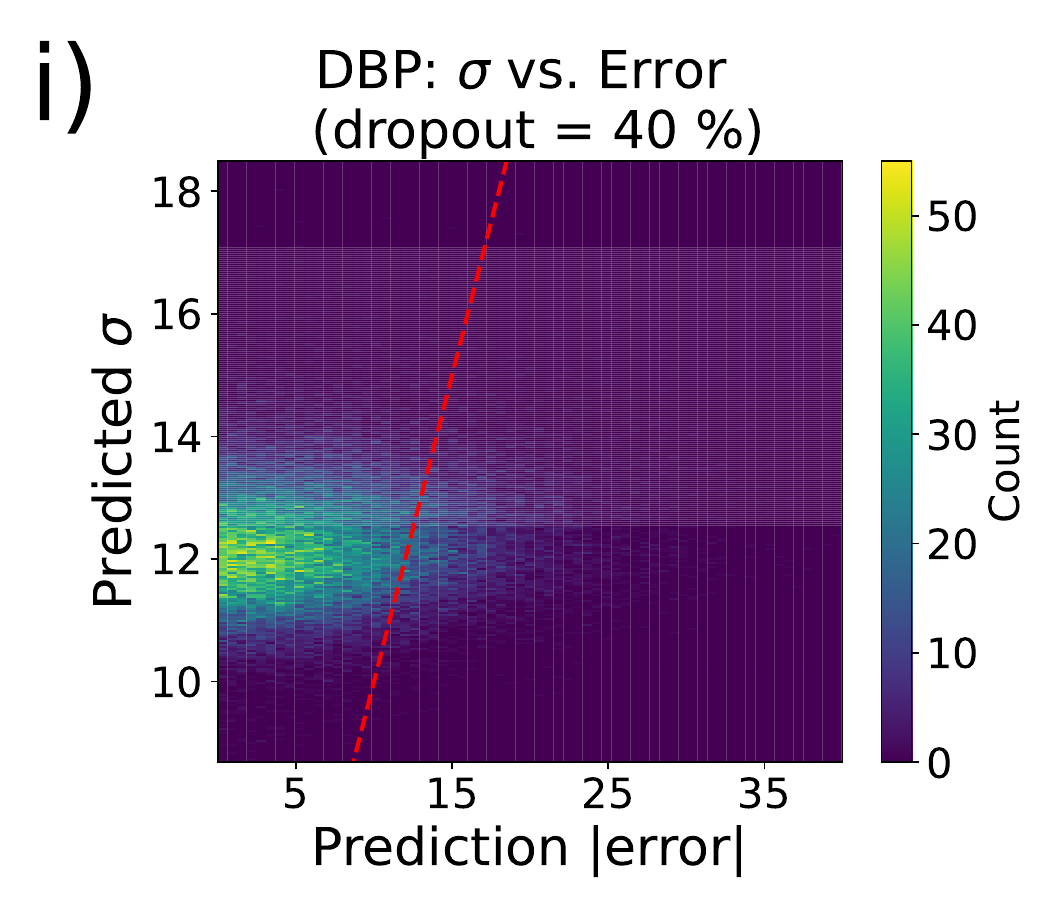}
    \includegraphics[width=.32\columnwidth]{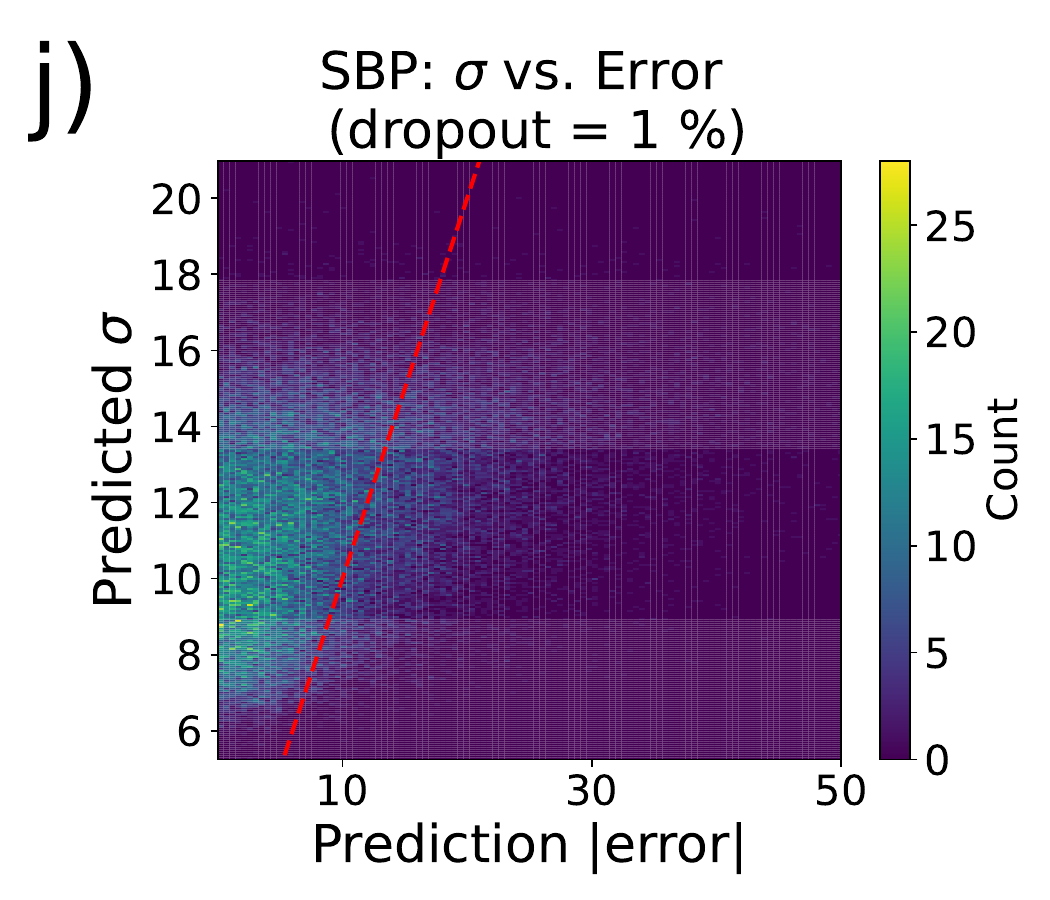}
    \includegraphics[width=.32\columnwidth]{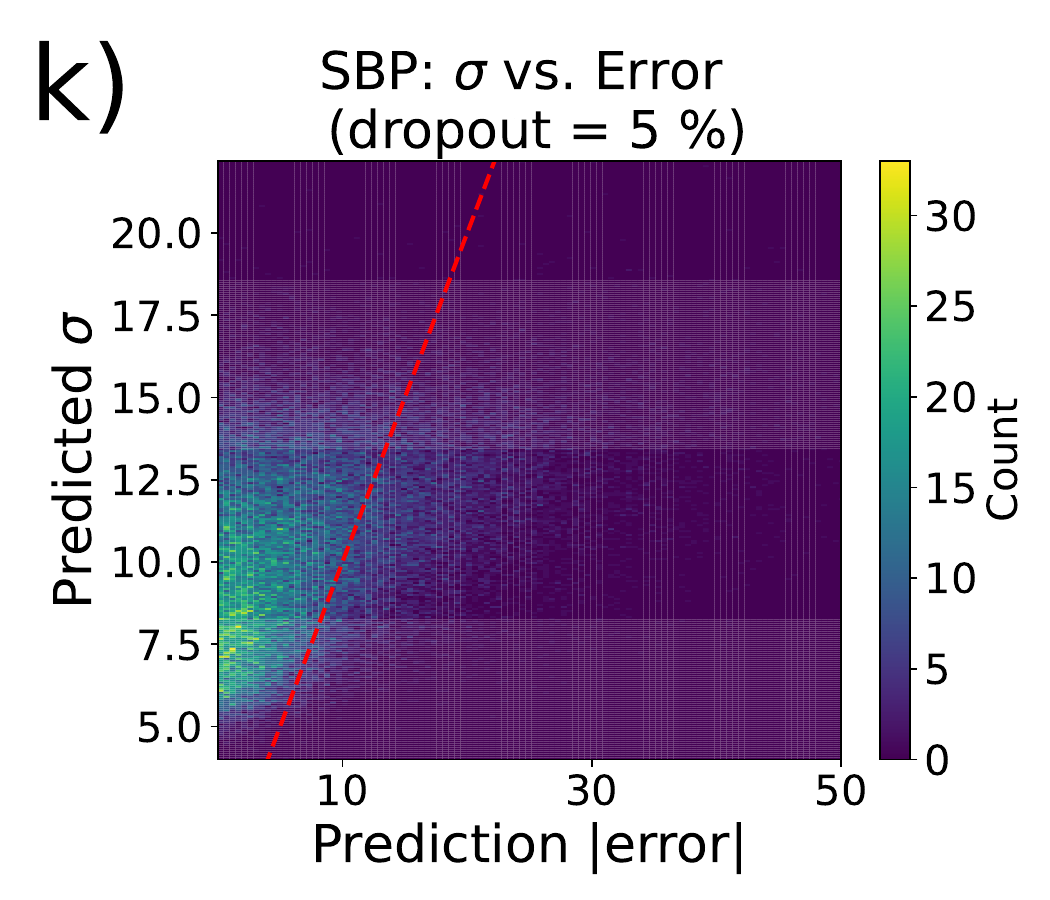}
    \includegraphics[width=.32\columnwidth]{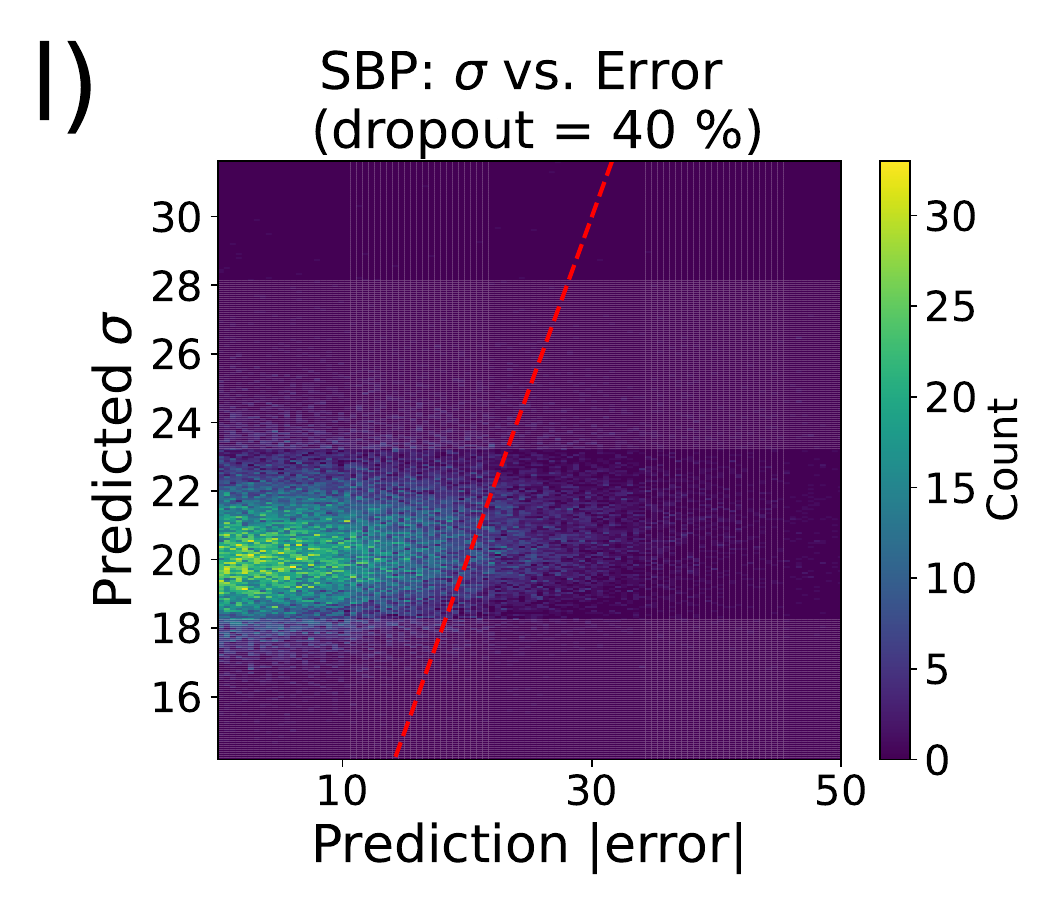}

    \caption{Evaluation of uncertainty calibration for blood pressure (BP) regression models trained with Monte Carlo Dropout (MCD). a-c) show the ENCE and corresponding reliability diagrams for both systolic blood pressure (SBP) and diastolic blood pressure (DBP), where the RMV is the root mean variance of the uncertainties in a given bin. d-f) show the coverage-based reliability diagrams and corresponding coverage calibration error (CCE) values. g-l) show how the predicted uncertainty is distributed against the prediction error (truncated along the horizontal axis to improve visualisation) for SBP and DBP.}
    \label{fig:BP_MCD_calib}
\end{figure}

\begin{figure}
    \centering
    \LARGE{\sffamily BP Regression: MCD \\ Uncertainty Disentanglement}\\
    \includegraphics[width=.32\columnwidth]{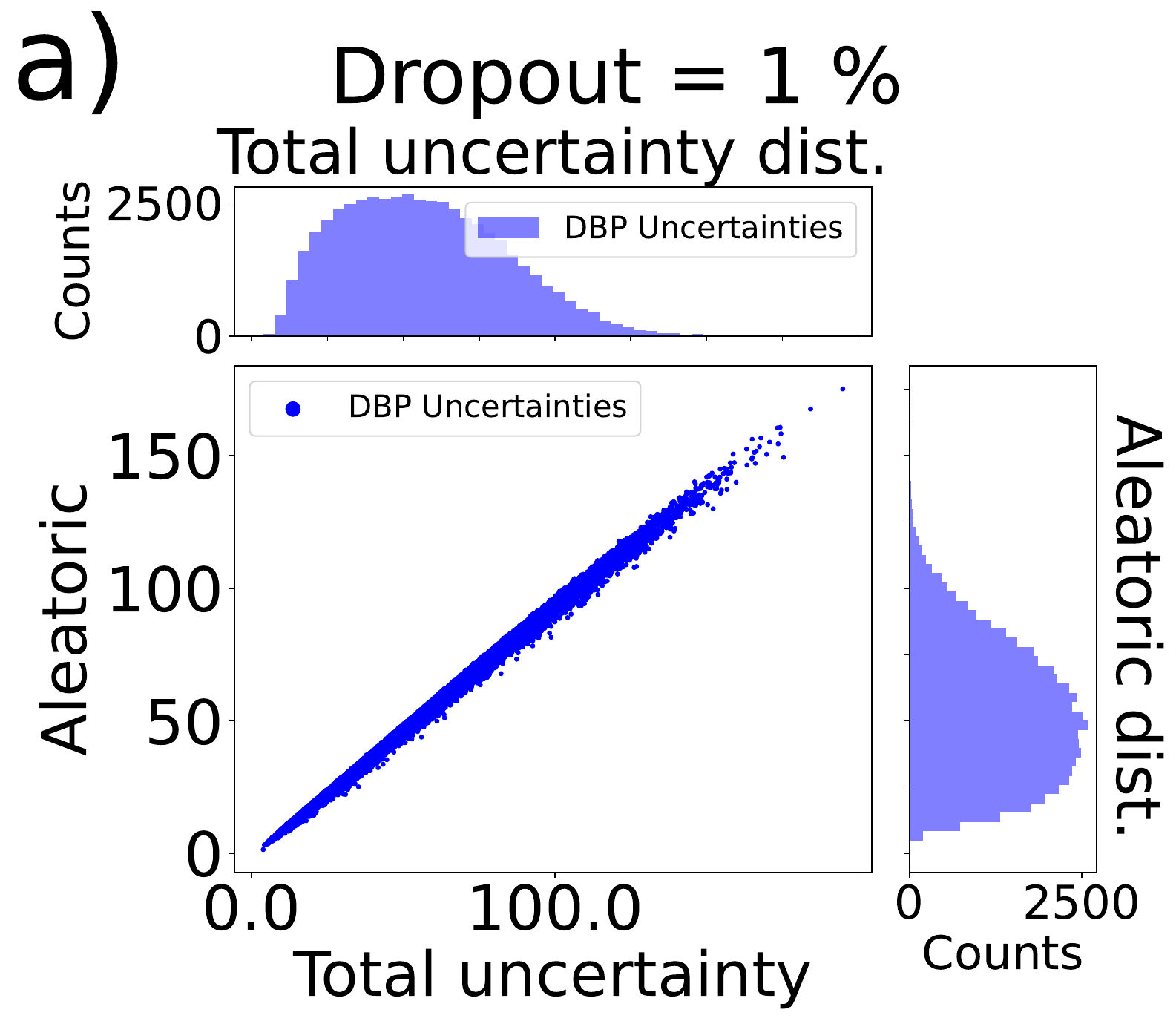}
    \includegraphics[width=.32\columnwidth]{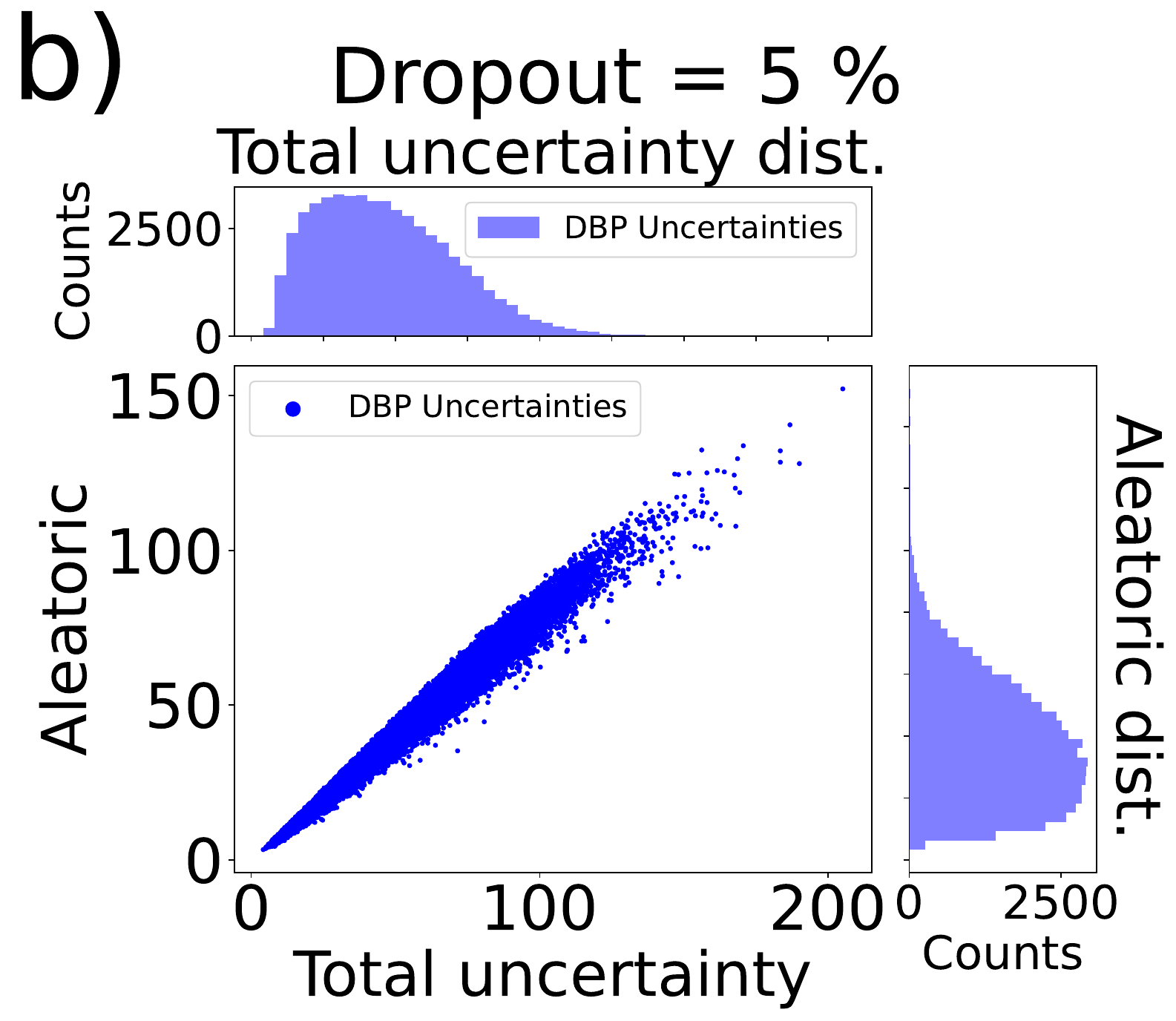}
    \includegraphics[width=.32\columnwidth]{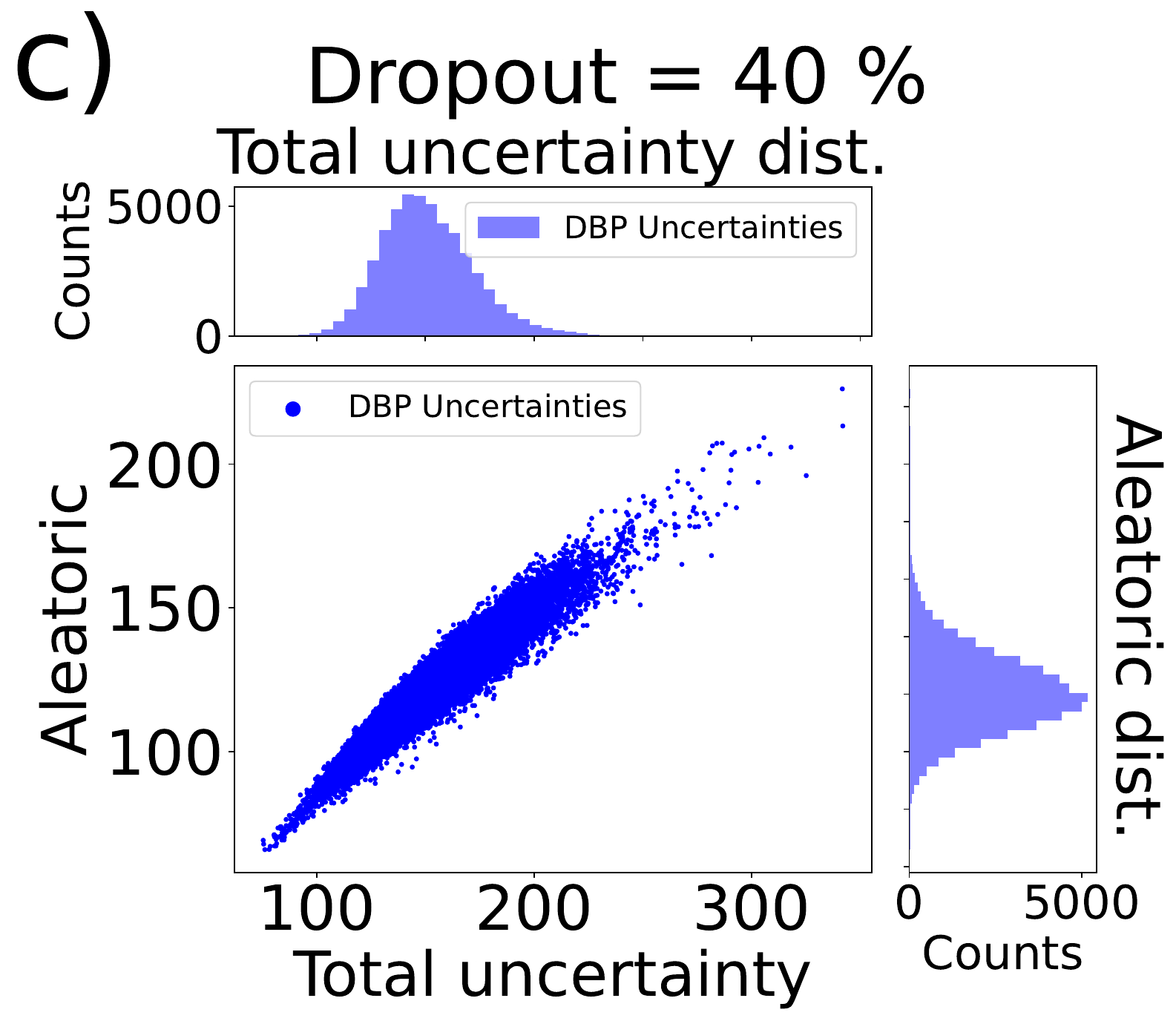}
    \includegraphics[width=.32\columnwidth]{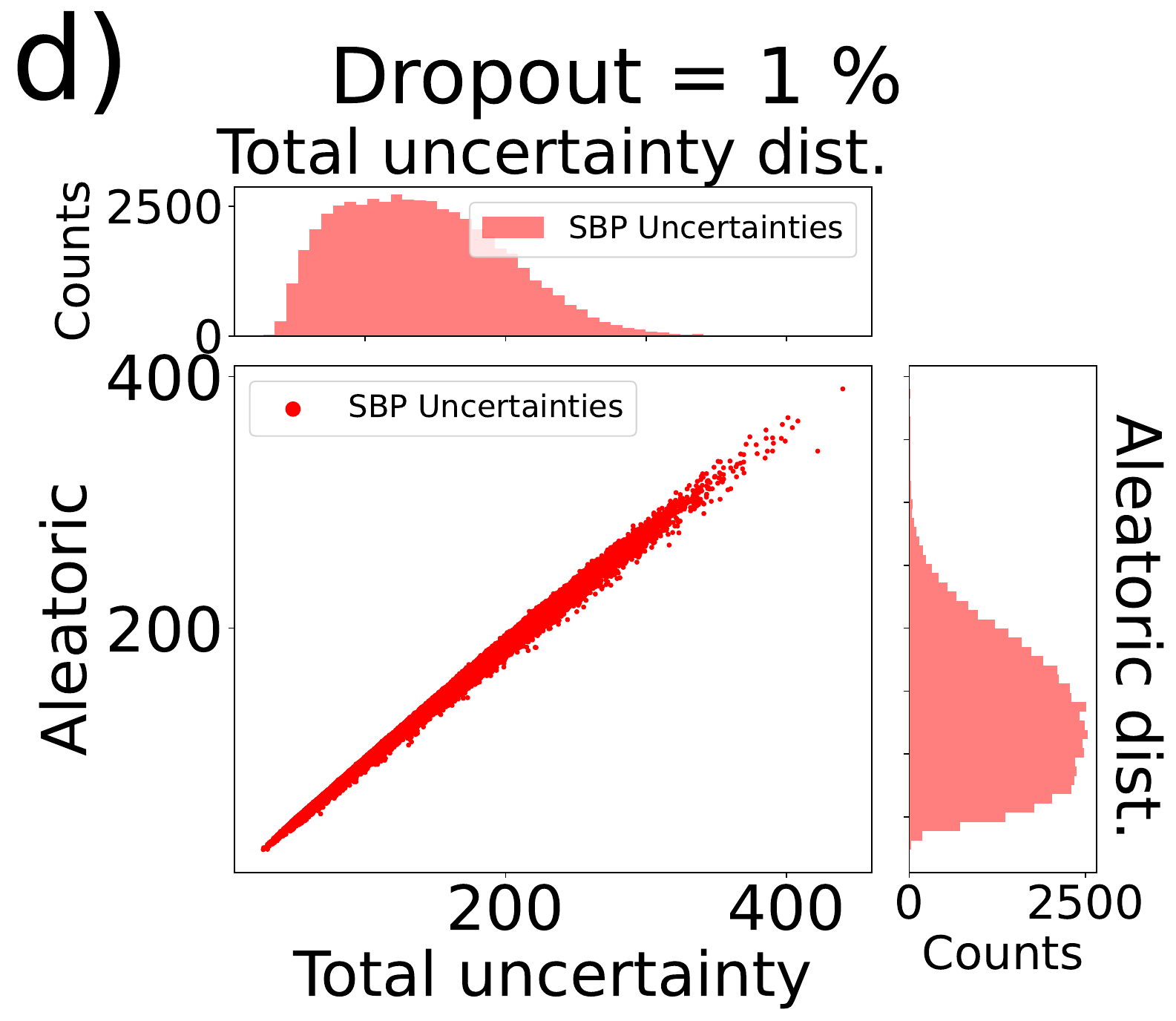}
    \includegraphics[width=.32\columnwidth]{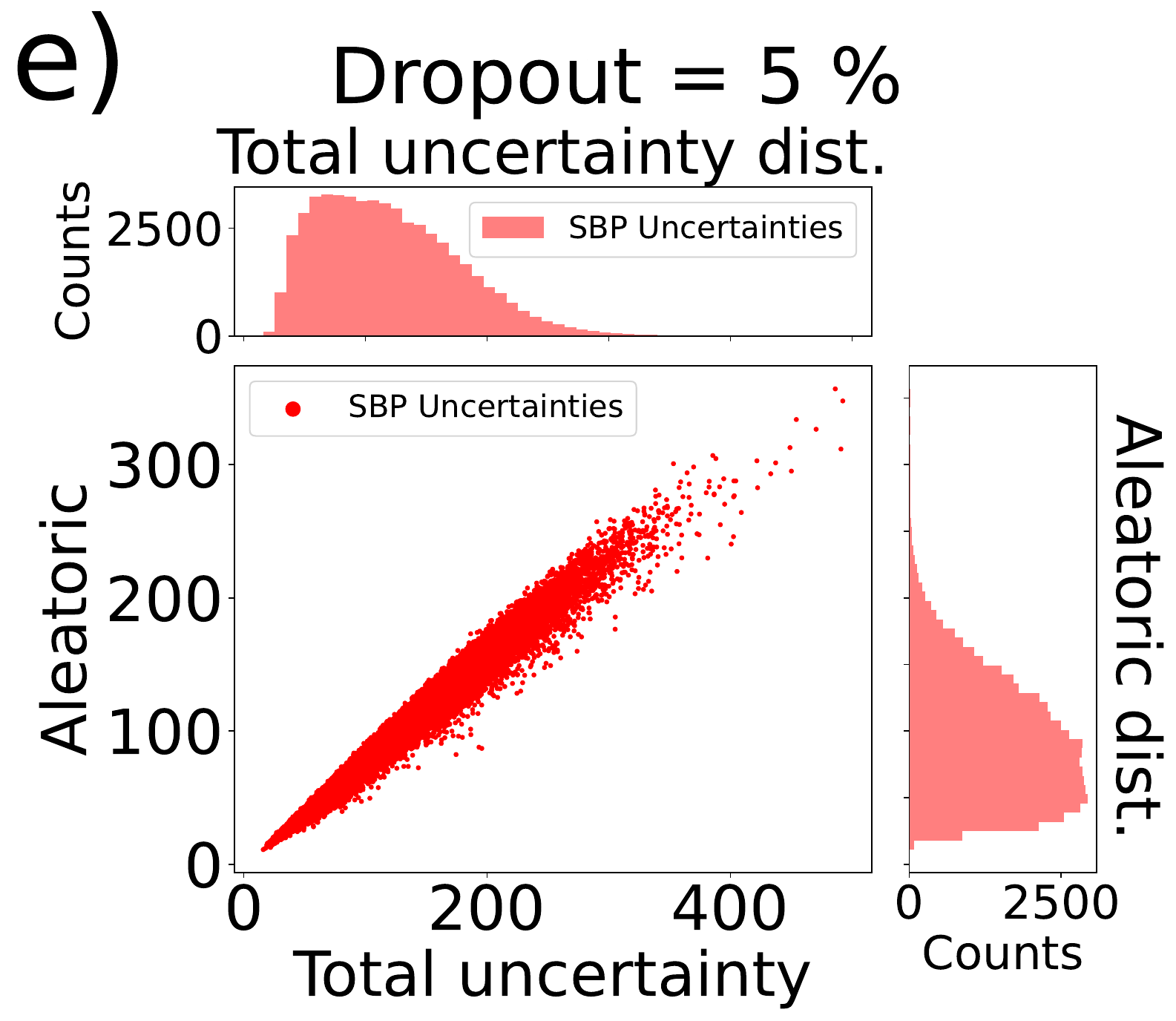}
    \includegraphics[width=.32\columnwidth]{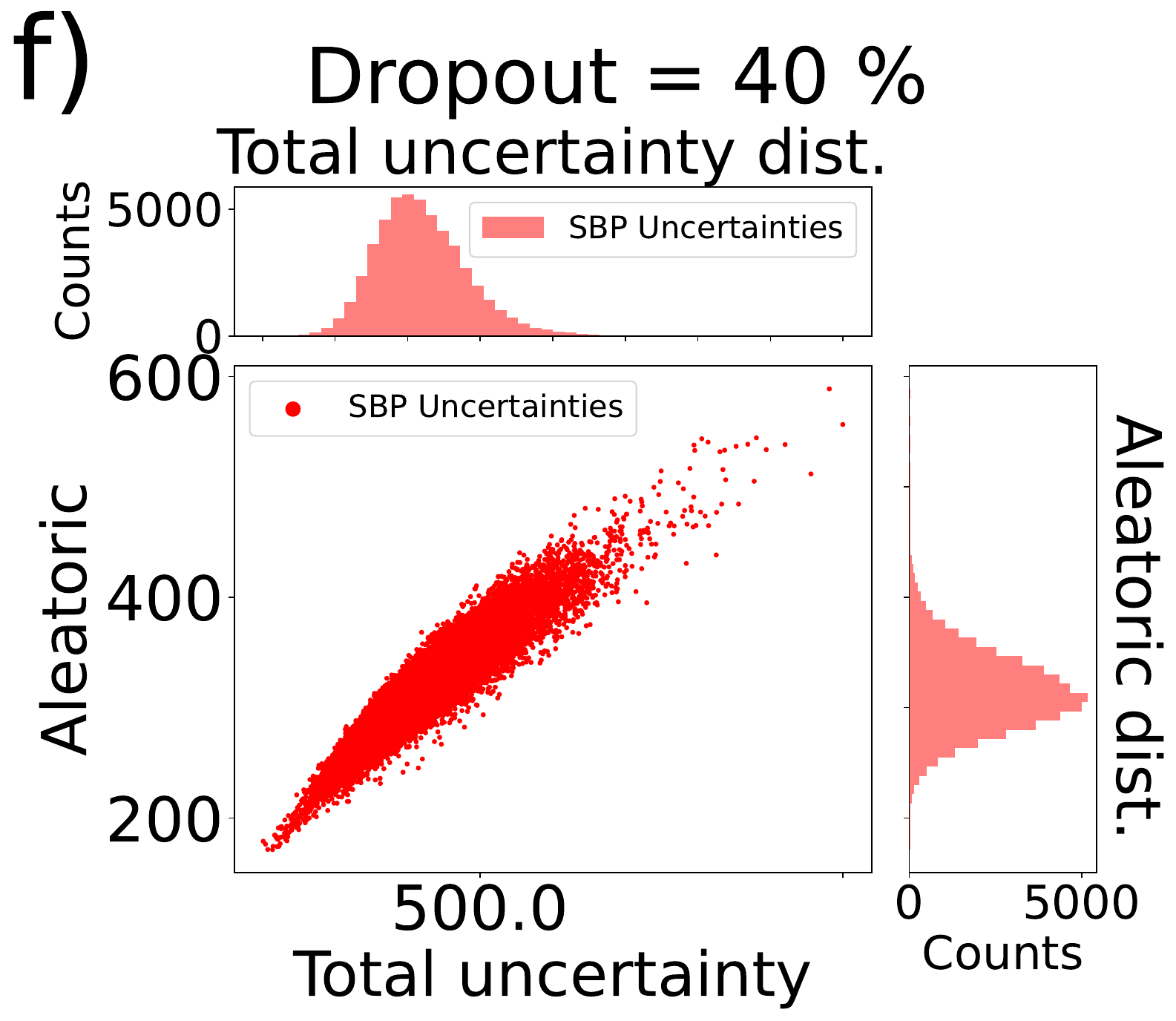}
    
    \caption{Uncertainty disentanglement for blood pressure (BP) regression models trained with Monte Carlo Dropout (MCD). a-c) show scatterplots of the aleatoric vs total uncertainty estimated for each test example expressed as variances, along with histograms showing the distributions of each value for diastolic blood pressure (DBP). d-f) show the corresponding plots for systolic blood pressure (SBP). We find that increasing the dropout rate increases how much of the total uncertainty is composed of epistemic uncertainty.}
    \label{fig:MCD_disentangle_reg}
\end{figure}

\begin{table}[h]
\centering
\begin{tabular}{|l|c|c|c|c|c|}
\hline
\multirow{2}{*}{\textbf{Dropout rate}} & \multirow{2}{*}{\textbf{MAE} (mmHg)} & \multicolumn{4}{c|}{\textbf{IEEE Grade}} \\
\cline{3-6}
& & \textbf{A} & \textbf{B} & \textbf{C} & \textbf{D} \\
\hline
1 \% & 6.22 & 52.6 \%& 7.4 \%& 6.2 \%& 33.8 \%\\
\hline
5 \% & 5.39 & 60.2 \%& 6.9 \%& 5.5 \%& 27.5 \%\\
\hline
40 \% & 7.47 & 41.2 \%& 7.2 \%& 6.7 \%& 44.9 \%\\
\hline
\end{tabular}
\caption{Predictive performance of blood pressure regression models trained with Monte Carlo Dropout for diastolic blood pressure. IEEE grades are defined in accordance with the standard 1708a-2019 \cite{ieee_standard}.}
\label{tab:model_comparison_dbp}
\end{table}

\begin{table}[h]
\centering

\begin{tabular}{|l|c|c|c|c|c|}
\hline
\multirow{2}{*}{\textbf{Dropout rate}} & \multirow{2}{*}{\textbf{MAE} (mmHg)} & \multicolumn{4}{c|}{\textbf{IEEE Grade}} \\
\cline{3-6}
& & \textbf{A} & \textbf{B} & \textbf{C} & \textbf{D} \\
\hline
1 \% & 9.51 & 37.3 \%& 6.4 \%& 5.8 \%& 50.5 \%\\
\hline
5 \% & 8.59 & 42.6 \%& 6.5 \%& 5.8 \%& 45.1 \%\\
\hline
40 \% & 11.20 & 29.0 \%& 5.4 \%& 5.2 \%& 60.5 \%\\
\hline
\end{tabular}
\caption{Predictive performance of BP regression models trained with Monte Carlo Dropout for systolic blood pressure. IEEE grades are defined in accordance with the standard 1708a-2019 \cite{ieee_standard}.}
\label{tab:model_comparison_sbp}
\end{table}

The best performing model trained with MCD (with a dropout rate of 5 \%) achieved a test mean absolute error (MAE) of 8.59 mmHg, and 5.39 mmHg for SBP and DBP respectively. We provide MAEs for the other models and IEEE grades in accordance with the standard 1708a-2019 \cite{ieee_standard} in Tables \ref{tab:model_comparison_dbp} and \ref{tab:model_comparison_sbp}.

The reliability diagrams and bivariate histograms shown in Fig. \ref{fig:BP_MCD_calib} indicate that calibration quality varies with the dropout rate, and that smaller dropout rates appear to output uncertainties that qualitatively exhibit a greater degree of small-scale/local calibration. With that said, the bivariate histograms suggest that the models likely overpredict uncertainty at the level of individual predictions; this is not as evident from the reliability diagrams, and highlights the importance of using several calibration metrics to assess model performance. For a  given population of predicted uncertainties binned by magnitude, there can be significant variance in the corresponding prediction error. Models with smaller dropout rates have greater predictive performance (see Tables \ref{tab:model_comparison_dbp} and \ref{tab:model_comparison_sbp}). We hypothesise that the degree of individual calibration is dependent on how well the model learns the predictive task. The scatter plots in Fig. \ref{fig:MCD_disentangle_reg} also show that the proportion of the total predicted uncertainty that is epistemic is larger for models parameterised with larger dropout rates.

\begin{table}[h]
\centering
\begin{tabular}{|l|c|c|}
\hline
\textbf{Dropout Rate} & \textbf{DBP r value} & \textbf{SBP r value} \\
\hline
1 \% & 0.75 & 0.79\\
\hline

5 \% & 0.81 & 0.81\\
\hline

40 \% &0.72 & 0.72\\
\hline
\end{tabular}
\caption{Pearson's correlation coefficient (r) for aleatoric vs epistemic uncertainty estimated from the blood pressure regression models, for both diastolic blood pressure (DBP) and systolic blood pressure (SBP).}
\label{tab:r_values}
\end{table}

A Pearson's correlation coefficient was computed comparing the aleatoric and epistemic uncertainties of each test example for each model. The values shown in Table \ref{tab:r_values} indicate that the aleatoric and epistemic uncertainty exhibit high correlation. While some correlation is unavoidable given aleatoric uncertainty may not be normally distributed within the training data (and hence, result in higher epistemic uncertainty for these estimates), disentangled uncertainties should ideally exhibit lower correlation  \cite{mucsanyi2024benchmarking}. These results suggest that our own estimates are likely not completely disentangled. This raises concerns about whether uncertainties disentangled using this generic decomposition can be faithfully interpreted as representing a single source of uncertainty. Caution should be exercised when using disentangled estimates to inform decisions related to model prototyping or dataset curation.

\subsection{Classification}

We assess the quality of the uncertainties predicted by our models conditioned over their magnitudes by plotting an uncertainty calibration curve and calibration curve for each model over the whole test set \cite{pernot2023calibration,pernot2023validation,laves2020calibration}. We also assess the adaptivity of the model with respect to each class by making a separate plot for non-AF and AF examples. We also show scatterplots of the aleatoric vs.\ total uncertainty for each test example, along with histograms showing the distributions of the estimated values. We compute the AUC-ROC score (AUC), F1 score and Matthew's Correlation Coefficient, each calculated with a decision threshold resulting in a sensitivity of 0.8 (MCC Sens) and specificity of 0.8 (MCC Spec), as well as the sensitivity (Sens) and specificity (Spec) at a threshold achieving 0.8 specificity and 0.8 sensitivity respectively.

\subsection{MCD}
\begin{figure}
    \centering
    \Large{\sffamily AF Classifier: MCD}\\
    \includegraphics[width=0.32\columnwidth]{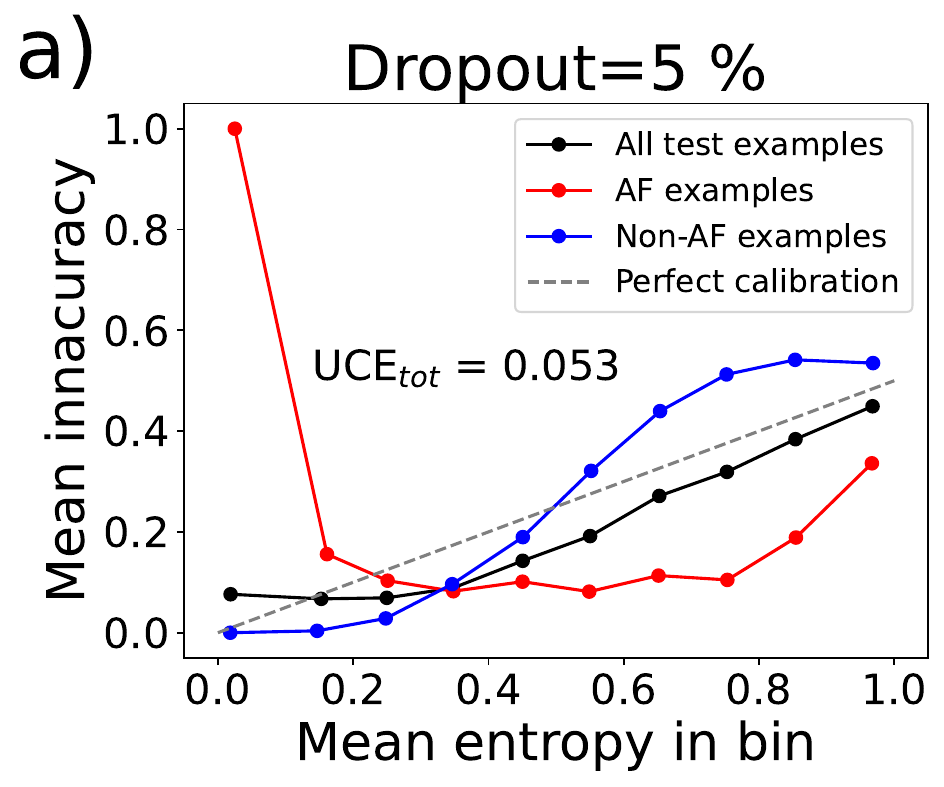}
    \includegraphics[width=0.32\columnwidth]{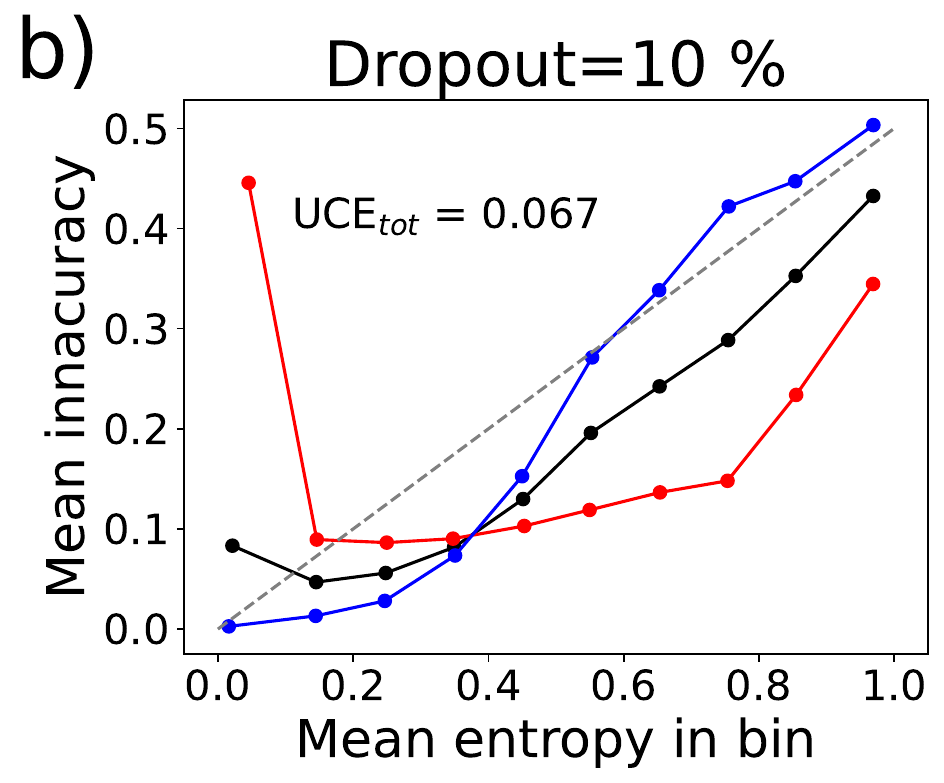}
    \includegraphics[width=0.32\columnwidth]{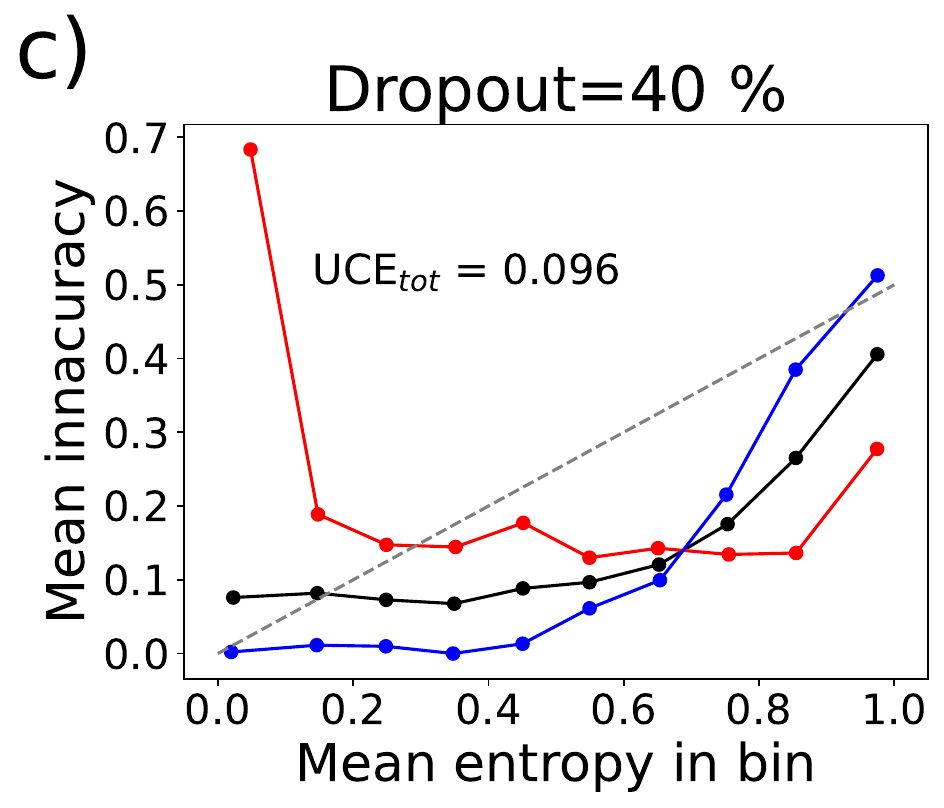}
    \includegraphics[width=0.32\columnwidth]{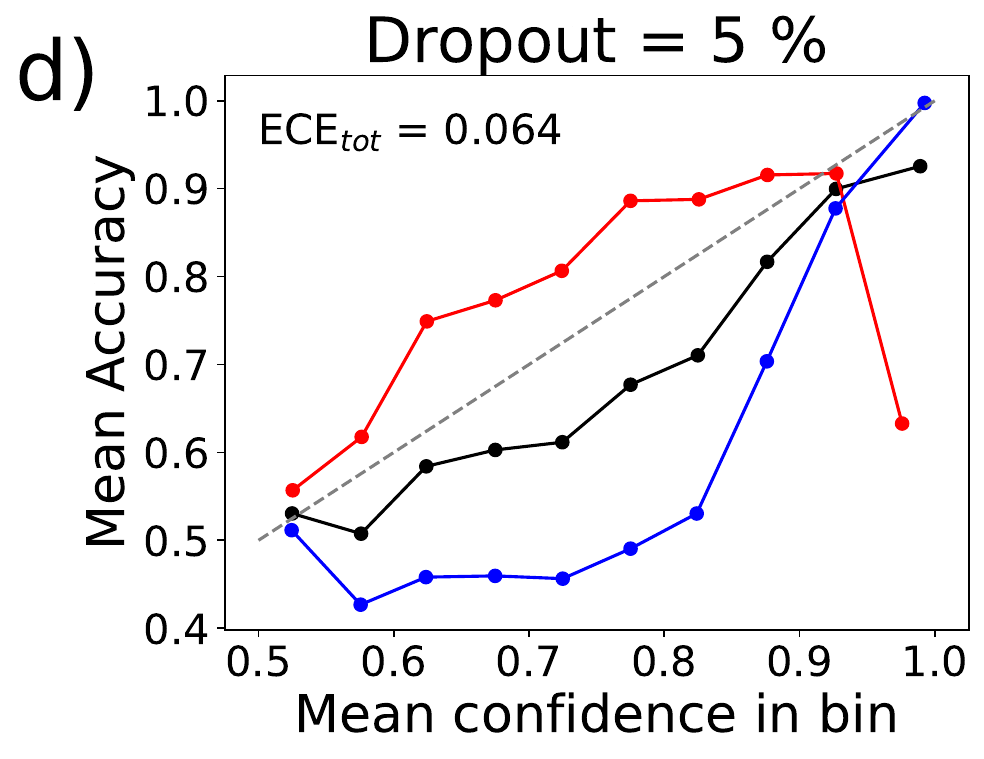}
    \includegraphics[width=0.32\columnwidth]{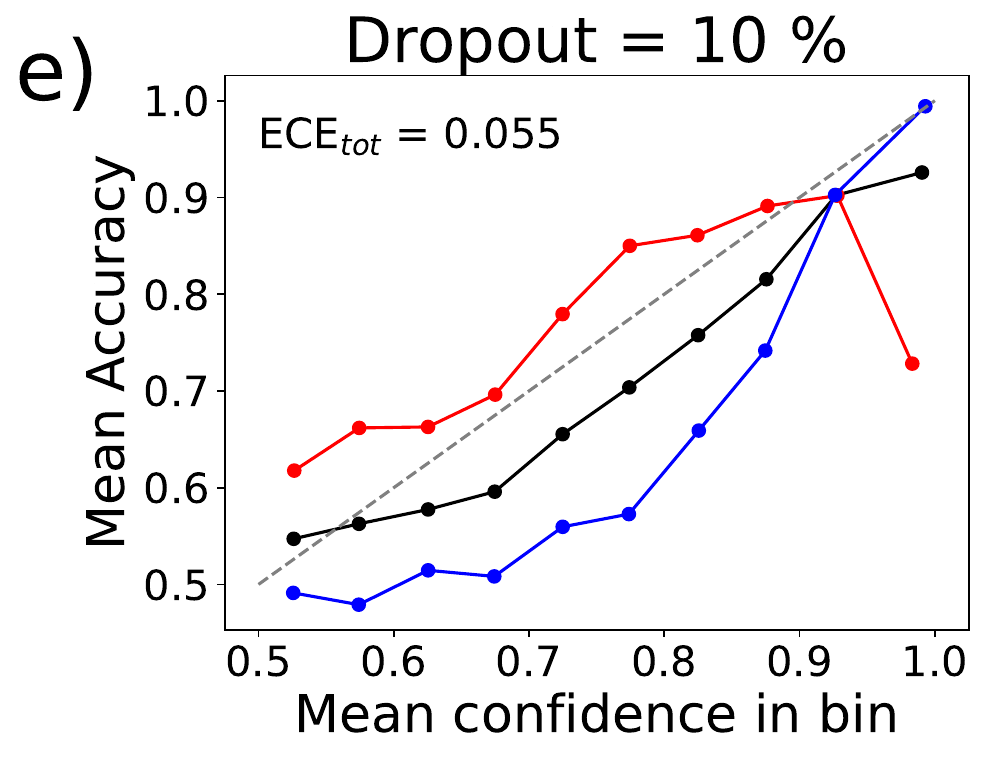}
    \includegraphics[width=0.32\columnwidth]{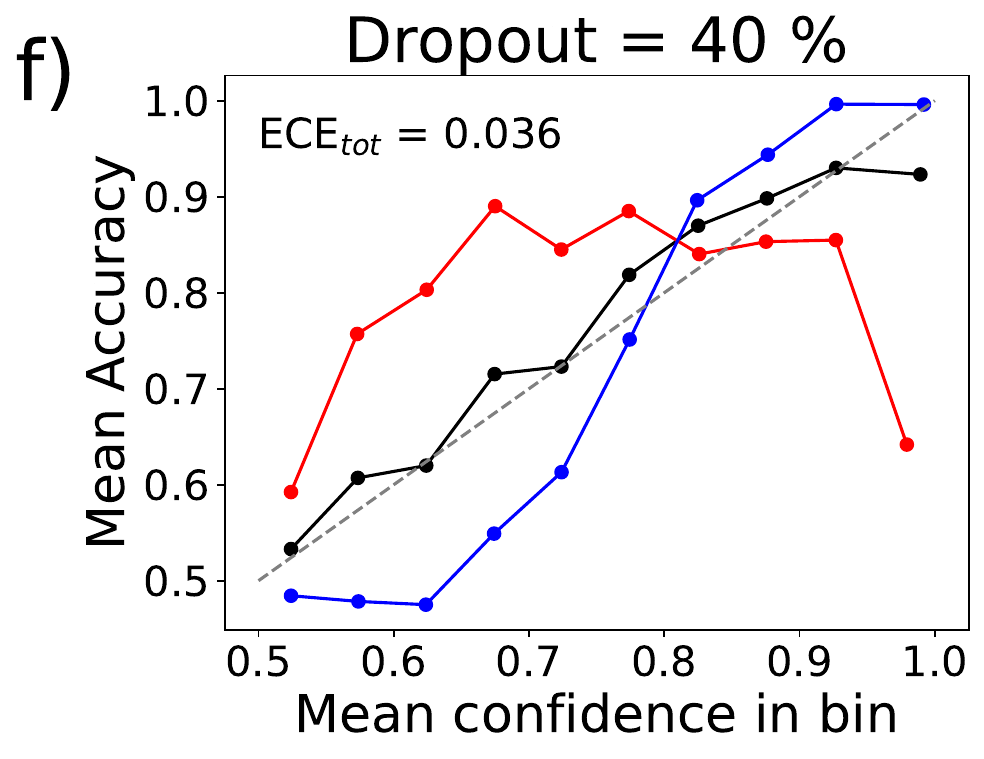}
    \includegraphics[width=.32\columnwidth]{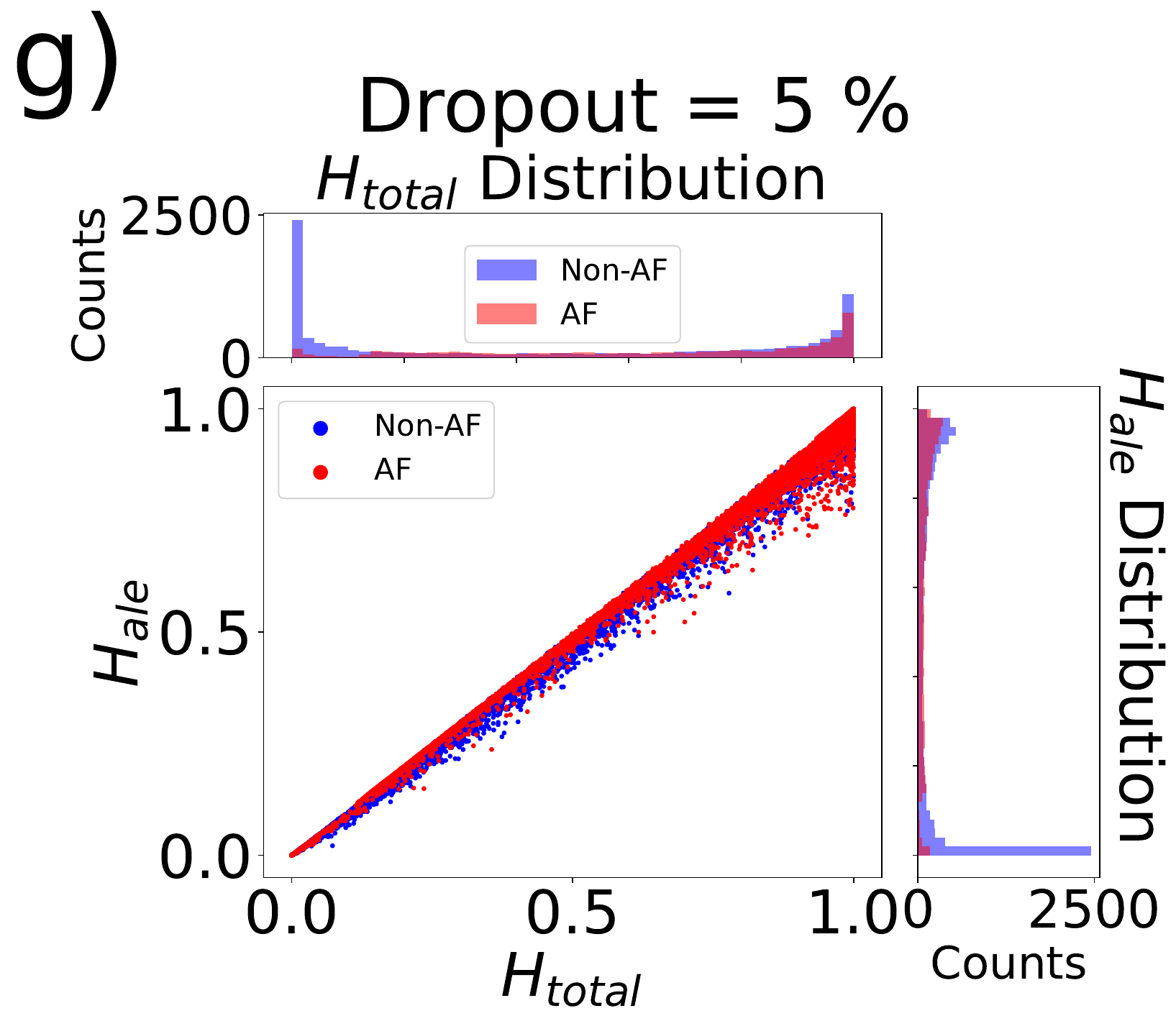}
    \includegraphics[width=.32\columnwidth]{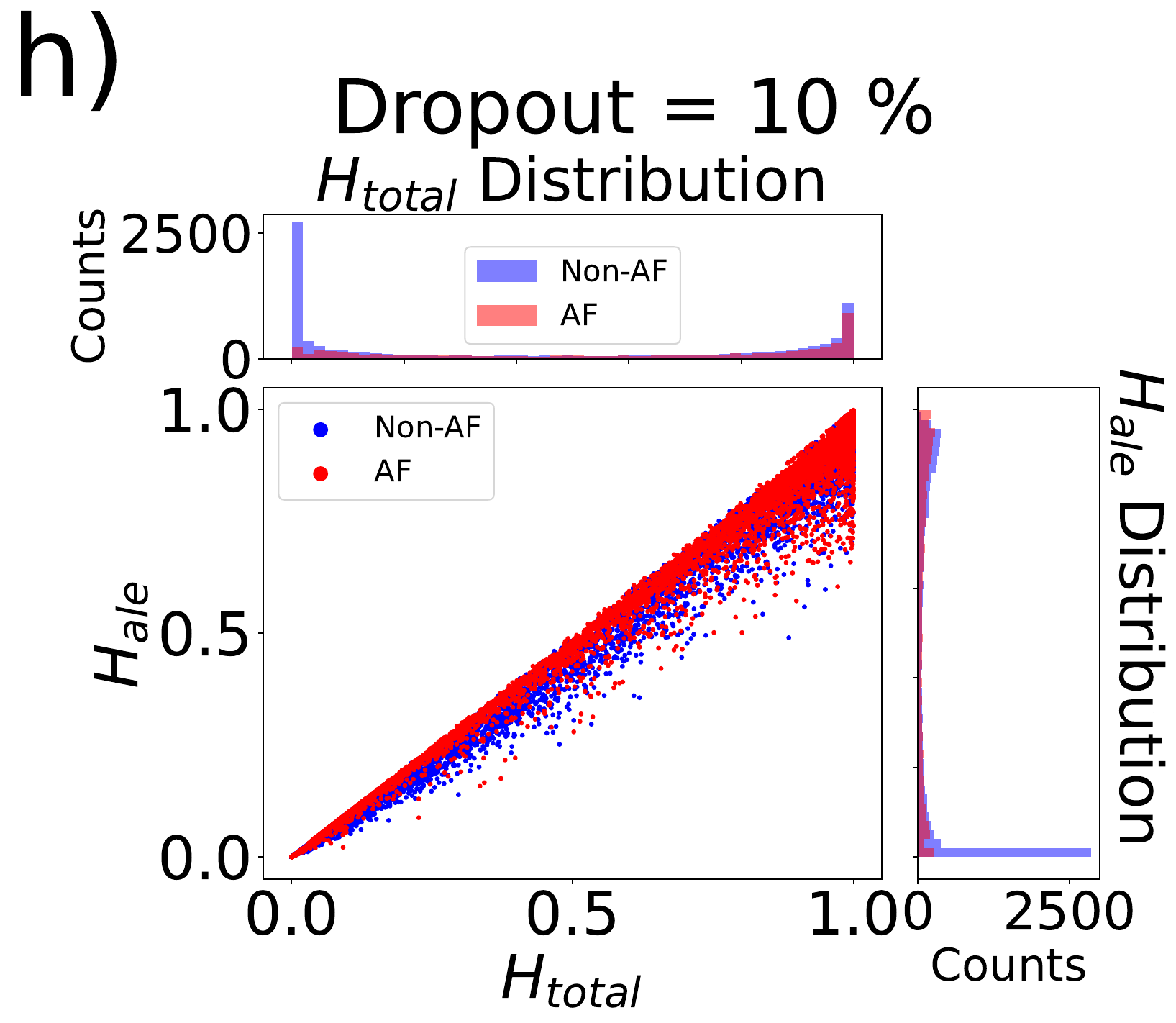}
    \includegraphics[width=.32\columnwidth]{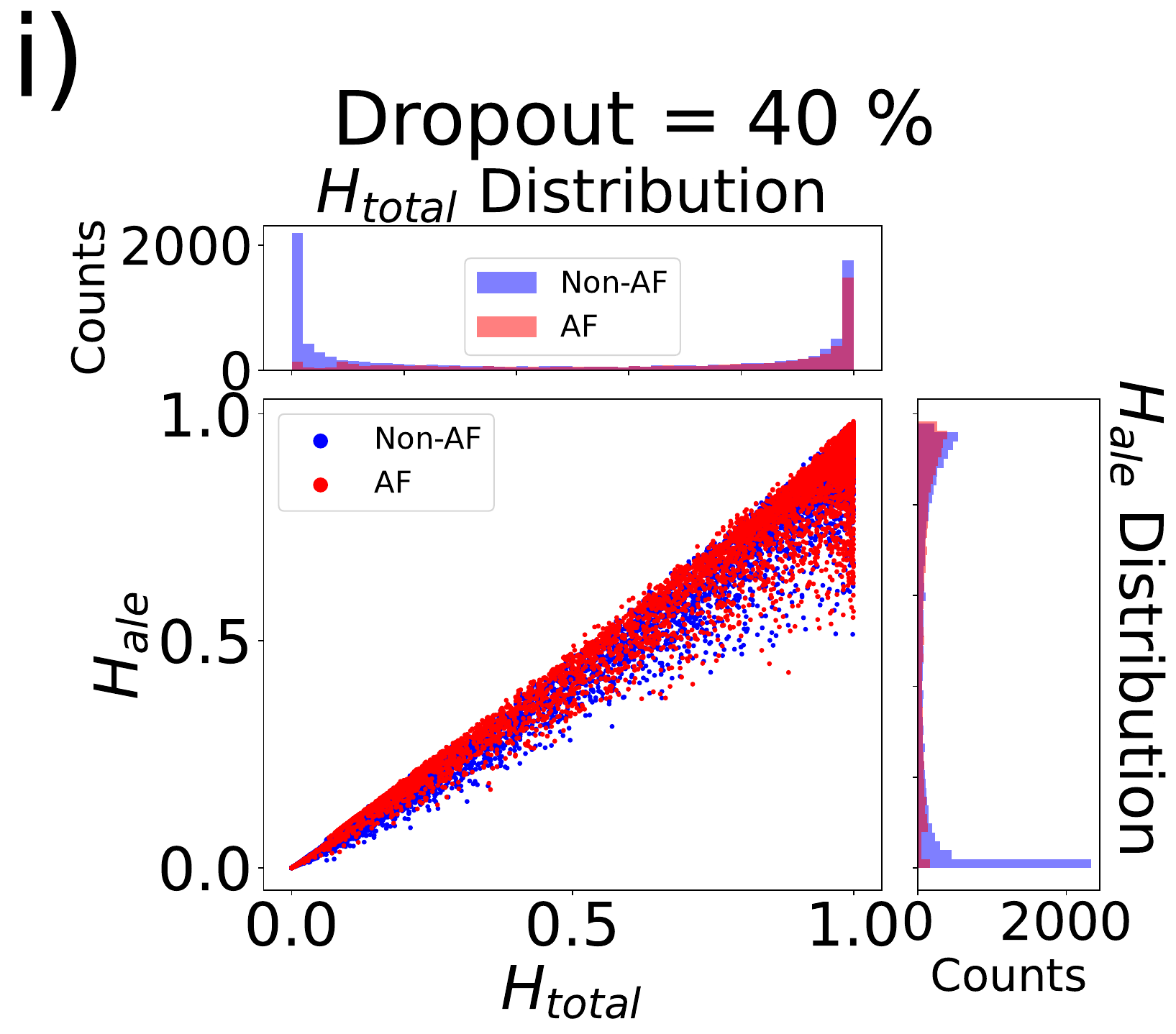}
    \caption{Evaluation of uncertainty calibration for Atrial Fibrillation (AF) classification models trained with Monte carlo Dropout (MCD). a-c) show the uncertainty calibration curves, d-f) show the calibration curves, and g-i) are scatterplots showing what proportion of the the total uncertainty for each test example that is aleatoric, along with the distribution of predicted uncertainties for each class. We find that the proportion of epistemic uncertainty is higher for models trained with higher dropout rates.}
    \label{fig:MCD_AF_calib}
\end{figure}

We show the results from the model trained and evaluated with dropout rates of 5 \%, 10 \% and 40 \% in Table \ref{tab:performance_metrics_AF_MCD}, and the calibration metrics in Table \ref{tab:mcd_af_metrics}. The models' predictive performance is comparable to those trained on similar datasets \cite{Torres-Soto_Ashley_2020,moulaeifard2025machine}, noting that we employ a custom split of our dataset \cite{moulaeifard2025machine}. The reliability diagrams shown in Figs. \ref{fig:MCD_AF_calib} a-f) illustrate the importance of using adaptive calibration metrics to assess model performance; we find significant differences in the classification curves when assessed over the whole population of test examples compared to when they are evaluated per class.

This assessment of adaptivity reveals interesting behaviours of the model. For example, very low uncertainty predictions tend to correspond with predictions of non-AF; the use of only very confident predictions would result in high instances of false-negative diagnoses. The scatterplots in Figs. \ref{fig:MCD_AF_calib} g-i) show the aleatoric vs.\ total uncertainty estimated for each test example, along with histograms showing the distributions of the estimated values. A significant proportion of total uncertainty estimates are found in the most extreme uncertainty bins. The models do not appear to predict noticeably more aleatoric uncertainty for a given class. The use of a higher dropout rate increases the proportion of the total uncertainty that is epistemic, illustrating how the value of the dropout rate used during training/evaluation has a considerable impact on the how various sources of uncertainty are modelled. With that said, without verifying our method for disentangling uncertainties, we cannot be certain that the observed increases in our predicted epistemic uncertainty is truly epistemic uncertainty. Our suspicions on this point align with prior reports outlining the challenges with disentangling uncertainties estimated with MCD \cite{mucsanyi2024benchmarking}.

We find that smaller dropout rates produce better calibrated entropies. Larger dropout rates resulted in better calibrated predicted probabilities (according to each model's UCE and ECE scores). In the multi-class case, prioritising the UCE when attempting to choose the model parameterisation that results in the highest calibration quality is preferable given this accounts for the full distribution of probabilities. However, for binary classification, this choice is less clear. One important consideration could be whether the chosen model can accommodate larger dropout rates; i.e.\ if a model cannot tolerate dropout, then the UCE may be preferable given optimal calibration for this task appeared to occur with less regularisation. However, it is not clear whether this behaviour generalises to any given task. Indeed, this point has been mentioned in other works \cite{seligmann2024beyond}.

\begin{table}[h]
\centering
AF MCD \\
\begin{tabular}{|l|c|c|c|c|c|c|}
\hline
\textbf{Dropout} & \textbf{AUC} & \textbf{F1} & \textbf{MCC Sens} & \textbf{MCC Spec} & \textbf{Sens} & \textbf{Spec} \\
\hline
5 \%& 0.84 & 0.70 & 0.50 & 0.50 & 0.70 & 0.72 \\
\hline

10 \% & 0.85 & 0.72 & 0.53 & 0.54 & 0.74 & 0.74 \\
\hline

40 \% & 0.85 & 0.72 & 0.55 & 0.54 & 0.75 & 0.76 \\
\hline
\end{tabular}
\caption{Test metrics for evaluating the performance of the Atrial Fibrillation (AF) classifiers trained with Monte Carlo Dropout (MCD).}
\label{tab:performance_metrics_AF_MCD}
\end{table}

\begin{table}[h]
\centering
Uncertainty calibration for AF classification: MCD
\begin{tabular}{|l|c|c|c|c|c|c|}
\hline
\textbf{Dropout Rate} & \textbf{Total UCE} & \textbf{non-AF UCE} & \textbf{AF UCE} & \textbf{Total ECE} & \textbf{non-AF ECE} & \textbf{AF ECE} \\
\hline

5 \% & 0.053 & 0.052& 0.198&  0.064 & 0.097& 0.114\\
\hline
10 \% & 0.067 & 0.024 & 0.176 & 0.055 & 0.060 &0.108\\
\hline
40 \% &0.096 &0.055 &0.212 & 0.036& 0.047& 0.155\\
\hline
\end{tabular}
\caption{Uncertainty calibration metrics for Atrial Fibrillation (AF) classification models trained with Monte Carlo Dropout (MCD).}
\label{tab:mcd_af_metrics}
\end{table}

\begin{table}[h]
\centering
AF IVON \\
\begin{tabular}{|l|c|c|c|c|c|c|}
\hline
\textbf{$h_0$} & \textbf{AUC} & \textbf{F1} & \textbf{MCC Sens} & \textbf{MCC Spec} & \textbf{Sens} & \textbf{Spec} \\
\hline
0.001 & 0.80 & 0.64 & 0.44 & 0.42 & 0.62 & 0.65 \\
\hline
0.01 & 0.81 & 0.69 & 0.46 & 0.42 & 0.62 & 0.68 \\
\hline
0.5 & 0.78 & 0.65 & 0.40 & 0.36 & 0.55 & 0.61 \\
\hline
\end{tabular}
\caption{Test metrics for evaluating the performance of the Atrial Fibrillation (AF) classifiers trained with IVON.}
\label{tab:performance_metrics_AF_IVON}
\end{table}

\begin{table}[h]
\centering
Uncertainty calibration for AF classification: IVON
\begin{tabular}{|l|c|c|c|c|c|c|}
\hline
\textbf{$h_0$} & \textbf{Total UCE} & \textbf{non-AF UCE} & \textbf{AF UCE} & \textbf{Total ECE} & \textbf{non-AF ECE} & \textbf{AF ECE} \\
\hline
0.001  & 0.106 & 0.118 &0.102 & 0.026& 0.036& 0.089 \\
\hline

0.01  & 0.040 & 0.127&0.252 & 0.073& 0.167&0.170\\
\hline

0.5 &0.044 & 0.0821& 0.188& 0.074& 0.119&0.122\\
\hline
\end{tabular}
\caption{Uncertainty calibration metrics for Atrial Fibrillation (AF) classification models trained with IVON.}
\label{tab:IVON_AF_metrics}
\end{table}

\begin{figure}
    \centering
    \Large{\sffamily AF Classifier: IVON}\\
    \includegraphics[width=.32\columnwidth]{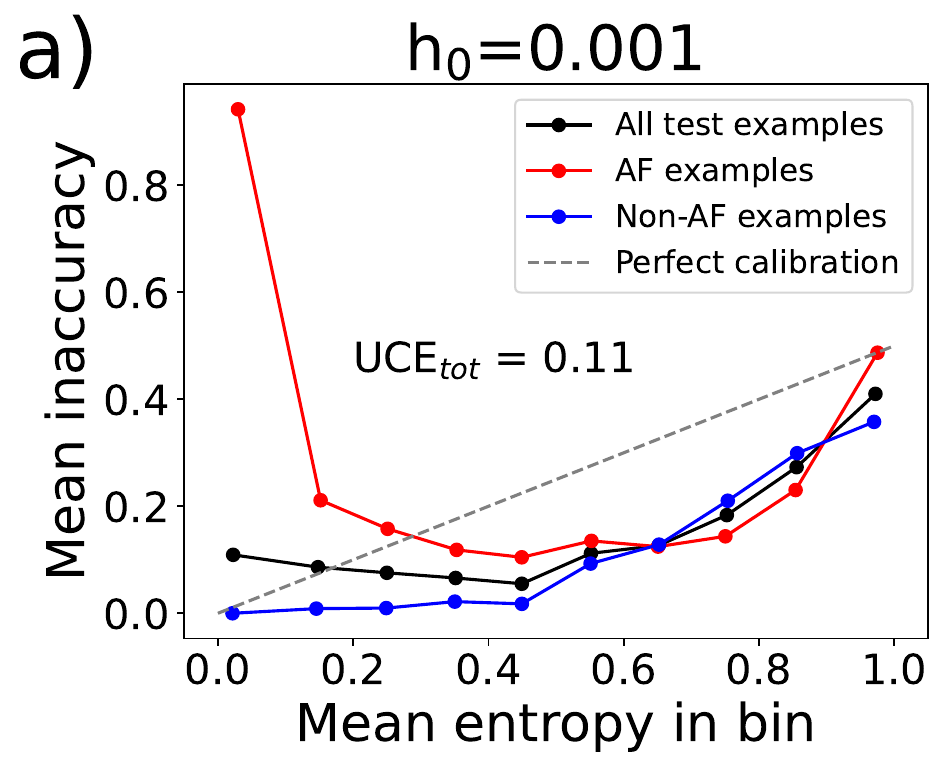}
    \includegraphics[width=.32\columnwidth]{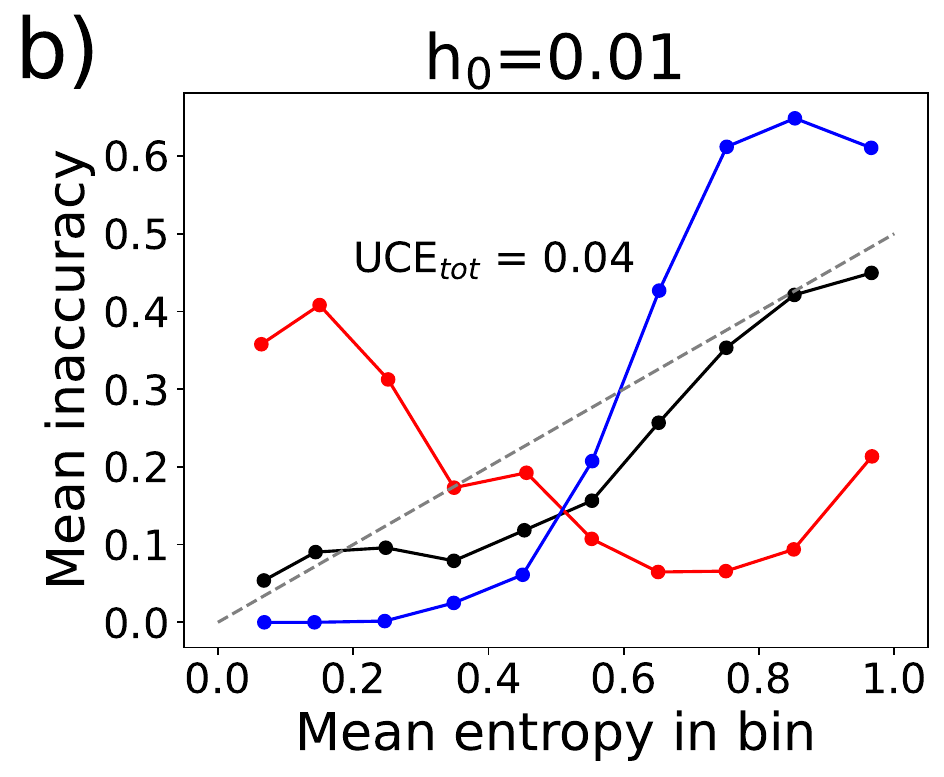}
    \includegraphics[width=.32\columnwidth]{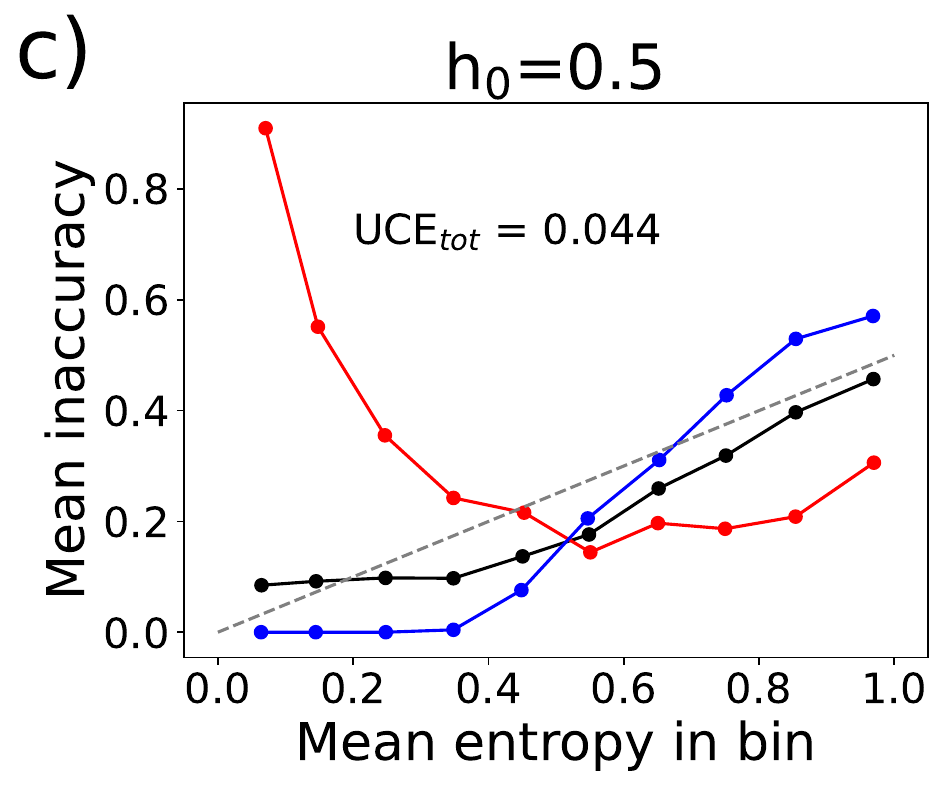}
    \includegraphics[width=.32\columnwidth]{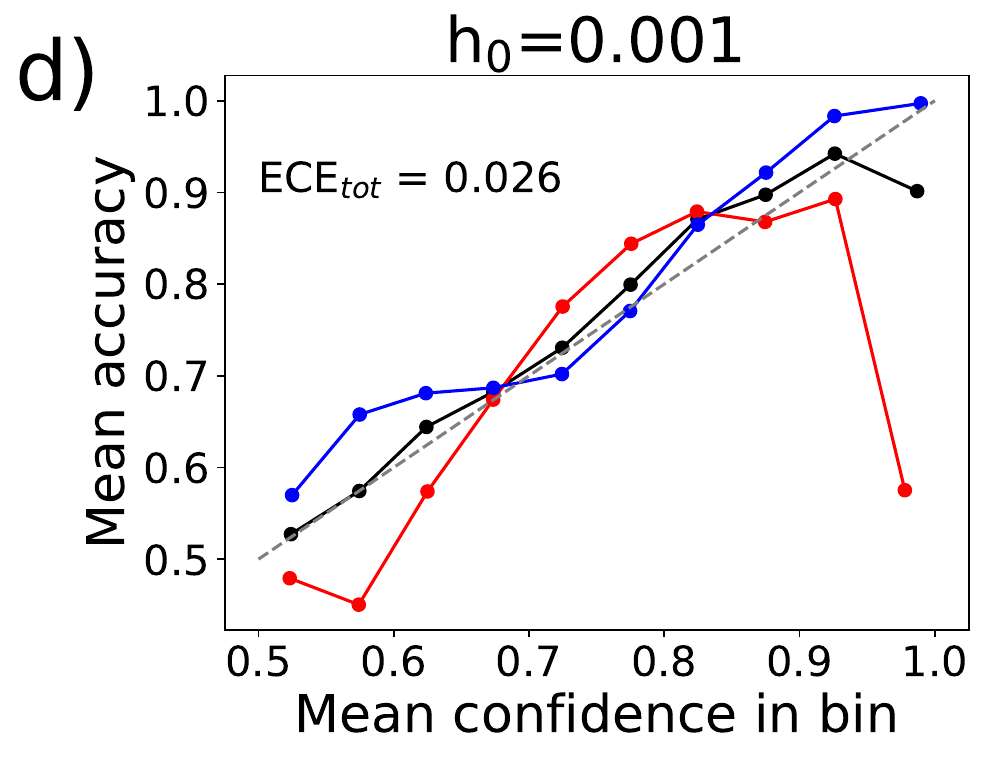}
    \includegraphics[width=.32\columnwidth]{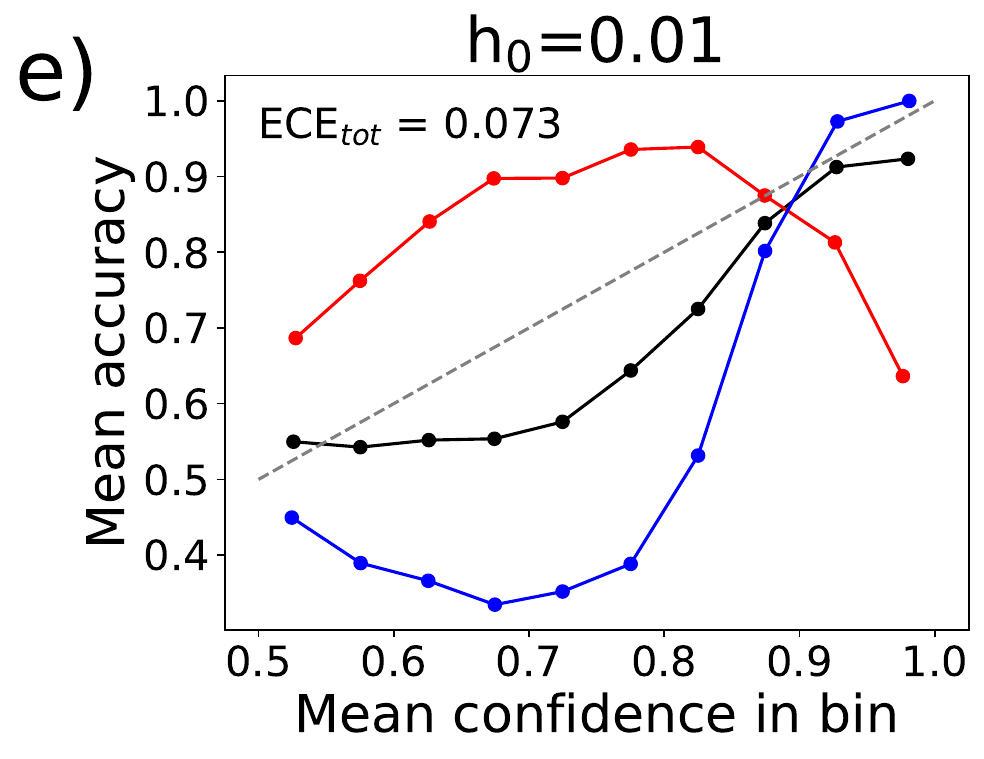}
    \includegraphics[width=.32\columnwidth]{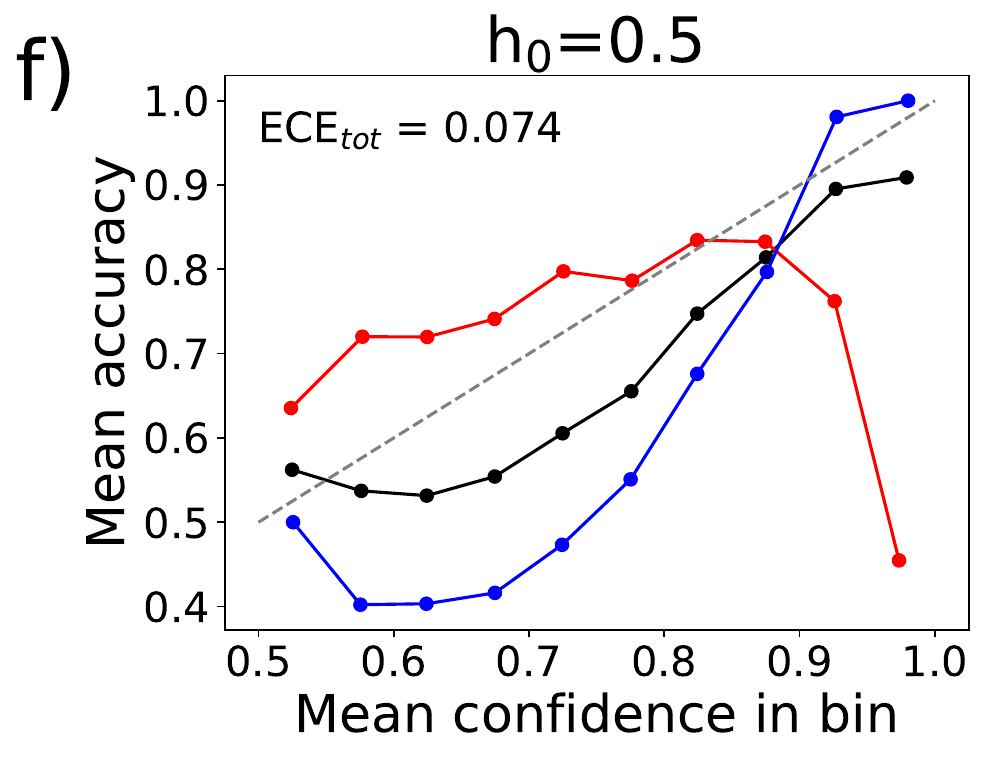}
    \includegraphics[width=.32\columnwidth]{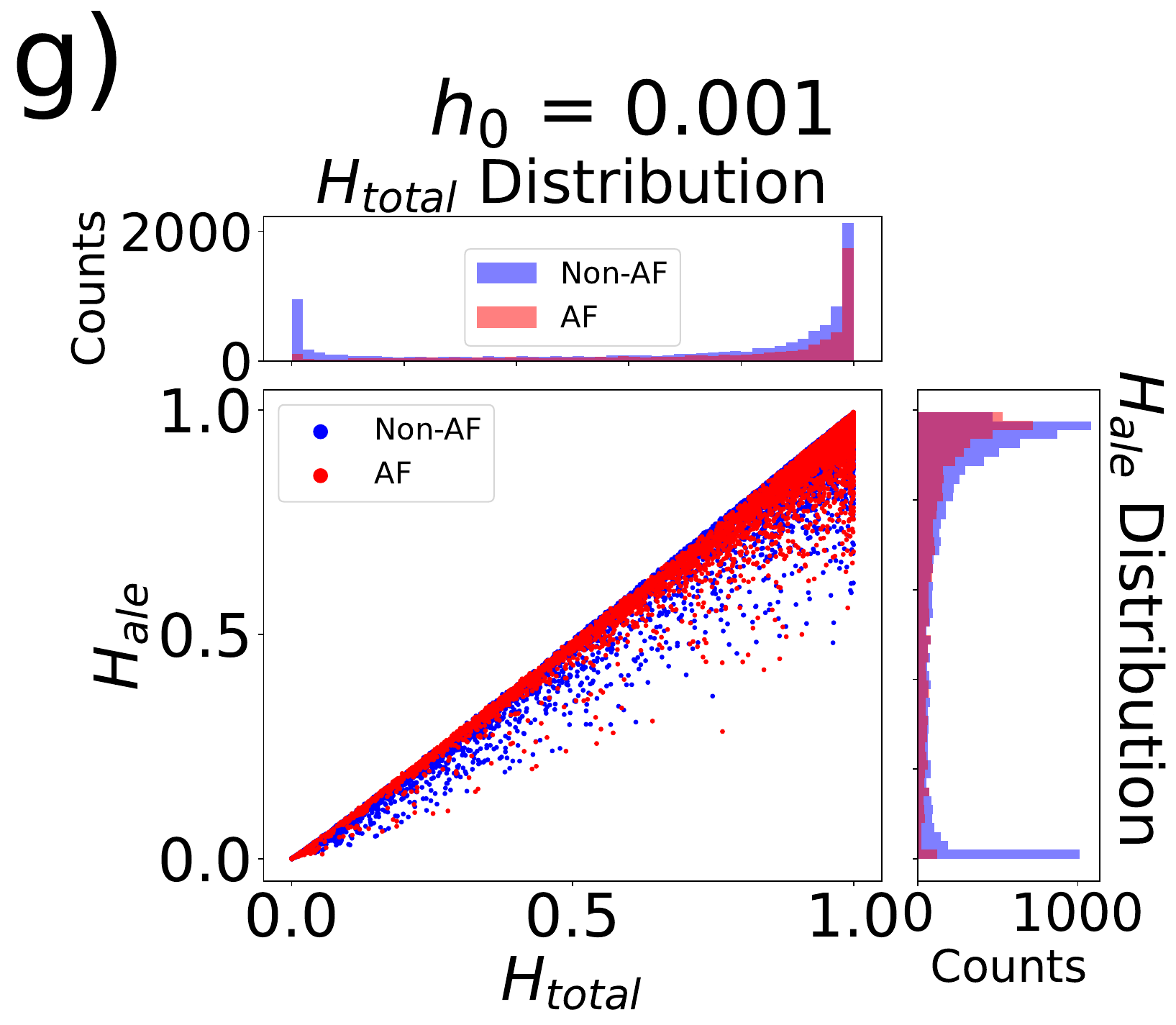}
    \includegraphics[width=.32\columnwidth]{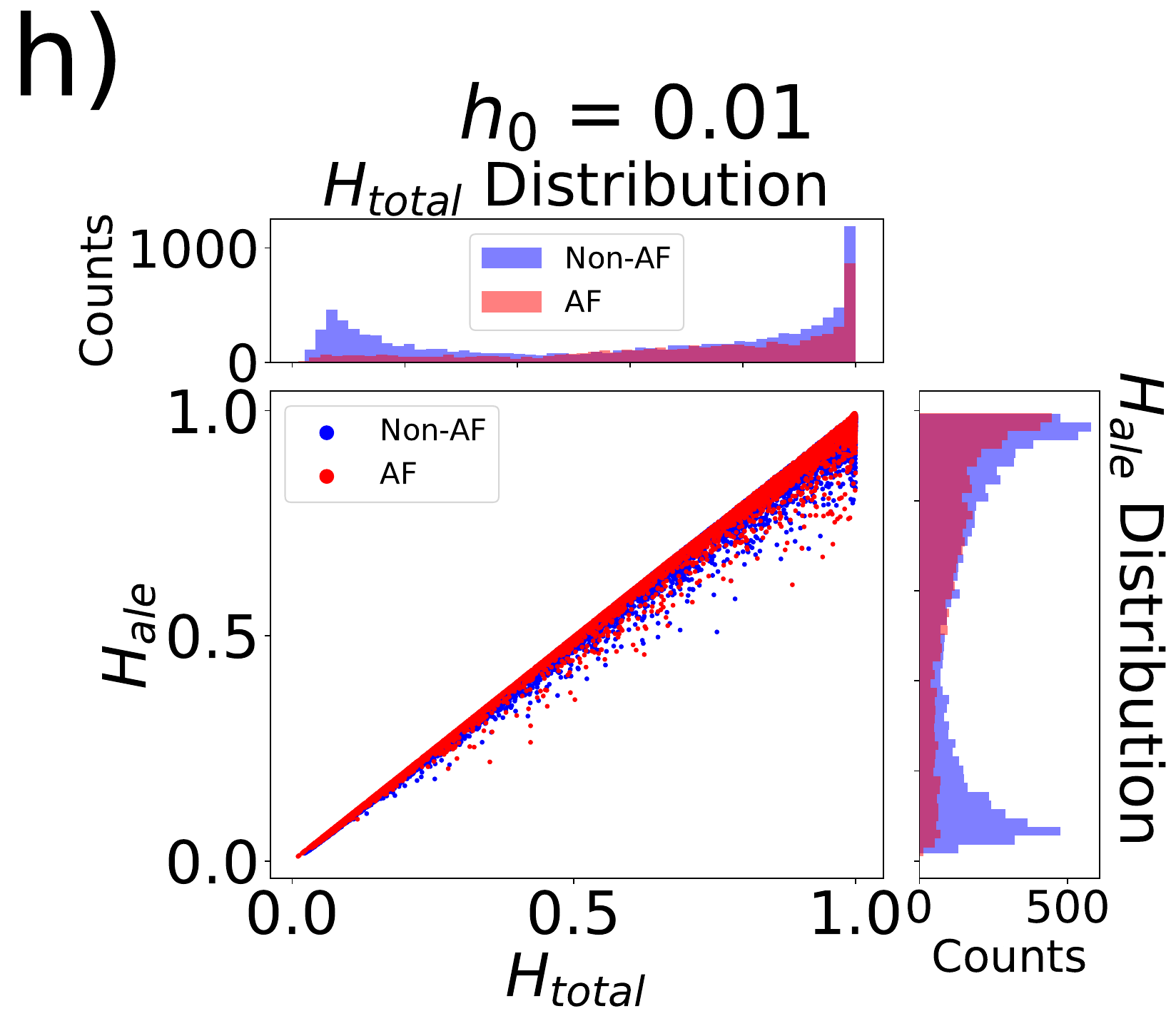}
    \includegraphics[width=.32\columnwidth]{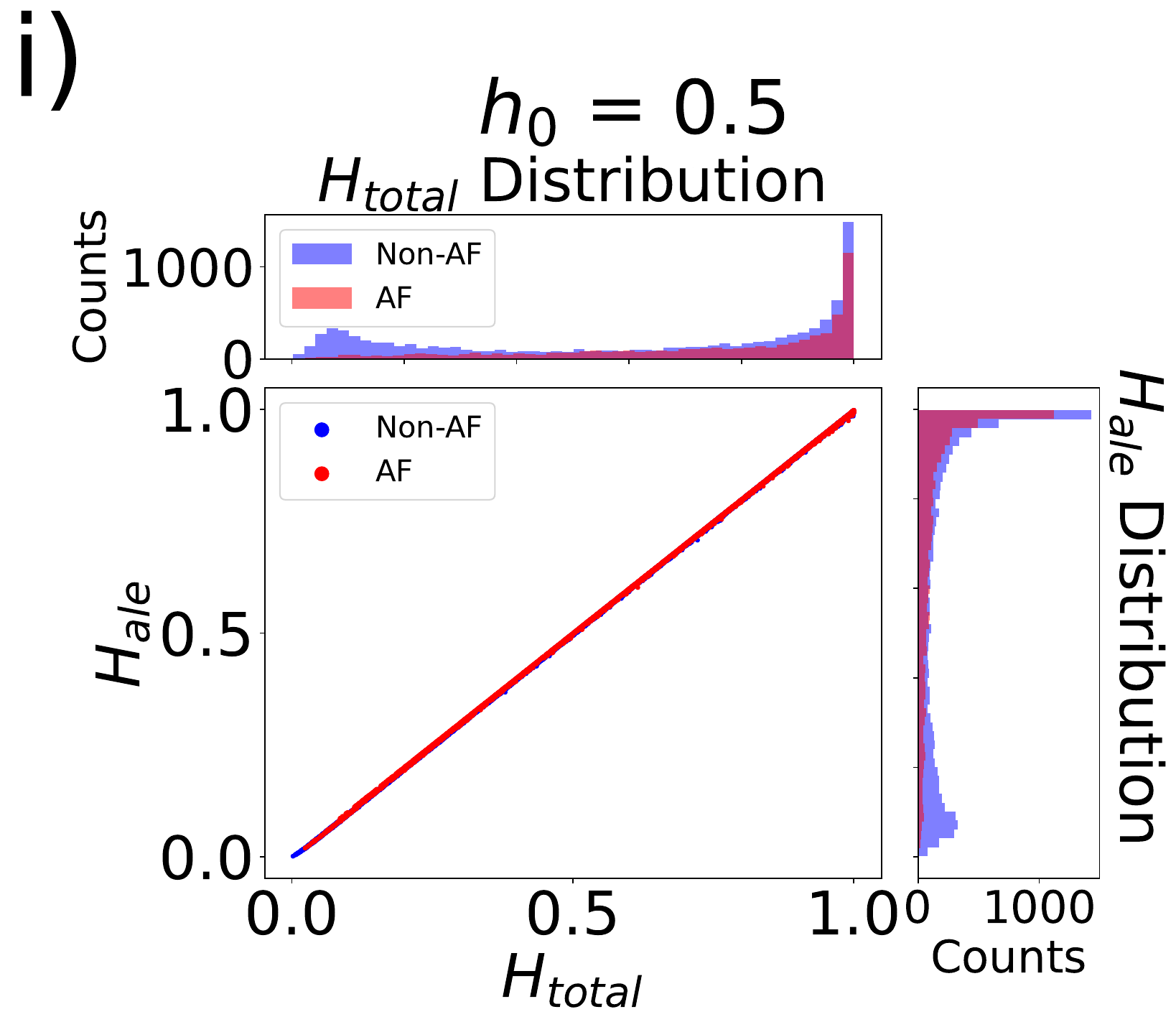}
    \caption{Evaluation of uncertainty calibration for Atrial Fibrillation (AF) classification models trained with IVON. a-c) show the uncertainty calibration curves, d-f) show the calibration curves, and g-i) are scatterplots showing what proportion of the the total uncertainty for each test example that is aleatoric, along with the distribution of predicted uncertainties for each class. We find that the proportion of epistemic uncertainty is higher for models trained with lower $h_0$ values.}
    \label{fig:scatter_ivon_htot_hale}
\end{figure}

\subsection{IVON}
While predictive performance is lower than that achieved with MCD (as seen in Table \ref{tab:performance_metrics_AF_IVON}), we find the most optimally performing models have comparable calibration quality (assessed over the whole population of test examples).
The scatterplots in Fig. \ref{fig:scatter_ivon_htot_hale} show that the Hessian initialisation has an effect akin to the role the dropout rate plays in MCD in determining how much of the total estimated uncertainty is epistemic. Interestingly, we also find that greater stochasticity in model sampling with lower $h_0$ values result in better calibrated ECEs, akin to what was observed with larger dropout rates for MCD. This suggests that, for this task, IVON may be preferable in cases where one wishes to calibrate the predicted probabilities for a model that does not tolerate dropout regularisation.

\section{Discussion}

Deep networks have clear potential to enable the continuous monitoring of AF and BP. However, some indication of the uncertainty in a given prediction is needed to determine whether they may reliably inform diagnoses. Here, we have considered the use of MCD for uncertainty quantification given that it is appealingly inexpensive and straightforward to implement. We also consider IVON for the classification task, which is a natural gradient method. 

We have observed that the quality of uncertainty estimates acquired using MCD for both the regression and classification problems are strongly influenced by the choice of dropout rate. Interestingly, we found that larger dropout rates increase how much of the total uncertainty was composed of epistemic uncertainty. We found similar behaviour when decreasing the initialisation of the Hessian for models trained with IVON, which is known to decrease the concentration of the initial posterior. These results suggest that the degree of stochasticity in the model sampling plays a significant role in determining the magnitude of the estimated epistemic uncertainty. We also found that more stochasticity in the sampling (i.e.\ higher dropout rate, or lower Hessian initialisation) resulted in better calibrated output probabilities, as determined from the ECE scores.

Although MCD is advantageous for large networks (requiring only a single training run over a generic loss function), the quality of estimated uncertainties can be quite poor. One hypothesised reason for this is that the technique is only capable of approximating/sampling from small regions of the posterior (i.e.\ a single mode) \cite{fort2019deep}.

While we assume that the uncertainty disentanglement approaches are effective, we found that for the regression task the estimated aleatoric uncertainty was highly correlated with the estimated epistemic uncertainty. This suggests that the two are not completely disentangled, which is in line with recent analysis investigating the use of this disentanglement formula on uncertainties estimated with a range of uncertainty quantification techniques \cite{mucsanyi2024benchmarking}. Therefore, we advise considerable caution when using uncertainty estimates that have been disentangled using this approach to inform decisions on model configuration or dataset curation.

The use of several evaluation metrics indicates a complex picture of uncertainty calibration for each model. For classification, we find that the per-class calibration may differ considerably from that computed over the whole population of test examples, which highlights the importance of assessing adaptivity. For example, an assessment of calibration over the whole test set for the model trained with a dropout rate of 5 \% does not reveal how the use of low uncertainty estimates would result in high instances of false negatives.

For the regression models, global and local calibration metrics (e.g.\ the coverage-based reliability score, and ENCE) indicate some degree of calibration for most models, but the bivariate histograms qualitatively suggest poor individual calibration across all models. This is concerning, as individual calibration is the most practically relevant metric, given clinical decisions may be derived from single estimates.

\section{Conclusion}
In this paper, we have implemented MCD and IVON on AF classification models trained to predict physiological parameters from raw PPG time-series. We have also shown the results of applying MCD to BP regression models. We have shown that the quality and composition of uncertainties estimated with either technique depend considerably on the hyperparameters used for training. Given that the use of different hyperparameters may also affect the predictive performance of the models, we advise that a range of model parameterisations should be investigated to acquire the values that provide the best combination of uncertainty calibration and predictive performance needed for a given task. We also found that greater stochasticity in model sampling for both techniques results in better calibrated class probabilities (as assessed over the whole population of test examples). Therefore, the optimal model parameterisation depends on the chosen expression of uncertainty.

Furthermore, we emphasise the need to thoroughly evaluate calibration quality, assessing individual calibration as well as adaptivity (e.g.\ assessing calibration per class) where possible. This is needed to best inform decisions about the viability of uncertainty estimates for clinical diagnosis. While we study the composition of uncertainty estimates using both standard and custom methods for disentangling aleatoric and epistemic uncertainty, we advise using caution when interpreting disentangled estimates given the observed correlations between aleatoric and epistemic uncertainty. 

This work ultimately describes important considerations when implementing and evaluating these uncertainty quantification techniques, and provides interesting observations derived from models trained to predict physiological parameters from raw PPG signals. This work provides a useful starting point for establishing principles of trustworthy model development in medical sensing, and helps to realise the potential new PPG sensing technologies can have for continuous monitoring of important physiological parameters outside of the clinic. 

\section*{Code availability}
The software associated with this work is under preparation for public release.
  
\section*{Data availability}
Information about how to access the datasets used in this work can be found in \cite{moulaeifard2025machine}.

\section*{Ethics}
All of the datasets used in this study are freely available/open access.

\section*{Acknowledgments}
The project (22HLT01 QUMPHY) has received funding from the European Partnership on Metrology, co-financed from the European Union’s Horizon Europe Research and Innovation Programme and by the Participating States. Funding for NPL was provided by Innovate UK under the Horizon Europe Guarantee Extension, grant number 10084125.

\section*{Appendix}
\subsection{Model architectures/parameterisations}
For AF classification, we used a generic convolutional network composed of five 1D convolutional layers with a kernel length of 8, each outputting the following number of activation maps: 128$\rightarrow$64$\rightarrow$32$\rightarrow$16$\rightarrow$1. The activations were passed through a Leaky ReLU activation function (with a negative slope of 0.01), followed by dropout and a 1D maxpool with a kernel length of 2 and a stride of 2. The outputs of the final convolutional layer were flattened and processed by a fully connected layer to output four values. No activation or dropout was applied to the outputs of the fully connected layer. We used stochastic gradient descent when implementing MCD with a learning rate of $1e\mathord{-}3$, weight decay of $1e\mathord{-}10$, batch size of 64, and momentum of 0.9. The hyperparameters for optimising the model with SGD are reported to be similar to those that should work when using IVON. For IVON we used a learning rate of $2e\mathord{-}2$, weight decay of $1e\mathord{-}10$, batch size of 128, and momentum of 0.9.

For BP regression, we found that the architecture used for AF classification did not have enough capacity to effectively learn the task with dropout regularisation. We instead employed a 1D ResNet-style architecture. The input is first processed by a 1D convolutional layer (kernel length 7, with a stride of 2, and 64 output activations), whose outputs are processed with batch norm, then dropout and then a ReLU activation, followed by a max pooling layer with a kernel size of 3 and a stride 2. The outputs of this initial convolutional layer are then processed by a sequence of residual blocks. Each residual block is composed of two convolutional layers, each followed by batch normalisation and then dropout, where a ReLU activation is applied to the first layer in the block. The appropriately downsampled input to this layer is added to the post-dropout outputs of the second convolutional layer, followed by a ReLU activation. The model is composed of 8 residual blocks, where each convolutional layer has a kernel length of 9, and where the stride is 1 for the first convolutional layer and 2 for the second. The convolutions in each block output the following number of activation maps: 64$\rightarrow$64$\rightarrow$128$\rightarrow$128$\rightarrow$256$\rightarrow$256$\rightarrow$1$\rightarrow$1. The outputs of the final block were average pooled/flattened with an output length of 100 for each activation map. This was then put through a fully connected layer with four output nodes and Softplus activations.  We used Adam optimisation with a learning rate of $5e-5$, weight decay of $1e-8$, batch size of 64, and momentum of 0.9.

\bibliographystyle{unsrt}  
\bibliography{references}

\begin{thebibliography}{10}

\bibitem{park2022photoplethysmogram}
Junyung Park, Hyeon~Seok Seok, Sang-Su Kim, and Hangsik Shin.
\newblock {Photoplethysmogram analysis and applications: an integrative review}.
\newblock {\em {Frontiers in Physiology}}, 12:808451, 2022.

\bibitem{charlton2022wearable}
Peter~H Charlton, Panicos~A Kyriacou, Jonathan Mant, Vaidotas Marozas, Phil Chowienczyk, and Jordi Alastruey.
\newblock {Wearable photoplethysmography for cardiovascular monitoring}.
\newblock {\em {Proceedings of the {IEEE}}}, 110(3):355--381, 2022.

\bibitem{Torres-Soto_Ashley_2020}
Jessica Torres-Soto and Euan~A. Ashley.
\newblock {Multi-task deep learning for cardiac rhythm detection in wearable devices}.
\newblock {\em {{NPJ} Digital Medicine}}, 3(1):116, September 2020.

\bibitem{shcherbina2017accuracy}
Anna Shcherbina, C~Mikael Mattsson, Daryl Waggott, Heidi Salisbury, Jeffrey~W Christle, Trevor Hastie, Matthew~T Wheeler, and Euan~A Ashley.
\newblock {Accuracy in wrist-worn, sensor-based measurements of heart rate and energy expenditure in a diverse cohort}.
\newblock {\em {Journal of {P}ersonalized {M}edicine}}, 7(2):3, 2017.

\bibitem{wasserlauf2019smartwatch}
Jeremiah Wasserlauf, Cindy You, Ruchi Patel, Alexander Valys, David Albert, and Rod Passman.
\newblock {Smartwatch performance for the detection and quantification of {A}trial {F}ibrillation}.
\newblock {\em {Circulation: Arrhythmia and Electrophysiology}}, 12(6):e006834, 2019.

\bibitem{krakoff2016blood}
Lawrence~R Krakoff.
\newblock {Blood pressure out of the office: its time has finally come}.
\newblock {\em {American Journal of Hypertension}}, 29(3):289--295, 2016.

\bibitem{o2008ambulatory}
Eoin O'Brien.
\newblock {Ambulatory blood pressure measurement: the case for implementation in primary care}.
\newblock {\em {Hypertension}}, 51(6):1435--1441, 2008.

\bibitem{piper2015diagnostic}
Margaret~A Piper, Corinne~V Evans, Brittany~U Burda, Karen~L Margolis, Elizabeth O'Connor, and Evelyn~P Whitlock.
\newblock {Diagnostic and predictive accuracy of blood pressure screening methods with consideration of rescreening intervals: a systematic review for the US Preventive Services Task Force}.
\newblock {\em {Annals of {I}nternal {M}edicine}}, 162(3):192--204, 2015.

\bibitem{parati2008european}
Gianfranco Parati, George~S Stergiou, Roland Asmar, Grzegorz Bilo, Peter De~Leeuw, Yutaka Imai, Kazuomi Kario, Empar Lurbe, Athanasios Manolis, Thomas Mengden, et~al.
\newblock {European Society of Hypertension guidelines for blood pressure monitoring at home: a summary report of the Second International Consensus Conference on Home Blood Pressure Monitoring}.
\newblock {\em {Journal of {H}ypertension}}, 26(8):1505--1526, 2008.

\bibitem{zungsontiporn2018newer}
Nath Zungsontiporn and Mark~S Link.
\newblock {Newer technologies for detection of {A}trial {F}ibrillation}.
\newblock {\em {BMJ}}, 363, 2018.

\bibitem{ding2020emerging}
Eric~Y Ding, Gregory~M Marcus, and David~D McManus.
\newblock {Emerging technologies for identifying {A}trial {F}ibrillation}.
\newblock {\em {Circulation {R}esearch}}, 127(1):128--142, 2020.

\bibitem{almarshad2022diagnostic}
Malak~Abdullah Almarshad, Md~Saiful Islam, Saad Al-Ahmadi, and Ahmed~S BaHammam.
\newblock {Diagnostic features and potential applications of {PPG} signal in healthcare: A systematic review}.
\newblock In {\em Healthcare}, volume~10, page 547. MDPI, 2022.

\bibitem{brillante2008arterial}
Divina~G Brillante, Anthony~J O'sullivan, and Laurence~G Howes.
\newblock {Arterial stiffness indices in healthy volunteers using non-invasive digital photoplethysmography}.
\newblock {\em {Blood {P}ressure}}, 17(2):116--123, 2008.

\bibitem{lei2020estimation}
Ruisheng Lei, Bingo Wing-Kuen Ling, Peihua Feng, and Jinrong Chen.
\newblock {Estimation of heart rate and respiratory rate from {PPG} signal using complementary ensemble empirical mode decomposition with both independent component analysis and non-negative matrix factorization}.
\newblock {\em {Sensors}}, 20(11):3238, 2020.

\bibitem{nitzan2014calibration}
Meir Nitzan, Salman Noach, Elias Tobal, Yair Adar, Yaacov Miller, Eran Shalom, and Shlomo Engelberg.
\newblock {Calibration-free pulse oximetry based on two wavelengths in the infrared—A preliminary study}.
\newblock {\em {{S}ensors}}, 14(4):7420--7434, 2014.

\bibitem{gonzalez2023benchmark}
Sergio Gonz{\'a}lez, Wan-Ting Hsieh, and Trista Pei-Chun Chen.
\newblock {A benchmark for machine-learning based non-invasive blood pressure estimation using photoplethysmogram}.
\newblock {\em {Scientific Data}}, 10(1):149, 2023.

\bibitem{el2020review}
Chadi El-Hajj and Panayiotis~A Kyriacou.
\newblock {A review of machine learning techniques in photoplethysmography for the non-invasive cuff-less measurement of blood pressure}.
\newblock {\em {Biomedical Signal Processing and Control}}, 58:101870, 2020.

\bibitem{maqsood2022survey}
Sumbal Maqsood, Shuxiang Xu, Son Tran, Saurabh Garg, Matthew Springer, Mohan Karunanithi, and Rami Mohawesh.
\newblock {A survey: From shallow to deep machine learning approaches for blood pressure estimation using biosensors}.
\newblock {\em {Expert Systems with Applications}}, 197:116788, 2022.

\bibitem{paviglianiti2022comparison}
Annunziata Paviglianiti, Vincenzo Randazzo, Stefano Villata, Giansalvo Cirrincione, and Eros Pasero.
\newblock {A comparison of deep learning techniques for arterial blood pressure prediction}.
\newblock {\em {Cognitive Computation}}, 14(5):1689--1710, 2022.

\bibitem{shashikumar2017deep}
Supreeth~Prajwal Shashikumar, Amit~J Shah, Qiao Li, Gari~D Clifford, and Shamim Nemati.
\newblock {A deep learning approach to monitoring and detecting {A}trial {F}ibrillation using wearable technology}.
\newblock In {\em 2017 {IEEE} EMBS {I}nternational {C}onference on {B}iomedical \& {H}ealth {I}nformatics (BHI)}, pages 141--144. {IEEE}, 2017.

\bibitem{tison2018passive}
Geoffrey~H Tison, Jos{\'e}~M Sanchez, Brandon Ballinger, Avesh Singh, Jeffrey~E Olgin, Mark~J Pletcher, Eric Vittinghoff, Emily~S Lee, Shannon~M Fan, Rachel~A Gladstone, et~al.
\newblock {Passive detection of {A}trial {F}ibrillation using a commercially available smartwatch}.
\newblock {\em {{JAMA} {C}ardiology}}, 3(5):409--416, 2018.

\bibitem{shen2019ambulatory}
Yichen Shen, Maxime Voisin, Alireza Aliamiri, Anand Avati, Awni Hannun, and Andrew Ng.
\newblock {Ambulatory {A}trial {F}ibrillation monitoring using wearable photoplethysmography with deep learning}.
\newblock In {\em Proceedings of the 25th {ACM} {SIGKDD} {I}nternational {C}onference on {K}nowledge {D}iscovery \& {D}ata {M}ining}, pages 1909--1916, 2019.

\bibitem{gotlibovych2018end}
Igor Gotlibovych, Stuart Crawford, Dileep Goyal, Jiaqi Liu, Yaniv Kerem, David Benaron, Defne Yilmaz, Gregory Marcus, and Yihan Li.
\newblock {End-to-end deep learning from raw sensor data: Atrial {F}ibrillation detection using wearables}.
\newblock {\em {ar{X}iv preprint ar{X}iv:1807.10707}}, 2018.

\bibitem{kurylyak2013neural}
Yuriy Kurylyak, Francesco Lamonaca, and Domenico Grimaldi.
\newblock {A Neural Network-based method for continuous blood pressure estimation from a {PPG} signal}.
\newblock In {\em 2013 {IEEE} International {I}nstrumentation and {M}easurement {T}echnology {C}onference ({I2MTC})}, pages 280--283. {IEEE}, 2013.

\bibitem{witkovsky2017brief}
V~Witkovsk{\`y}, G~Wimmer, Z~{\v{D}}uri{\v{s}}ov{\'a}, S~{\v{D}}uri{\v{s}}, and R~Palen{\v{c}}{\'a}r.
\newblock {Brief overview of methods for measurement uncertainty analysis: {GUM} uncertainty framework, {M}onte {C}arlo method, characteristic function approach}.
\newblock In {\em 2017 11th International Conference on Measurement}, pages 35--38. {IEEE}, 2017.

\bibitem{gal2016dropout}
Yarin Gal and Zoubin Ghahramani.
\newblock {Dropout as a {B}ayesian approximation: Representing model uncertainty in deep learning}.
\newblock In {\em international conference on machine learning}, pages 1050--1059. PMLR, 2016.

\bibitem{hullermeier2021aleatoric}
Eyke H{\"u}llermeier and Willem Waegeman.
\newblock {Aleatoric and epistemic uncertainty in machine learning: An introduction to concepts and methods}.
\newblock {\em {Machine learning}}, 110(3):457--506, 2021.

\bibitem{gruber2023sources}
Cornelia Gruber, Patrick~Oliver Schenk, Malte Schierholz, Frauke Kreuter, and G{\"o}ran Kauermann.
\newblock Sources of uncertainty in machine learning--a statisticians' view.
\newblock {\em arXiv preprint arXiv:2305.16703}, 2023.

\bibitem{mucsanyi2024benchmarking}
B{\'a}lint Mucs{\'a}nyi, Michael Kirchhof, and Seong~Joon Oh.
\newblock {Benchmarking uncertainty disentanglement: Specialized uncertainties for specialized tasks}.
\newblock {\em {ar{X}iv preprint ar{X}iv:2402.19460}}, 2024.

\bibitem{kendall2017uncertainties}
Alex Kendall and Yarin Gal.
\newblock {What uncertainties do we need in {B}ayesian deep learning for computer vision?}
\newblock {\em {Advances in neural information processing systems}}, 30, 2017.

\bibitem{lakshminarayanan2017simple}
Balaji Lakshminarayanan, Alexander Pritzel, and Charles Blundell.
\newblock {Simple and scalable predictive uncertainty estimation using deep ensembles}.
\newblock {\em {Advances in neural information processing systems}}, 30, 2017.

\bibitem{jospin2022hands}
Laurent~Valentin Jospin, Hamid Laga, Farid Boussaid, Wray Buntine, and Mohammed Bennamoun.
\newblock {Hands-on {B}ayesian neural networks—{A} tutorial for deep learning users}.
\newblock {\em {{IEEE} Computational Intelligence Magazine}}, 17(2):29--48, 2022.

\bibitem{mackay1995bayesian}
David~JC MacKay.
\newblock {{B}ayesian neural networks and density networks}.
\newblock {\em {Nuclear Instruments and Methods in Physics Research Section A: Accelerators, Spectrometers, Detectors and Associated Equipment}}, 354(1):73--80, 1995.

\bibitem{mackay1992practical}
David~JC MacKay.
\newblock {A practical {B}ayesian framework for backpropagation networks}.
\newblock {\em {Neural {C}omputation}}, 4(3):448--472, 1992.

\bibitem{duffield2024scalable}
Samuel Duffield, Kaelan Donatella, Johnathan Chiu, Phoebe Klett, and Daniel Simpson.
\newblock {Scalable {B}ayesian Learning with posteriors}.
\newblock {\em {ar{X}iv preprint ar{X}iv:2406.00104}}, 2024.

\bibitem{osawa2019practical}
Kazuki Osawa, Siddharth Swaroop, Mohammad Emtiyaz~E Khan, Anirudh Jain, Runa Eschenhagen, Richard~E Turner, and Rio Yokota.
\newblock {Practical deep learning with {B}ayesian principles}.
\newblock {\em {Advances in neural information processing systems}}, 32, 2019.

\bibitem{kingma2014adam}
Diederik~P Kingma and Jimmy Ba.
\newblock Adam: A method for stochastic optimization.
\newblock {\em arXiv preprint arXiv:1412.6980}, 2014.

\bibitem{shen2024variational}
Yuesong Shen, Nico Daheim, Bai Cong, Peter Nickl, Gian~Maria Marconi, Clement Bazan, Rio Yokota, Iryna Gurevych, Daniel Cremers, Mohammad~Emtiyaz Khan, et~al.
\newblock {Variational learning is effective for large deep networks}.
\newblock {\em {ar{X}iv preprint ar{X}iv:2402.17641}}, 2024.

\bibitem{kendall2015bayesian}
Alex Kendall, Vijay Badrinarayanan, and Roberto Cipolla.
\newblock {Bayesian {S}eg{N}et: {M}odel uncertainty in deep convolutional encoder-decoder architectures for scene understanding}.
\newblock {\em {ar{X}iv preprint ar{X}iv:1511.02680}}, 2015.

\bibitem{fort2019deep}
Stanislav Fort, Huiyi Hu, and Balaji Lakshminarayanan.
\newblock {Deep ensembles: A loss landscape perspective}.
\newblock {\em {ar{X}iv preprint ar{X}iv:1912.02757}}, 2019.

\bibitem{gal2015bayesian}
Yarin Gal and Zoubin Ghahramani.
\newblock {Bayesian convolutional neural networks with Bernoulli approximate variational inference}.
\newblock {\em {ar{X}iv preprint ar{X}iv:1506.02158}}, 2015.

\bibitem{gal2017concrete}
Yarin Gal, Jiri Hron, and Alex Kendall.
\newblock {Concrete dropout}.
\newblock {\em {Advances in {N}eural {I}nformation {P}rocessing {S}ystems}}, 30, 2017.

\bibitem{amini2020deep}
Alexander Amini, Wilko Schwarting, Ava Soleimany, and Daniela Rus.
\newblock {Deep evidential regression}.
\newblock {\em {Advances in neural information processing systems}}, 33:14927--14937, 2020.

\bibitem{gao2024comprehensive}
Junyu Gao, Mengyuan Chen, Liangyu Xiang, and Changsheng Xu.
\newblock {A {C}omprehensive {S}urvey on {E}vidential {D}eep {L}earning and {I}ts {A}pplications}.
\newblock {\em {ar{X}iv preprint ar{X}iv:2409.04720}}, 2024.

\bibitem{wilson2020bayesian}
Andrew~G Wilson and Pavel Izmailov.
\newblock Bayesian deep learning and a probabilistic perspective of generalization.
\newblock {\em Advances in neural information processing systems}, 33:4697--4708, 2020.

\bibitem{valdenegro2022deeper}
Matias Valdenegro-Toro and Daniel~Saromo Mori.
\newblock {A deeper look into aleatoric and epistemic uncertainty disentanglement}.
\newblock In {\em 2022 {IEEE}/CVF Conference on Computer Vision and Pattern Recognition Workshops (CVPRW)}, pages 1508--1516. IEEE, 2022.

\bibitem{song2023uncertainty}
Rencheng Song, Han Wang, Haojie Xia, Juan Cheng, Chang Li, and Xun Chen.
\newblock {Uncertainty quantification for deep learning-based remote photoplethysmography}.
\newblock {\em {{IEEE} Transactions on Instrumentation and Measurement}}, 2023.

\bibitem{harper2019end}
Ross Harper and Joshua Southern.
\newblock {End-to-end prediction of emotion from heartbeat data collected by a consumer fitness tracker}.
\newblock In {\em 2019 8th {I}nternational {C}onference on {A}ffective {C}omputing and {}ntelligent {I}nteraction ({ACII})}, pages 1--7. {IEEE}, 2019.

\bibitem{trudaquantifying}
Gianluca Truda, Serafim Korovin, and Adam Kantorik.
\newblock {Quantifying Uncertainty in Blood Oxygen Estimation Models from Real-World Data}.

\bibitem{liu2022videocad}
Xuenan Liu, Xuezhi Yang, Rencheng Song, Jie Zhang, and Longwei Li.
\newblock {VideoCAD: an uncertainty-driven neural network for coronary artery disease screening from facial videos}.
\newblock {\em {{IEEE} Transactions on Instrumentation and Measurement}}, 72:1--12, 2022.

\bibitem{asgharnezhad2023improving}
Hamzeh Asgharnezhad, Afshar Shamsi, Ivan Bakhshayeshi, Roohallah Alizadehsani, Somayyeh Chamaani, and Hamid Alinejad-Rokny.
\newblock {Improving {PPG} Signal Classification with Machine Learning: The Power of a Second Opinion}.
\newblock In {\em 2023 24th {I}nternational {C}onference on {D}igital {S}ignal {P}rocessing ({DSP})}, pages 1--5. {IEEE}, 2023.

\bibitem{han2023non}
Xuesong Han, Xuezhi Yang, Shuai Fang, Rencheng Song, Longwei Li, and Jie Zhang.
\newblock {Non-contact blood pressure estimation using BP-related cardiovascular knowledge: an uncalibrated method based on consumer-level camera}.
\newblock {\em {{IEEE} Transactions on Instrumentation and Measurement}}, 2023.

\bibitem{vranken2021uncertainty}
Jeroen~F Vranken, Rutger~R van~de Leur, Deepak~K Gupta, Luis~E Juarez~Orozco, Rutger~J Hassink, Pim van~der Harst, Pieter~A Doevendans, Sadaf Gulshad, and Ren{\'e} van Es.
\newblock {Uncertainty estimation for deep learning-based automated analysis of 12-lead electrocardiograms}.
\newblock {\em {European Heart Journal-Digital Health}}, 2(3):401--415, 2021.

\bibitem{chen2022quantifying}
Brian Chen, Golara Javadi, Alexander Hamilton, Stephanie Sibley, Philip Laird, Purang Abolmaesumi, David Maslove, and Parvin Mousavi.
\newblock {Quantifying deep neural network uncertainty for {A}trial {F}ibrillation detection with limited labels}.
\newblock {\em {Scientific Reports}}, 12(1):20140, 2022.

\bibitem{das2020bayesbeat}
S~Snigdha~Sarathi Das, S~Karmaker Shanto, Masum Rahman, M~Islam, Atif Rahman, Mohammad~Mehedy Masud, and Mohammed~Eunus Ali.
\newblock {Bayes{B}eat: A {B}ayesian deep learning approach for {A}trial {F}ibrillation detection from noisy photoplethysmography data}.
\newblock {\em {ar{X}iv preprint ar{X}iv:2011.00753}}, 2020.

\bibitem{pernot2023validation}
Pascal Pernot.
\newblock {Validation of uncertainty quantification metrics: a primer based on the consistency and adaptivity concepts}.
\newblock 2023.

\bibitem{khan2018fast}
Mohammad Khan, Didrik Nielsen, Voot Tangkaratt, Wu~Lin, Yarin Gal, and Akash Srivastava.
\newblock {Fast and scalable {B}ayesian deep learning by weight-perturbation in adam}.
\newblock In {\em International conference on machine learning}, pages 2611--2620. PMLR, 2018.

\bibitem{kuleshov2018accurate}
Volodymyr Kuleshov, Nathan Fenner, and Stefano Ermon.
\newblock {Accurate uncertainties for deep learning using calibrated regression}.
\newblock In {\em International conference on machine learning}, pages 2796--2804. PMLR, 2018.

\bibitem{levi2022evaluating}
Dan Levi, Liran Gispan, Niv Giladi, and Ethan Fetaya.
\newblock {Evaluating and calibrating uncertainty prediction in regression tasks}.
\newblock {\em {Sensors}}, 22(15):5540, 2022.

\bibitem{guo2017calibration}
Chuan Guo, Geoff Pleiss, Yu~Sun, and Kilian~Q Weinberger.
\newblock {On calibration of modern neural networks}.
\newblock In {\em International conference on machine learning}, pages 1321--1330. PMLR, 2017.

\bibitem{laves2020calibration}
Max-Heinrich Laves, Sontje Ihler, Karl-Philipp Kortmann, and Tobias Ortmaier.
\newblock {Calibration of model uncertainty for dropout variational inference}.
\newblock {\em {ar{X}iv preprint ar{X}iv:2006.11584}}, 2020.

\bibitem{moulaeifard2025machine}
Mohammad Moulaeifard, Loic Coquelin, Mantas Rinkevi{\v{c}}ius, Andrius Solo{\v{s}}enko, Oskar Pfeffer, Ciaran Bench, Nando Hegemann, Sara Vardanega, Manasi Nandi, Jordi Alastruey, et~al.
\newblock Machine-learning for photoplethysmography analysis: Benchmarking feature, image, and signal-based approaches.
\newblock {\em arXiv preprint arXiv:2502.19949}, 2025.

\bibitem{wang2023pulsedb}
Weinan Wang, Pedram Mohseni, Kevin~L Kilgore, and Laleh Najafizadeh.
\newblock {PulseDB: A large, cleaned dataset based on MIMIC-III and VitalDB for benchmarking cuff-less blood pressure estimation methods}.
\newblock {\em {Frontiers in Digital Health}}, 4:1090854, 2023.

\bibitem{pereira2020photoplethysmography}
Tania Pereira, Nate Tran, Kais Gadhoumi, Michele~M Pelter, Duc~H Do, Randall~J Lee, Rene Colorado, Karl Meisel, and Xiao Hu.
\newblock {Photoplethysmography based {A}trial {F}ibrillation detection: a review}.
\newblock {\em {{NPJ} digital medicine}}, 3(1):1--12, 2020.

\bibitem{chong2018motion}
Jo~Woon Chong, Chae~Ho Cho, Fatemehsadat Tabei, Duy Le-Anh, Nada Esa, David~D McManus, and Ki~H Chon.
\newblock {Motion and noise artifact-resilient {A}trial {F}ibrillation detection using a smartphone}.
\newblock {\em {{IEEE} journal on emerging and selected topics in circuits and systems}}, 8(2):230--239, 2018.

\bibitem{corino2017detection}
Valentina~DA Corino, Rita Laureanti, Lorenzo Ferranti, Giorgio Scarpini, Federico Lombardi, and Luca~T Mainardi.
\newblock {Detection of {A}trial {F}ibrillation episodes using a wristband device}.
\newblock {\em {Physiological measurement}}, 38(5):787, 2017.

\bibitem{tang2017identification}
Sung-Chun Tang, Pei-Wen Huang, Chi-Sheng Hung, Shih-Ming Shan, Yen-Hung Lin, Jiann-Shing Shieh, Dar-Ming Lai, An-Yeu Wu, and Jiann-Shing Jeng.
\newblock {Identification of {A}trial {F}ibrillation by quantitative analyses of fingertip photoplethysmogram}.
\newblock {\em {Scientific reports}}, 7(1):1--7, 2017.

\bibitem{mukhoti2020batch}
Jishnu Mukhoti, Puneet~K Dokania, Philip~HS Torr, and Yarin Gal.
\newblock {On batch normalisation for approximate {B}ayesian inference}.
\newblock {\em {ar{X}iv preprint ar{X}iv:2012.13220}}, 2020.

\bibitem{haddad2021continuous}
Serj Haddad, Assim Boukhayma, and Antonino Caizzone.
\newblock {Continuous {PPG}-based blood pressure monitoring using multi-linear regression}.
\newblock {\em {{IEEE} journal of biomedical and health informatics}}, 26(5):2096--2105, 2021.

\bibitem{sola2009parametric}
Josep Sola, Rolf Vetter, Philippe Renevey, Olivier Ch{\'e}telat, Claudio Sartori, and Stefano~F Rimoldi.
\newblock {Parametric estimation of pulse arrival time: a robust approach to pulse wave velocity}.
\newblock {\em {Physiological measurement}}, 30(7):603, 2009.

\bibitem{atomi2017cuffless}
Kengo Atomi, Haruki Kawanaka, Md~Shoaib Bhuiyan, Koji Oguri, et~al.
\newblock {Cuffless blood pressure estimation based on data-oriented continuous health monitoring system}.
\newblock {\em {Computational and mathematical methods in medicine}}, 2017, 2017.

\bibitem{mehta2023can}
Suril Mehta, Nipun Kwatra, Mohit Jain, and Daniel McDuff.
\newblock {"{C}an't {T}ake the {P}ressure?": Examining the Challenges of Blood Pressure Estimation via Pulse Wave Analysis}.
\newblock {\em {ar{X}iv preprint ar{X}iv:2304.14916}}, 2023.

\bibitem{suzuki2008cuffless}
Satomi Suzuki and Koji Oguri.
\newblock {Cuffless and non-invasive systolic blood pressure estimation for aged class by using a photoplethysmograph}.
\newblock In {\em 2008 30th Annual International Conference of the {IEEE} Engineering in Medicine and Biology Society}, pages 1327--1330. {IEEE}, 2008.

\bibitem{li2021central}
Peihao Li and Taous-Meriem Laleg-Kirati.
\newblock {Central blood pressure estimation from distal {PPG} measurement using semiclassical signal analysis features}.
\newblock {\em {{IEEE} Access}}, 9:44963--44973, 2021.

\bibitem{samimi2022cuffless}
Hamed Samimi and Hilmi~R Dajani.
\newblock {Cuffless blood pressure estimation using cardiovascular dynamics}.
\newblock In {\em 2022 International Conference on Electrical, Computer and Energy Technologies (ICECET)}, pages 1--8. {IEEE}, 2022.

\bibitem{samimi2023ppg}
Hamed Samimi and Hilmi~R Dajani.
\newblock {A {PPG}-based calibration-free cuffless blood pressure estimation method using cardiovascular dynamics}.
\newblock {\em {Sensors}}, 23(8):4145, 2023.

\bibitem{ding2015continuous}
Xiao-Rong Ding, Yuan-Ting Zhang, Jing Liu, Wen-Xuan Dai, and Hon~Ki Tsang.
\newblock {Continuous cuffless blood pressure estimation using pulse transit time and photoplethysmogram intensity ratio}.
\newblock {\em {{IEEE} Transactions on Biomedical Engineering}}, 63(5):964--972, 2015.

\bibitem{khalid2018blood}
Syed~Ghufran Khalid, Jufen Zhang, Fei Chen, Dingchang Zheng, et~al.
\newblock {Blood pressure estimation using photoplethysmography only: comparison between different machine learning approaches}.
\newblock {\em {Journal of healthcare engineering}}, 2018, 2018.

\bibitem{gesche2012continuous}
Heiko Gesche, Detlef Grosskurth, Gert K{\"u}chler, and Andreas Patzak.
\newblock {Continuous blood pressure measurement by using the pulse transit time: comparison to a cuff-based method}.
\newblock {\em {European journal of applied physiology}}, 112(1):309--315, 2012.

\bibitem{mase2011feasibility}
Michela Mase, Walter Mattei, Roberta Cucino, Luca Faes, and Giandomenico Nollo.
\newblock {Feasibility of cuff-free measurement of systolic and diastolic arterial blood pressure}.
\newblock {\em {Journal of electrocardiology}}, 44(2):201--207, 2011.

\bibitem{ieee_standard}
{{IEEE} Standard for Wearable, Cuffless Blood Pressure Measuring Devices - Amendment 1}.
\newblock {\em {{IEEE} Std 1708a-2019 (Amendment to {IEEE} Std 1708-2014)}}, pages 1--35, 2019.

\bibitem{pernot2023calibration}
Pascal Pernot.
\newblock {Calibration in Machine Learning Uncertainty Quantification: beyond consistency to target adaptivity}.
\newblock {\em {APL Machine Learning}}, 1(4), 2023.

\bibitem{seligmann2024beyond}
Florian Seligmann, Philipp Becker, Michael Volpp, and Gerhard Neumann.
\newblock {Beyond deep ensembles: A {L}arge-{S}cale {E}valuation of {B}ayesian {D}eep {L}earning {U}nder {D}istribution {S}hift}.
\newblock {\em {Advances in Neural Information Processing Systems}}, 36, 2024.

\end{thebibliography}

\end{document}